\documentclass{article} 
\usepackage{iclr2027_conference,times}
\usepackage{xcolor}

\usepackage[colorlinks=true,linkcolor=magenta,citecolor=blue,urlcolor=blue]{hyperref}
\usepackage{url}

\usepackage{smile}
\usepackage{math}
\usepackage{mathtools}
\usepackage{algorithmic}
\usepackage{titletoc}

\usepackage{mdframed}
\usepackage{etoolbox}
\usepackage{subcaption}

\definecolor{myred}{RGB}{0, 0, 0}
\colorlet{red}{myred}
\newcommand{\myred}{\color{myred}}

\definecolor{shadecolor}{gray}{0.9}
\definecolor{revcolor}{HTML}{8B1A89}

\newmdenv[
skipabove=3pt,
skipbelow=-16pt,
hidealllines=true,
backgroundcolor=shadecolor,
innerleftmargin=2pt,
innerrightmargin=2pt,
innertopmargin=3.5pt,
innerbottommargin=3.5pt
]{shadedbox}

\AtBeginEnvironment{theorem}{\begin{shadedbox}}
	\AtEndEnvironment{theorem}{\end{shadedbox}}

\AtBeginEnvironment{lemma}{\begin{shadedbox}}
	\AtEndEnvironment{lemma}{\end{shadedbox}}

\AtBeginEnvironment{proposition}{\begin{shadedbox}}
	\AtEndEnvironment{proposition}{\end{shadedbox}}

\AtBeginEnvironment{corollary}{\begin{shadedbox}}
	\AtEndEnvironment{corollary}{\end{shadedbox}}

\newcommand{\papertitle}{A Free Lunch: Gradient Averaging Achieves Optimal Statistical Efficiency in Newton Sketch}
\title{A Free Lunch: Gradient \mbox{Averaging}~Achieves~Optimal Statistical Efficiency in Newton Sketch}

\author{%
\makebox[\dimexpr\textwidth-2\tabcolsep\relax][c]{%
Haoxuan Wang\thanks{School of Industrial and Systems Engineering, Georgia Institute of Technology.}%
\hspace{1.2cm}%
Jianliang Ye\thanks{School of Mathematics, Georgia Institute of Technology.}%
\hspace{1.2cm}%
Sen Na\footnotemark[1]%
}}

\iclrfinalcopy 
\begin{document}

\maketitle
\lhead{}
\renewcommand{\headrulewidth}{0pt}

\begin{abstract}
We study statistical inference for online sketched Newton methods, in which each Newton system is approximately solved by an inner randomized sketching solver. Existing work reveals a computational--statistical trade-off: the sketching solver (and its accelerated variants) reduces the per-step cost of Newton methods from $O(d^3)$ to $O(d^2)$, but the randomness introduced by sketching also inflates the limiting covariance of both the last and the averaged Newton iterates. In this work, we address this trade-off via gradient averaging. We show that simply replacing the stochastic gradient with a moving average at each step can mitigate, and in some cases eliminate, the trade-off in Newton sketch. Importantly, gradient averaging requires no additional oracle evaluations and leaves the contraction bound of the inner sketching solver unchanged.
In particular, we prove that the last sketched Newton iterate with gradient averaging attains a limiting covariance no larger than that without gradient averaging, while the averaged iterate even attains the \textit{minimax optimal} limiting covariance, matching that of the exact Newton method. To our knowledge, this is the first analysis showing that randomness introduced to improve computational efficiency need not compromise statistical efficiency as long as proper gradient estimates are used. The key analytical challenge is that gradient averaging destroys the crucial martingale difference structure underlying central limit theorem arguments.
We overcome this challenge by analyzing the coupled two-timescale recursions governing the iterate and the gradient estimator. Numerical experiments on linear and logistic regression support our theoretical \mbox{covariance}~comparisons.
\end{abstract}

\section{Introduction}\label{sec:intro}

We consider the stochastic optimization problem:
\begin{equation}\label{problem}
\min_{\bx\in\mathbb{R}^d} f(\bx)=\mE_{\xi\sim\mathcal P}[F(\bx;\xi)],
\end{equation}
where $f:\mathbb{R}^d\rightarrow\mathbb{R}$ is a stochastic objective and $F(\cdot;\xi):\mathbb{R}^d\rightarrow\mathbb{R}$ is its realization with $\xi\sim\mathcal P$. Problem~\eqref{problem} is one of the most fundamental building blocks for many decision-making tasks in statistics and machine learning, such as online recommendation~\citep{Li2010contextual}, precision medicine~\citep{Kosorok2019Precision}, and portfolio allocation~\citep{Fan2012Vast}. In these applications,~\eqref{problem} is interpreted as a parameter estimation problem, where $\bx$ denotes the model parameter, $\xi$ denotes an observation, and the minimizer $\bx^\star=\argmin_{\bx\in\mathbb R^d}f(\bx)$ denotes the true model parameter.

The ubiquity of streaming data has made online methods increasingly attractive, where only one \textit{single} sample is processed at each step. Stochastic gradient descent (SGD) and its many acceleration variates are prevalent examples~\citep{Robbins1951Stochastic}, with extensively studied convergence~properties \citep{Pelletier1998almost,Moulines2011Non}. {\myred Although first-order methods enjoy inexpensive updates}, their performance can be sensitive to ill-conditioning and heterogeneous gradient noise.
{\myred Moreover, online inference with SGD still requires updating a proper covariance matrix, incurring $O(d^2)$ time and memory complexity. The resulting confidence intervals can also exhibit substantial undercoverage even in relatively low-dimensional regression problems~\citep{Zhu2021Online}.}
Online Newton methods address these issues by preconditioning each stochastic gradient with curvature information \citep{Bottou2018Optimization}:
\begin{equation}\label{equ:newton_update}
\bx_{t+1}=\bx_t+\varphi_t\Delta\bx_t, \qquad \text{with} \qquad B_t\Delta\bx_t=-\nabla F(\bx_t;\xi_t),
\end{equation}
where $B_t$ approximates the Hessian $\nabla^2 f(\bx_t)$.

Beyond convergence, reliable decision-making also requires uncertainty quantification for the resulting estimator. A standard approach for SGD is Polyak--Ruppert averaging, which averages the iterates to yield an asymptotically normal estimator~\citep{Ruppert1988Efficient,Polyak1992Acceleration}. Building on prior works, the asymptotic properties of the last and averaged iterates of online Newton methods have also been established in unconstrained and constrained settings~\citep{Leluc2023Asymptotic,Gao2025Online}. In particular, under suitable regularity conditions and $B_t\rightarrow B^\star = \nabla^2f(\bx^\star)$, we have$\quad$
\begin{equation}\label{nsequ:1}
1/\sqrt{\varphi_t}\cdot(\bx_t-\bx^\star) \stackrel{d}{\rightarrow} \mN(\0,\; \iota\cdot\Omega)\qquad \text{ and }\qquad \sqrt{t}\rbr{\bar{\bx}_t - \bx^\star}\stackrel{d}{\rightarrow} \mN(\0,\; \Omega),
\end{equation}
where $\bar{\bx}_t \coloneqq \frac{1}{t} \sum_{i=0}^{t-1}\bx_i $, $\Omega = (B^\star)^{-1}\Cov(\nabla F(\bx^\star;\xi))(B^\star)^{-1}$ (see \eqref{eq:single-limiting-covariances}), and $\iota = C_\varphi / (2C_\varphi-1)$ if $\varphi_t=C_\varphi / (t+1)^\varphi$ with $\varphi=1$ while $\iota=0.5$ if $\varphi\in(0.5,1)$. The limiting covariance $\Omega$ is further known to be \textit{asymptotic minimax optimal} in H\'ajek and Le Cam sense \citep{Duchi2021Asymptotic, Davis2024Asymptotic}.
{\myred Though averaged SGD also attains $\Omega$~\citep{Ruppert1988Efficient,Polyak1992Acceleration}, the last iterate of SGD is not asymptotically optimal and its limiting covariance solves a Lyapunov equation involving the Hessian. For online Newton, preconditioning gives the scalar form $\iota\Omega$ in~\eqref{nsequ:1}. In particular, when $\varphi_t=1/(t+1)$, the last iterate attains $\Omega$, so inference can be based directly on it without iterate averaging~\citep{Bercu2020Efficient,Leluc2023Asymptotic}. Appendix~\ref{app:additional-discussion} compares the rate orders and last-iterate stepsize conditions of SGD and online Newton. Online inference also requires covariance estimation. For averaged SGD, the batch-means estimator studied in~\citet{Zhu2021Online} has rate $O(1/\sqrt[4]{t\varphi_t})$. For online Newton methods, the weighted sample covariance estimators studied in~\citet{Kuang2025Online,Wang2026Inference} achieve $O(1/\sqrt{t\varphi_t})$, a quadratic improvement over the batch-means rate.}
Despite promising statistical properties, a key computational bottleneck of online Newton methods is solving the Newton system in \eqref{equ:newton_update}, which requires $O(d^3)$ flops at each~step.$\quad$

Online sketched Newton methods approximately solve the Newton system using an inner randomized iterative solver that performs a fixed number of sketch-and-project updates, reducing the per-step cost to $O(d^2)$~\citep{Na2025Statistical}. A recent Nesterov-accelerated variant further improves the decay rate of the solver's approximation error while preserving the same order of computational cost~\citep{Derezinski2025Fine}. Despite these computational gains, the sketching solver degrades the favorable statistical efficiency of the outer Newton iterates. In particular, \citet{Wang2026Inference} and~\citet{Du2026Inference} established asymptotic normality analogous to \eqref{nsequ:1} for the last and averaged sketched Newton iterates, with limiting covariances $\Sigma_{\bg,\mathrm{as}}^{l}$ and $\Sigma_{\bg,\mathrm{as}}^{a}$, respectively. Their results reveal an interesting computational--statistical trade-off: randomized sketching avoids the exact Newton solve but introduces an additional covariance term that inflates the limiting covariance. Specifically, the~last~and averaged iterates have limiting covariances satisfying $\Omega\preceq \Sigma_{\bg,\mathrm{as}}^{a}\preceq\Sigma_{\bg,\mathrm{as}}^{l}$. Therefore, we~ask:

\emph{Can sketching solver reduce the cost of online Newton without inflating the limiting covariance?}

In this paper, we provide an affirmative answer to this question by replacing the single-sample gradient $\bg_t\coloneqq\nabla F(\bx_t;\xi_t)$ with its moving average $\bbm_t=(1-\eta_t)\bbm_{t-1}+\eta_t\bg_t$. Gradient averaging has long been used to smooth stochastic estimates~\citep{Ruszczynski1983GradientAveraging}, and recent work studies its interaction with Polyak--Ruppert averaging in first-order methods~\citep{Tang2023Acceleration}. Here, however, we use gradient averaging for quite a different purpose: to weaken the interaction between sampling noise and the randomness introduced by the sketching solver in Newton methods. 
We establish convergence and asymptotic normality of both the last and averaged sketched Newton iterates with gradient averaging. In particular, {\myred under weaker assumptions than~\cite{Wang2026Inference, Du2026Inference}}, we show that \textbf{(i)} the last-iterate covariance satisfies $\Sigma_{\bbm,\mathrm{as}}^{l}\preceq\Sigma_{\bg,\mathrm{as}}^{l}$, {\myred where $\Sigma_{\bbm,\mathrm{as}}^{l}$ and $\Sigma_{\bg,\mathrm{as}}^{l}$ denote the last-iterate covariances of accelerated sketched Newton using moving-average and single-sample gradients, respectively}; and \textbf{(ii)} the averaged-iterate covariance satisfies $\Sigma_{\bbm,\mathrm{as}}^{a}=\Omega$. Thus, gradient averaging mitigates the computational--statistical trade-off for the last iterate, and {\myred even eliminates it for the averaged iterate. 
In this sense, gradient averaging provides a ``free lunch'' compared to existing sketched Newton methods: our covariance gains require only an $O(d)$ vector update~per~iteration, without additional gradient, Hessian, or sketching evaluations, and hold under weaker assumptions.}

The analysis requires addressing the following \textbf{challenges} introduced by gradient averaging.

\textbf{(a)} $\bbm_t$ averages stochastic gradients evaluated at different iterates and is therefore not conditionally unbiased for $\nabla f(\bx_t)$, unlike $\bg_t$, whose centered noise forms crucial martingale difference structure underlying central limit theorem arguments. More precisely, the tracking error $\be_t \coloneqq \bbm_t-\nabla f(\bx_t)$ contains inherited error from earlier steps, gradient drift, and martingale noise, and must be analyzed jointly with the optimization error $\bDelta_t\coloneqq\bx_t-\bx^\star$. As such, we derive coupled recursions for $(\be_t,\bDelta_t)$ and control their evolution on the two timescales governed by $\eta_t$ and $\varphi_t$.

\textbf{(b)} To characterize how gradient averaging changes the leading sketching noise, we must separate sampling noise, sketching fluctuations, and variations in the Hessian approximation and sketching parameters across iterations. Since the resulting perturbation terms are sufficiently small only when $(\bDelta_t, B_t-B^\star, \be_t)$ remain in a local neighborhood of zero, we introduce a stopping time at the first exit from this neighborhood and localize the convergence rate analysis (see \eqref{eq:local-stopping-time} in the proof of Lemma \ref{lem:local-rate}). The effect of the localization is then removed by our almost-sure convergence guarantee.

\textbf{(c)} To isolate $\nabla F(\bx^\star;\xi_t)$ as the leading stochastic noise in the asymptotic linear representations~established in Theorem~\ref{thm:sketch-noise-reduction}, we rewrite the outer Newton recursion in terms of the stochastic gradient $\bg_t$. This introduces weighted differences $\bbm_t-\bbm_{t-1}$ that do not appear in the analysis of vanilla online Newton methods. We use summation by parts to transfer the differences from the moving-average gradients to their weights, and then bound the resulting boundary and weighted-sum terms to~show that they are asymptotically negligible under both the last- and averaged-iterate normalizations.

Our main \textbf{contributions} are summarized as follows.

\textbf{(a)} We establish global almost-sure convergence of online sketched Newton with gradient averaging. Under standard two-timescale conditions, we show that the iterate $\bx_t$ converges to $\bx^\star$ and the moving-average gradient $\bbm_t$ tracks $\nabla f(\bx_t)$ almost surely (Theorem~\ref{thm:global-convergence}). {\myred Thus, using gradients evaluated at earlier iterates does not prevent the outer iterates from converging.} Moreover, gradient averaging requires no additional oracle evaluations and preserves the inner-solver contraction bound.

\textbf{(b)} We establish last-iterate asymptotic normality, $1/\sqrt{\varphi_t}\cdot(\bx_t-\bx^\star)\xrightarrow{d}\mN(\zero,\Sigma_{\bbm,\mathrm{as}}^{l})$ (Theorem~\ref{thm:last-normality}), and prove $\Sigma_{\bbm,\mathrm{as}}^{l}\preceq\Sigma_{\bg,\mathrm{as}}^{l}$ (Prop.~\ref{prop:covariance-comparison}). Thus, gradient averaging yields no larger asymptotic variance in any direction than vanilla online sketched Newton. 
{\myred This covariance reduction (same for (c) below) uses the \textbf{same} inner sketching solver and the \textbf{same} number of sketching steps as existing~methods using single-sample gradients.}
Our local convergence rates and asymptotic linear representations reveal the mechanism behind this variance reduction: the random matrix $I-\tK_i^\star$, which captures the limiting effect of the inner solver on gradient noise, is replaced by its mean $I-K^\star$ (Theorem~\ref{thm:sketch-noise-reduction}).~Consequently, sketching enters the limiting covariance only through its mean effect.

\textbf{(c)} We further establish averaged-iterate asymptotic normality, $\sqrt{t}\,(\bar{\bx}_t-\bx^\star)\xrightarrow{d}\mN(\zero,\Sigma_{\bbm,\mathrm{as}}^{a})$, with $\Sigma_{\bbm,\mathrm{as}}^{a} = \Omega\preceq\Sigma_{\bg,\mathrm{as}}^{a}$ (Theorem~\ref{thm:averaged-normality}). Polyak--Ruppert averaging eliminates the remaining mean effect of sketching and recovers the covariance $\Omega$ of exact online Newton method, which is asymptotically \textit{minimax optimal} \citep{Duchi2021Asymptotic}.
{\myred Therefore, with a fixed number of sketching steps per iteration, our method gains computational efficiency without sacrificing statistical efficiency: its~averaged iterate has the same limiting covariance as if each Newton system were solved exactly.}

\noindent \textbf{Notation.}
We use $\|\cdot\|$ to denote the Euclidean norm for vectors and the spectral norm for matrices. We write $O(\cdot)$ and $o(\cdot)$ ($O_p(\cdot)$ and $o_p(\cdot)$) for the usual big- and small-$o$ notation (in probability). We use $I$ and $\zero$ to denote identity matrices and zero vectors/matrices, with dimensions clear from context. For symmetric $A$ and $B$, $A\succeq B$ means $A-B$ is positive semidefinite. We abbreviate $\nabla f_t=\nabla f(\bx_t)$ and $\nabla^2 f_t = \nabla^2f(\bx_t)$. 
We write $\Sigma_{u,s}^{q}$ for limiting covariances, where $u\in\{\bg,\bbm\}$ denotes the stochastic gradient or its moving average; $s\in\{\mathrm{e},\mathrm{as}\}$ denotes the solver (exact or accelerated sketching); and $q\in\{l,a\}$ denotes the iterate (last or averaged). In particular, we know from \eqref{nsequ:1} that $\Omega=\Sigma_{\bg,\mathrm{e}}^{a}$.$\quad$

\section{Preliminaries}\label{sec:preliminaries}

We review online sketched Newton methods, which replace the exact Newton solve with an inexact randomized sketching solver \citep{Na2025Statistical,Kuang2025Online}. These methods were subsequently improved by leveraging Nesterov's acceleration \citep{Wang2026Inference,Du2026Inference}. In what follows, we first introduce the vanilla online Newton, then describe the accelerated sketching solver (of which the unaccelerated solver is a special case), and finally present the preliminary~guarantees.

\noindent$\bullet$ \textbf{Online Newton with {\myred a Hessian approximation.}}
Online Newton methods precondition stochastic gradients using curvature information. Given $\bx_t$, we form {\myred a positive definite matrix $B_t$ from past information to approximate $\nabla^2 f_t$. We then draw $\xi_t\sim\cP$ independently of the past and compute $\bg_t\coloneqq\nabla F(\bx_t;\xi_t)$ and $H_t\coloneqq\nabla^2F(\bx_t;\xi_t)$. A common starting point for constructing $B_t$ is the Hessian average $\tilde B_t\coloneqq\frac{1}{t}\sum_{i=0}^{t-1}H_i$, which stabilizes the curvature estimate and improves local convergence behavior~\citep{Na2022Hessian}. Section~\ref{sec:normality} uses a clipped version of this average to form $B_t$.}

Given a prescribed stepsize sequence $\{\varphi_t\}_{t\geq0}$, the vanilla online Newton method updates as
\begin{equation}\label{eq:newton-system}
\bx_{t+1}=\bx_t+\varphi_t\Delta\bx_t, \qquad \text{with} \qquad B_t \Delta \bx_t = -\bg_t.
\end{equation}
Solving this Newton system costs $O(d^3)$ flops per step, motivating the sketching solver next.

\noindent$\bullet$ \textbf{Nesterov's accelerated sketch-and-project.}
The inner solver for~\eqref{eq:newton-system} combines randomized sketch-and-project updates with Nesterov's acceleration, yielding faster decay rate of the approximation error at the same order of per-step cost~\citep{Gower2018Accelerated,Derezinski2025Fine}. We suppress the outer index $t$ and present the accelerated solver $\texttt{NASketch}(B,-\bg;\alpha,\beta,\gamma,\tau)$ in Algorithm~\ref{alg:nasketch}, where $(\alpha,\beta,\gamma)$ are the acceleration parameters and $\tau$ is the number of sketching steps.

\vspace{-0.1cm}
At each outer iteration, $(B,\bg)$ are held fixed, and we draw $S_j\stackrel{\mathrm{iid}}{\sim}S$ for $0\leq j\leq\tau-1$. Without acceleration, the sketch-and-project solver performs the update $\bz_{j+1}=\bz_j-BS_j(S_j^\top B^2S_j)^\dagger S_j^\top(B\bz_j+\bg)$, where $\dagger$ denotes the pseudoinverse. Nesterov's acceleration introduces an auxiliary momentum sequence $\bv_j$. Given $(\bz_j,\bv_j)$, Algorithm~\ref{alg:nasketch} first forms the middle point $\by_j=\alpha\bv_j+(1-\alpha)\bz_j$, computes a sketching correction at $\by_j$, and updates $(\bz_j,\bv_j)$ jointly. By Algorithm \ref{alg:nasketch}, the setup $(\alpha,\beta,\gamma)=(0.5,0,1)$ leads to $\bv_j=\bz_j$ and reduces the method to unaccelerated sketch-and-project.

\vspace{0.1cm}

\begingroup
\setlength{\intextsep}{2pt}
\begin{algorithm}[thb]
\caption{Sketch-and-Project with Nesterov's Acceleration:
$\texttt{NASketch}(B,-\bg;\alpha,\beta,\gamma,\tau)$}\label{alg:nasketch}
\begin{algorithmic}[1]
\linespread{0.88}\selectfont
\setlength{\lineskip}{1.5pt}
\STATE \textbf{Initialize:} $\bz_0=\bv_0=\zero$, $\alpha\in(0,1)$, $\beta\in[0,1)$, $\gamma\geq1$, $\tau\geq 1$.
\FOR{$j=0,1,\ldots,\tau-1$}
\STATE $\by_j\leftarrow\alpha\bv_j+(1-\alpha)\bz_j$ and draw a sketching matrix $S_j\sim S$, where $S_j\in\mR^{d\times s}$, $1\leq s\leq d$.\vskip-0.16cm
\STATE $\bw_j\leftarrow BS_j(S_j^\top B^2S_j)^\dagger S_j^\top(B\by_j+\bg)$.
\STATE $(\bz_{j+1},\bv_{j+1})\leftarrow(\by_j-\bw_j,\beta\bv_j+(1-\beta)\by_j-\gamma\bw_j)$.
\ENDFOR    
\STATE \textbf{Output:} $\bz_\tau$. 
\end{algorithmic}
\end{algorithm}
\endgroup

\vspace{5pt}

Following \citet{Gower2018Accelerated,Derezinski2025Fine}, we parameterize $(\alpha,\beta,\gamma)$ by $(\mu,\nu)$ as
\begingroup
\setlength{\abovedisplayskip}{4pt}
\setlength{\belowdisplayskip}{4pt}
\begin{equation}\label{eq:acceleration-parameters}
\alpha=1/(1+\gamma\nu), \qquad \beta=1-\sqrt{\mu/\nu}, \qquad \gamma=1/\sqrt{\mu \nu}.
\end{equation}
\endgroup
For a generic sketching matrix $S$, define $\tilde{Z}\coloneqq BS(S^\top B^2S)^\dagger S^\top B$ and $Z\coloneqq\mE[\tilde{Z}]$, where the expectations below are taken over $S$. When $Z\succ\zero$, define \vspace{-0.1cm}
\begin{equation}\label{eq:mu-nu}
\mu\coloneqq\inf_{\ba\neq\zero}\frac{\ba^\top Z\ba}{\ba^\top\ba},
\qquad\qquad
\nu\coloneqq\sup_{\ba\neq\zero}\frac{\ba^\top\mE[\tilde{Z}Z^{-1}\tilde{Z}]\ba}{\ba^\top Z\ba}.
\end{equation}
At outer iteration $t$, evaluating \eqref{eq:mu-nu} at $B_t$ gives $(\mu_t,\nu_t)$ and hence $(\alpha_t,\beta_t,\gamma_t)$. 
Since the exact computation of these parameters requires expectations over the sketching distribution, \eqref{eq:mu-nu} primarily serves as a theoretical parameterization for analyzing the solver's convergence; 
implementations may approximate these quantities using periodically updated estimates or fixed parameter choices \citep{Wang2026Inference}. With specified parameters, the online sketched Newton method replaces the exact Newton direction $\Delta\bx_t$ in \eqref{eq:newton-system} with $\bz_{t,\tau}$:
\begin{equation}\label{nsequ:2}
\bz_{t,\tau}=\texttt{NASketch}\!\left(B_t,-\bg_t;\alpha_t,\beta_t,\gamma_t,\tau\right), \qquad \bx_{t+1}=\bx_t+\varphi_t\bz_{t,\tau}.
\end{equation}
\noindent $\bullet$ \textbf{Theory of accelerated sketched Newton.}
We next review the inner-solver contraction bound and existing asymptotic normality results for online sketched Newton methods. At any fixed outer iteration $t$, each inner update of the solution approximation in Algorithm \ref{alg:nasketch} is linear in $\bg_t$. More specifically, by \citet[Lemma 3.5]{Wang2026Inference}, there exists a realized coefficient factor matrix $\tK_t$ for the solution approximation error, independent of $\bg_t$, such that (see \eqref{def:tilde_K_t} for the precise form of $\tK_t$)
\begingroup
\setlength{\abovedisplayskip}{4pt}
\setlength{\belowdisplayskip}{4pt}
\begin{equation}\label{eq:sketching-residual-matrix}
\bz_{t,\tau}-\Delta\bx_t=\tK_t(\bz_{t,0}-\Delta\bx_t),
\qquad\qquad
\Delta\bx_t=-B_t^{-1}\bg_t.
\end{equation}
\endgroup
With the above identity, the accelerated inner solver satisfies the following contraction bound.

\AddToHookNext{env/mdframed/begin}{\mdfsetup{nobreak=true}}
\begin{lemma}[\cite{Derezinski2025Fine}]\label{lem:accelerated-inner-rate}
Suppose $Z_t = \mE[\tilde{Z}_t] = \mE[B_tS(S^\top B_t^2S)^\dagger S^\top B_t]\succ\zero$,~then
\begin{equation*}
\mE[\|\bz_{t,j}-\Delta\bx_t\|^2]\leq2(1-\sqrt{\mu_t/\nu_t})^j\|\bz_{t,0}-\Delta\bx_t\|^2,\qquad \text{for any } t\geq0, \; j\geq 0,
\end{equation*}
where the expectation is taken over the inner sketches $\{S_j\}_j$.
\end{lemma}

By \citet[Lemma 2]{Gower2018Accelerated}, $1\leq\nu_t\leq1/\mu_t$; thus, $1-\sqrt{\mu_t/\nu_t}\leq1-\mu_t$, where the right-hand side is the contraction rate of vanilla unaccelerated sketch-and-project \mbox{\citep{Gower2015Randomized}}.

\vskip-0.1cm

To introduce asymptotic normality, we need additional notation. Let $B^\star\coloneqq\nabla^2f(\bx^\star)$ be the Hessian at $\tx$ and $(\mu^\star,\nu^\star,\alpha^\star,\beta^\star,\gamma^\star)$ be parameters obtained from~\eqref{eq:acceleration-parameters} and \eqref{eq:mu-nu} with $B=B^\star$. We also define $\tK^\star$ by evaluating~\eqref{eq:sketching-residual-matrix} at these limiting quantities (cf. \eqref{def:tilde_K_t}), $K^\star\coloneqq\mE[\tK^\star]$, and two matrices
\begin{equation}\label{eq:single-limiting-covariances}
\Omega=(B^\star)^{-1}\mE [\nabla F(\bx^\star;\xi)\nabla F(\bx^\star;\xi)^\top ](B^\star)^{-1}\quad \text{and}\quad
\Gamma_{\bg,\mathrm{as}}\coloneqq\mE [(I-\tK^\star)\Omega(I-\tK^\star)^\top ].
\end{equation}
Let $\varphi_t\coloneqq C_\varphi / (t+1)^\varphi$, $\zeta_\varphi\coloneqq\1_{\{\varphi=1\}}/(2C_\varphi)$, and $A_\varphi^\star\coloneqq(I-K^\star)-\zeta_\varphi I$. 
The following theorem {\myred extends the existing normality results} for both the last and the averaged iterates.

\begin{theorem}[\cite{Wang2026Inference, Du2026Inference}]\label{thm:single-normality}
\setlength{\abovedisplayskip}{3pt}
\setlength{\abovedisplayshortskip}{3pt}
\setlength{\belowdisplayskip}{3pt}
\setlength{\belowdisplayshortskip}{3pt}
Under the assumptions in Theorem~\ref{thm:sketch-noise-reduction}, we~have
\begin{equation}\label{eq:single-last-normality}
1/\sqrt{\varphi_t}\cdot(\bx_t-\bx^\star)\xrightarrow{d}\mN(\zero,\Sigma_{\bg,\mathrm{as}}^{l}),
\qquad\text{where } \quad
A_\varphi^\star\Sigma_{\bg,\mathrm{as}}^{l}+\Sigma_{\bg,\mathrm{as}}^{l}A_\varphi^\star=\Gamma_{\bg,\mathrm{as}}.
\end{equation}
Moreover, if $\varphi\in(0.5,1)$, then the averaged iterate $\bar{\bx}_t=\sum_{i=0}^{t-1}\bx_i/t$ satisfies
\begin{equation}\label{eq:single-averaged-normality}
\sqrt{t}\,(\bar{\bx}_t-\bx^\star)\xrightarrow{d}\mN(\zero,\Sigma_{\bg,\mathrm{as}}^{a}),
\qquad\text{where}\quad
\Sigma_{\bg,\mathrm{as}}^{a}\coloneqq(I-K^\star)^{-1}\Gamma_{\bg,\mathrm{as}}(I-K^\star)^{-1}.
\end{equation}
\end{theorem}

{\myred We clarify that the results of \citet{Wang2026Inference,Du2026Inference} were established under uniform bounds on sampled Hessian $\nabla^2F(\cdot;\xi)$. For clipped $B_t$ mentioned in Section~\ref{sec:normality}, our global convergence and local rate arguments extend to their update that uses $\bg_t$ and provide the bounds needed in their proofs. Their asymptotic normality arguments then yield~\eqref{eq:single-last-normality} and~\eqref{eq:single-averaged-normality} under the conditions stated~here.
}

For the last iterate, setting $\tK^\star=K^\star=\zero$ in~\eqref{eq:single-last-normality} recovers the asymptotic covariance of exact online Newton~\eqref{nsequ:1}. In general, however, when $\varphi=C_\varphi=1$, it has been shown that $\Sigma_{\bg,\mathrm{as}}^{l}\succeq\Omega$. Since $\Omega$ is the classical Polyak--Ruppert covariance and the local asymptotic minimax bound~\citep{Duchi2021Asymptotic}, this ordering reveals a computational--statistical trade-off: randomized sketching reduces the cost of the Newton solve but inflates the limiting covariance.

Analogously, the averaged-iterate result \eqref{eq:single-averaged-normality} also recovers exact online Newton \eqref{nsequ:1}. However, it has also been shown that $\Sigma_{\bg,\mathrm{as}}^{l}\succeq\Sigma_{\bg,\mathrm{as}}^{a}\succeq\Omega$. Thus, while iterate averaging reduces limiting covariance relative to the last iterate, it does not fully eliminate the sketch-induced covariance inflation, and the trade-off persists. In this paper, we completely eliminate the trade-off by gradient averaging.

\section{Global Convergence with Gradient Averaging}\label{sec:convergence}

In this section, we incorporate a moving-average gradient into accelerated sketched Newton method, introduce assumptions, and establish almost-sure convergence of the method. Specifically, instead of using $\bg_t$ directly, we initialize $\bbm_0=\bg_0$ and update \vspace{-0.05cm}
\begin{equation}\label{eq:debiased-gradient}
\bbm_t = (1 - \eta_t) \bbm_{t-1} + \eta_t \bg_t, \qquad \eta_t \in(0, 1], \quad t \geq 1,
\end{equation}
where $\{\eta_t\}_{t\geq1}$ is a prescribed averaging sequence. We then replace $\bg_t$ in \eqref{nsequ:2} with $\bbm_t$ and obtain
\begin{equation}\label{eq:outer-update}
\bz_{t,\tau}=\texttt{NASketch}\!\left(B_t,-\bbm_t;\alpha_t,\beta_t,\gamma_t,\tau\right),
\qquad
\bx_{t+1}=\bx_t+\varphi_t\bz_{t,\tau}.
\end{equation}
This replacement changes only the right-hand side of~\eqref{eq:newton-system}; hence, $B_t$, $(\alpha_t,\beta_t,\gamma_t)$, and the inner sketching procedure remain unchanged, as does the contraction bound in Lemma~\ref{lem:accelerated-inner-rate}. Moreover, updating $\bbm_t$ requires no additional oracle evaluations and incurs only $O(d)$ computation, preserving the per-step cost order. Since $\bbm_t$ averages gradients evaluated at different iterates, it is not conditionally unbiased for $\nabla f_t$ and destroys the crucial martingale difference structure. Thus, establishing global convergence requires joint control of the optimization error $\bDelta_t=\bx_t-\bx^\star$ and the gradient-tracking error $\be_t=\bbm_t-\nabla f_t$. We first specify the filtration and assumptions used here and in~Section~\ref{sec:normality}.

For $t\geq 0$, we define $\sigma$-algebras $\mathcal{F}_{t-0.5} = \sigma\left(\{\xi_i, \{S_{i, j}\}_{j}\}_{i=0}^{t-1}\cup \xi_t\right)$ and $\mathcal{F}_t = \sigma\left(\{\xi_i, \{S_{i, j}\}_{j} \}_{i=0}^{t}\right)$. For consistency, let $(\bx_0,B_0)$ be deterministic and $\mF_{-1}\coloneqq\sigma(\bx_0,B_0)$ be the trivial $\sigma$-algebra. Then, we know $(\bg_t,H_t,\bbm_t)$ and $(\bz_{t,\tau},\bx_{t+1},B_{t+1})$ are $\mF_{t-0.5}$- and $\mF_t$-measurable, respectively. 

The first assumption regards the strong convexity and Lipschitz smoothness of the objective.$\quad\qquad$

\begin{assumption}\label{ass:1}
{\myred We assume $f$ and its realization $F(\cdot;\xi)$ are twice continuously differentiable with $\gamma_H I \preceq \nabla^2 f(\bx) \preceq \Upsilon_H I$ for any $\bx$ and some constants $\gamma_H, \Upsilon_H > 0$, and $\nabla^2 f$ is $\Upsilon_L$-Lipschitz~continuous. We further assume there exist two positive sequences $\{\gamma_{B,t}, \Upsilon_{B,t}\}_{t \geq 0}$ such that $B_t$, $\gamma_{B,t}$ and $\Upsilon_{B,t}$ are $\mathcal F_{t-1}$-measurable and the following hold almost surely: $\gamma_{B,t} I \preceq B_t \preceq \Upsilon_{B,t} I$ for~all~$t \geq 0$; $\sum_{t=0}^{\infty} \varphi_t / \gamma_{B,t} = \infty$ and $\sum_{t=0}^{\infty} (\varphi_t / \gamma_{B,t})^2 < \infty$; and $\limsup_{t \rightarrow \infty} \Upsilon_{B,t} / \gamma_{B,t} < \infty$.}
\end{assumption}

We next impose growth conditions on the moments of the gradient and Hessian noises. 

\begin{assumption}\label{ass:2}
For any $t\geq 0$, we assume the gradient estimate is unbiased $\mE[\bg_t\mid \mathcal{F}_{t-1}]=\nabla f_t$ and satisfies a~growth condition on its $q_g$-th moment: for some constants $C_{g,1}, C_{g,2} > 0$,
\begin{equation*}
\mE[\|\bg_t-\nabla f_t\|^{q_g}\mid\mF_{t-1}]\leq C_{g,1}\|\bx_t-\bx^\star\|^{q_g}+C_{g,2}.
\end{equation*}
\end{assumption}

\begin{assumption}\label{ass:3}
For any $t\geq 0$, we assume the Hessian estimate is unbiased $\mE[H_t \hskip-1pt\mid\hskip-1pt \mathcal{F}_{t-1}]\hskip-1pt=\hskip-1pt\nabla^2 f_t$ and satisfies a growth condition on its second moment: for some constants $C_{H,1}, C_{H,2} > 0$,
\begin{equation*}
\mE[\|H_t-\nabla^2f_t\|^2\mid\mF_{t-1}] \leq C_{H,1}\|\bx_t-\bx^\star\|^2+C_{H,2}.
\end{equation*}
\end{assumption}

\vspace{-1pt}
We will specify $q_g$ in the theorem statements later. In particular, $q_g=2$ suffices for global convergence and $q_g>2$ for asymptotic normality; Assumption~\ref{ass:3} will be used only for asymptotic normality~in Section~\ref{sec:normality}. Finally, we present an assumption on the sketching distribution.

\begin{assumption}\label{ass:4}
For any $t \geq 0$, we assume there exists $\gamma_S > 0$ such that $\{S_{t,j}\}_{j=0}^{\tau-1}\stackrel{iid}{\sim}S$ satisfying $Z_t \coloneqq \mathbb{E}[B_t S(S^{\top}B_t^2 S)^{\dagger}S^{\top}B_t \mid \mathcal{F}_{t-1}] \succeq \gamma_S I$, and $\{S_{t,j}\}_{j=0}^{\tau-1}$ are independent of $\xi_t$.
Moreover, the condition number of $S$ has bounded expectation: $\mE[\|S\|\|S^\dagger\|] \leq \Upsilon_S$ for $\Upsilon_S>0$.
\end{assumption}

\vspace{-1pt}

{\myred Assumptions \ref{ass:1}--\ref{ass:4} are taken (in fact, weaken) from existing literature on sketched Newton inference \citep{Kuang2025Online, Wang2026Inference, Du2026Inference}, and the population strong convexity and smoothness conditions are also used in SGD inference \citep{Chen2020Statistical, Zhu2021Online}.~In~particular, Assumption \ref{ass:1} has two parts.
\textbf{(i)} The strong convexity and smoothness are imposed only on the population objective $f$, and no spectral bounds are required on sampled Hessians $\nabla^2F$. We note that global lower bound on $\nabla^2 f$ is used only to establish $\bx_t\stackrel{a.s.}{\rightarrow}\bx^\star$ in Theorem \ref{thm:global-convergence}, while inference is local in nature: the curvature entering $\Omega$ in \eqref{eq:single-limiting-covariances} is the Hessian at the solution, $B^\star=\nabla^2 f(\bx^\star)$.~\textbf{(ii)}~The conditions on the Hessian approximation $B_t$ are taken directly from conditioned SGD \citep[Assumption~9]{Leluc2023Asymptotic}. It is strictly weaker than its counterparts in sketched Newton inference \citep{Kuang2025Online, Wang2026Inference, Du2026Inference}, which impose uniform bounds on $\nabla^2F$. In linear regression with covariates $\ba\sim\mN(\b0,\Sigma_a)$, the sample Hessian $\ba\ba^\top$ is rank one and unbounded, so uniform bounds fail while $\nabla^2 f(\bx)=\Sigma_a$ satisfies Assumption \ref{ass:1}. Assumptions \ref{ass:2}--\ref{ass:4} follow sketched Newton literature, and the gradient condition in~Assumption \ref{ass:2} is also standard in SGD inference. Appendix~\ref{app:additional-discussion} further discusses Hessian conditions and their relation to conditioned~SGD.$\quad$
}

\vspace{-0.45cm}
For regression problems, the gradient moment conditions can be verified by the arguments in \citet[Appendix A]{Chen2020Statistical}. The lower bound on expected projection matrix $Z_t$ is required to ensure convergence of sketching solvers~\citep{Gower2018Accelerated}. These conditions hold for various sketching distributions, including Gaussian and Kaczmarz sketching~\citep{Strohmer2008Randomized}.

Under the assumptions shown above, the following theorem establishes joint global almost-sure convergence of the optimization error $\Delta_t$ and tracking error $\be_t$.

\begin{theorem}\label{thm:global-convergence}
Consider the Newton scheme~\eqref{eq:outer-update} with $\varphi_t = C_\varphi / (t+1)^\varphi$, $\eta_t = C_\eta / (t+1)^\eta$ satisfying $C_\varphi > 0$, $C_\eta \in (0, 1]$ and $\eta > 0.5$. Suppose Assumptions~\ref{ass:1},~\ref{ass:2}, and~\ref{ass:4} hold with $q_g = 2$, {\myred $\varphi_t/(\gamma_{B,t} \eta_t) \rightarrow 0$ almost surely}, and $\tau$ satisfies $\tau(1-\sqrt{\mu_t/\nu_t})^{\tau-2}\leq{\myred \gamma_{B,t}}/(8{\myred \Upsilon_{B,t}})$~for~all~$t\geq 0$. Then, we have~$\be_t = \bbm_t - \nabla f_t \stackrel{a.s.}{\rightarrow} \b0$ and $\bx_t \stackrel{a.s.}{\rightarrow} \bx^{\star}$ as $t \rightarrow \infty$.
\end{theorem}

Theorem~\ref{thm:global-convergence} requires only the bounded variance condition on the stochastic gradient, which is standard in stochastic optimization~\citep{Ghadimi2013Stochastic}. The condition {\myred $\varphi_t/(\gamma_{B,t}\eta_t)\rightarrow0$ ensures that $\bbm_t$ adopts an averaging sequence $\eta_t$ that asymptotically dominates the effective outer Newton stepsize $\varphi_t/\gamma_{B,t}$}, so as to preserve as much gradient information from $\bg_t$ as possible despite the continued evolution of $\bx_t$. Moreover, Lemma~\ref{lem:accelerated-inner-rate} and the condition on $\tau$ ensure that the approximate Newton direction is sufficiently accurate for global convergence.
{\myred By \citet[(3.5)]{Wang2026Inference}, $1-\sqrt{\mu_t/\nu_t}\leq1-\gamma_S<1$. Assumption~\ref{ass:1} gives $\sup_{t\geq0}\Upsilon_{B,t}/\gamma_{B,t}<\infty$ almost surely. Hence, along almost every realization, the sketching steps needed to meet the condition on $\tau$ are uniformly bounded in $t$; the required inner accuracy doesn't increase with $t$. See Appendix~\ref{app:additional-discussion} for more on $\tau$.}
Overall, Theorem~\ref{thm:global-convergence} suggests that though $\bbm_t$ is not conditionally unbiased for $\nabla f_t$, it tracks $\nabla f_t$ almost surely and $\bx_t$ also converges to $\bx^\star$.
This global result underlies the local analysis in Section~\ref{sec:normality}.

\section{Optimal Statistical Efficiency via Gradient Averaging}\label{sec:normality}

We now study how moving-average gradient estimators in~\eqref{eq:debiased-gradient} change the statistical effect of sketching. We {\myred first introduce a spectrally clipped Hessian average $B_t$ and then establish} local asymptotic rates and show that the centered sketching fluctuation is asymptotically negligible with gradient averaging. We then derive asymptotic normality of the last and averaged Newton iterates. The resulting limits reveal a sharp distinction (Theorem~\ref{thm:sketch-noise-reduction}): gradient averaging reduces the random sketching factor in the leading stochastic noise to its mean, thereby reducing the sketching-induced fluctuation of the last iterate, while averaging the outer iterates further cancels~the remaining mean factor~and completely eliminates the inflation in asymptotic~uncertainty.

{\myred 
Throughout this section, we consider the following $B_t$ based on Hessian averaging. For $t \geq 1$, let $\;\;$
\begin{equation}\label{eq:clipped-hessian-average}
\vspace{-0.15cm}
\tilde{B}_t \coloneqq \frac{t-1}{t}\tilde{B}_{t-1}+\frac{1}{t}H_{t-1}= \frac{1}{t}\sum_{i=0}^{t-1} H_i, \qquad B_t \coloneqq \operatorname{Clip}(\tilde{B}_t; \gamma_{B,t}, \Upsilon_{B,t}).
\end{equation}
For a symmetric matrix $A=U\operatorname{diag}(\lambda_1,\ldots,\lambda_d)U^\top$ and $0<\gamma\leq\Upsilon$, we define $\operatorname{Clip}(A;\gamma,\Upsilon) \coloneqq U\operatorname{diag}\bigl(\{\min\{\Upsilon,\max\{\gamma,\lambda_j\}\}\}_{j=1}^{d}\bigr)U^\top$. In practice, we refresh the clipping once every $d$ iterations and keep $B_t$, $\gamma_{B,t}$, and $\Upsilon_{B,t}$ fixed between refreshes. A spectral clipping costs $O(d^3)$, while updating $\tilde{B}_t$ costs $O(d^2)$ per iteration. Thus, the amortized matrix-update cost is $O(d^2)$ per iteration. The~clipping in~\eqref{eq:clipped-hessian-average} enforces the spectral bounds on $B_t$ used to control the search direction in the global convergence analysis. To identify the limiting covariance matrices, we also need the clipped Hessian average to approach $B^\star$ as $\bx_t\rightarrow\bx^\star$.
We therefore assume that almost~surely $\gamma_{B,t}\leq\lambda_{\min}(B^\star)$~and $\Upsilon_{B,t}\geq\lambda_{\max}(B^\star)$ for all sufficiently large $t$.}

{\myred 
The sketching cost depends on sketching distribution. For coordinate sketches $S=\be_j$ considered in Section \ref{sec:experiments}, each update in Algorithm~\ref{alg:nasketch} accesses only one column of $B_t$ and costs $O(d)$, so $\tau$ updates cost $O(\tau d)$. Moreover, for these sketches, \eqref{eq:mu-nu} yields $\mu_t\geq1/(d\kappa_t^2)$ and $\nu_t=d$, where $\kappa_t\coloneqq\lambda_{\max}(B_t)/\lambda_{\min}(B_t)\leq\Upsilon_{B,t}/\gamma_{B,t}$ is bounded uniformly in $t$ \citep{Gower2018Accelerated, Derezinski2025Fine, Wang2026Inference}. Thus, the conditions on $\tau$ in Theorems~\ref{thm:global-convergence} and~\ref{thm:sketch-noise-reduction} are met by $\tau$ of order $d$, yielding an $O(d^2)$ sketching solver cost per outer iteration, compared with $O(d^3)$ for an exact Newton solve. 
The constant hidden in this $O(d^2)$ bound depends on the condition number, which is where sketching acceleration helps: it improves the worst-case dependence from $\kappa_t^2$ to $\kappa_t$. Including the $O(d)$ moving average update, the total amortized cost per outer iteration remains $O(d^2)$ in terms of the dimension. See Appendix~\ref{app:additional-discussion} for details. We emphasize that the conditions on $\tau$ are sufficient, worst-case conditions and can be conservative in practice. In our experiment, even $\tau\in\{5,10\}$ suffices to achieve the predicted covariance reduction.
}

\subsection{Local rates and sketching noise reduction}\label{subsec:local-preparation}

We begin by showing the almost-sure convergence of the {\myred clipped} Hessian average $B_t$, the acceleration parameters $(\alpha_t,\beta_t,\gamma_t)$, and, more importantly, the evolution coefficient matrix of the sketching approximation error, $K_t\coloneqq\mE[\tK_t\mid\mF_{t-1}]$, from \eqref{eq:sketching-residual-matrix}. Explicit form of $\tK_t$ is given in \eqref{def:tilde_K_t} and not crucial here.

\begin{lemma}\label{lem:operator-stabilization}
Under the conditions of Theorem~\ref{thm:global-convergence} and Assumption~\ref{ass:3}, we have $B_t \stackrel{a.s.}{\rightarrow} B^{\star}$, $(\alpha_t,\beta_t,\gamma_t) \stackrel{a.s.}{\rightarrow} (\alpha^{\star}, \beta^{\star}, \gamma^{\star})$, $K_t \stackrel{a.s.}{\rightarrow} K^{\star}$; and $K^\star$ is symmetric with $\|K^\star\|\leq1/4$.
\end{lemma}

With the limiting quantities in place, we establish the local rates needed for asymptotic normality.

\begin{lemma}\label{lem:local-rate}
Suppose Assumptions~\ref{ass:1}--\ref{ass:4} hold with $q_g>2$; $\tau$ satisfies $\tau(1-\sqrt{\mu_t/\nu_t})^{\tau-2}\leq{\myred \gamma_{B,t}}/(8{\myred \Upsilon_{B,t}})$ for all $t\geq0$; and the stepsize parameters satisfy $C_\eta\in(0,1]$ and $0.5<\eta<\varphi\leq1$. If $\varphi=1$, we further suppose $C_\varphi>1/\{2\lambda_{\min}(I-K^\star)\}$. Then, we have \begin{equation*}
\max\big\{\|\bbm_t-\nabla f_t\|, \;\|\bx_t-\bx^\star\|,\;\|B_t-B^\star\|,\; \|(\alpha_t,\beta_t,\gamma_t) - (\alpha^\star,\beta^\star,\gamma^\star)\|,\; \|K_t-K^\star\|\big\}=O_p(\sqrt{\eta_t}).
\end{equation*}
\end{lemma}

The rates in Lemma~\ref{lem:local-rate} are obtained via a localization argument. Specifically, we introduce the first exit time in~\eqref{eq:local-stopping-time} to restrict the analysis to trajectories for which the iterate and gradient-tracking errors remain within a sufficiently small neighborhood of zero. This localization enables control of the resulting perturbation terms. The almost-sure convergence established in Theorem~\ref{thm:global-convergence} and Lemma~\ref{lem:operator-stabilization} ensures that the exit probability can be made arbitrarily small, thereby converting the localized bounds into the stated $O_p(\sqrt{\eta_t})$ rates. By Lemma \ref{lem:local-rate}, the centered sketching residual $(\tK_t-K_t)B_t^{-1}\bbm_t$ forms a martingale difference sequence whose conditional second moment is of order $O_p(\eta_t)$. Thus, its cumulative contribution is negligible under the iterate normalization.

The following theorem states the resulting asymptotic linear representations.

\begin{theorem}\label{thm:sketch-noise-reduction}
\setlength{\abovedisplayskip}{3pt}
\setlength{\belowdisplayskip}{3pt}
\setlength{\belowdisplayshortskip}{3pt}
Suppose Assumptions~\ref{ass:1}--\ref{ass:4} hold with $q_g>2$ and $\tau$ satisfies $\tau(1-\sqrt{\mu_t/\nu_t})^{\tau-2}\leq{\myred \gamma_{B,t}}/(8{\myred \Upsilon_{B,t}})$ for all $t\geq0$.~Let $C_\varphi>0$, $C_\eta\in(0,1]$, $0.5<\eta<\varphi\leq1$, and $C_\varphi>1/\{2\lambda_{\min}(I- K^\star)\}$
when $\varphi=1$. Then, the last-iterate admits the linear representation \begin{equation}\label{eq:last-effective-noise}
1/\sqrt{\varphi_t} \cdot (\bx_t-\bx^\star) = -1/\sqrt{\varphi_t} \cdot \sum_{i=0}^{t-1}\Pi_{t,i+1}\varphi_i(I-K^\star)(B^\star)^{-1}\nabla F(\bx^\star;\xi_i)+o_p(1),
\end{equation}
where $\Pi_{n,i}\coloneqq\prod_{j=i}^{n-1}\left\{I-\varphi_j(I-K^\star)\right\}$ for $n>i$ with $\Pi_{n,n}\coloneqq I$.
Moreover, if $\varphi<1$, we have
\begin{equation}\label{eq:averaged-effective-noise}
\sqrt{t}\,(\bar{\bx}_t-\bx^\star) = 1/\sqrt{t} \cdot \sum_{i=0}^{t-1}(\bx_i-\bx^\star) = -1/\sqrt{t} \cdot \sum_{i=0}^{t-1}(B^\star)^{-1}\nabla F(\bx^\star;\xi_i)+o_p(1).
\end{equation}
\end{theorem}

\begin{remark}[\textbf{Interaction between gradient averaging and randomized sketching}]\label{remark:interaction}
\setlength{\abovedisplayskip}{3pt}
\setlength{\belowdisplayskip}{3pt}
\setlength{\belowdisplayshortskip}{3pt}

{\myred The conditions of Theorem~\ref{thm:sketch-noise-reduction} relax the sampled Hessian bounds imposed for vanilla online sketched Newton methods~\citep{Wang2026Inference,Du2026Inference} (except that we have an additional averaging sequence $\eta_t$ satisfying $\varphi_t=o(\eta_t)$).}
To isolate the effect of gradient averaging, consider vanilla Newton sketch that uses $\bg_t$ in place of $\bbm_t$. For each $t\geq 0$, let $\tK_t^\star$ denote the coefficient matrix in \eqref{eq:sketching-residual-matrix} constructed from the step-$t$ sketches and evaluated at $B^\star$ and $(\alpha^\star,\beta^\star,\gamma^\star)$. Building upon \citet[Lemma~B.3]{Wang2026Inference}, we can derive the following linear representation for vanilla Newton sketch:
\begin{equation}\label{eq:single-effective-noise}
1/\sqrt{\varphi_t} \cdot (\bx_t-\bx^\star) = -1/\sqrt{\varphi_t} \cdot \sum_{i=0}^{t-1}\Pi_{t,i+1}\varphi_i(I-\tK_i^\star)(B^\star)^{-1}\nabla F(\bx^\star;\xi_i)+o_p(1).
\end{equation}
Comparing~\eqref{eq:last-effective-noise} and~\eqref{eq:single-effective-noise} reveals the effect of gradient averaging. For our method, the localized moment bounds in Lemma~\ref{lem:local-rate} imply that the centered sketching term $(\tK_t-K_t)B_t^{-1}\bbm_t$ has a second moment of order $O(\eta_t)$ and therefore does not contribute to the leading stochastic sum. In contrast, for vanilla Newton sketch, the corresponding centered sketching term $(\tK_t-K_t)B_t^{-1}\bg_t$ has a constant order of second moment, which inevitably inflates the leading stochastic sum.

The results in~\eqref{eq:last-effective-noise}--\eqref{eq:single-effective-noise} reveal two successive effects. First, replacing the stochastic gradient with its moving average turns the sketching factor $I-\tK_i^\star$ into its mean $I-K^\star$ in the leading stochastic term of the last iterate. Second, averaging the Newton iterates further introduces the factor $(I-K^\star)^{-1}$, which cancels the remaining mean sketching factor. Consequently,~\eqref{eq:averaged-effective-noise} contains no sketch-dependent matrix and has the same leading stochastic term as exact online Newton. In contrast, the averaged iterate of vanilla sketched Newton still contains the centered sketching factor $(I-K^\star)^{-1}(\tK_i^\star-K^\star)$ in its leading term, so the sketching randomness persists in its asymptotic distribution.

\end{remark}

\subsection{Asymptotic normality and covariance comparisons}\label{subsec:statistical-efficiency}

In this section, we build upon Theorem~\ref{thm:sketch-noise-reduction} to establish asymptotic normality and to explore how gradient averaging and randomized sketching affect statistical efficiency. We begin with the last \mbox{iterate}.

\begin{theorem}\label{thm:last-normality}
\setlength{\abovedisplayskip}{3pt}
\setlength{\belowdisplayskip}{3pt}
\setlength{\belowdisplayshortskip}{3pt}
Under the conditions of Theorem~\ref{thm:sketch-noise-reduction}, let $\Gamma_{\bbm,\mathrm{as}}\coloneqq(I-K^\star)\Omega(I-K^\star)$ and we~have
\begin{equation}\label{eq:last-normality}
1/\sqrt{\varphi_t} \cdot (\bx_t-\bx^\star) \xrightarrow{d} \mN(\zero,\Sigma_{\bbm,\mathrm{as}}^{l}),\qquad \text{where}\quad
A_\varphi^\star\Sigma_{\bbm,\mathrm{as}}^{l}+\Sigma_{\bbm,\mathrm{as}}^{l}A_\varphi^\star=\Gamma_{\bbm,\mathrm{as}}.
\end{equation}
\end{theorem}

Recall from Theorem \ref{thm:single-normality} that $A_\varphi^\star=(I-K^\star)-\zeta_\varphi I$ and $\zeta_\varphi=\1_{\{\varphi=1\}}/(2C_\varphi)$.
The result \eqref{eq:last-normality} characterizes $\Sigma_{\bbm,\mathrm{as}}^{l}$ through a Lyapunov equation 
in which randomized sketching enters only through the mean factor matrix $K^\star$. When $\varphi=1$, the condition $C_\varphi>1/\{2\lambda_{\min}(I-K^\star)\}$ ensures that~$A_\varphi^\star\succ\zero$. Hence, the Lyapunov equation in~\eqref{eq:last-normality} admits a unique positive semidefinite solution.

We next show the reduction in $\Sigma_{\bbm,\mathrm{as}}^{l}$ compared to $\Sigma_{\bg,\mathrm{as}}^{l}$ (the asymptotic covariance of vanilla online Newton sketch), and we relate both to the asymptotic covariance of exact online Newton.$\hskip1.5cm$

\begin{proposition}\label{prop:covariance-comparison}
Under the conditions of Theorem~\ref{thm:sketch-noise-reduction}, the following statements hold.
\vskip-0.1cm
\textbf{(a)} We have $\Gamma_{\bbm,\mathrm{as}}\preceq \Gamma_{\bg,\mathrm{as}}$ and $\Sigma_{\bbm,\mathrm{as}}^{l}\preceq\Sigma_{\bg,\mathrm{as}}^{l}$.
\vskip-0.2cm
\textbf{(b)} If the Newton system  \eqref{eq:newton-system} is solved exactly ($\tK^\star=K^\star=\zero$), the last-iterate covariances reduce to $\Sigma_{\bbm,\mathrm{e}}^{l}=\Sigma_{\bg,\mathrm{e}}^{l}=0.5\Omega$ for $\varphi\in(0.5,1)$ and $\Sigma_{\bbm,\mathrm{e}}^{l}=\Sigma_{\bg,\mathrm{e}}^{l}= C_\varphi / (2C_\varphi-1) \cdot \Omega$ for $\varphi=1$.
\vskip-0.2cm
\textbf{(c)} \textbf{($\sqrt{t}$-consistency)} Let $\varphi_t=1/t$ (i.e., $\varphi=C_\varphi=1$). Then, $\Omega=\Sigma_{\bg,\mathrm{e}}^{l}=\Sigma_{\bbm,\mathrm{e}}^{l}\preceq\Sigma_{\bbm,\mathrm{as}}^{l}\preceq\Sigma_{\bg,\mathrm{as}}^{l}$.
\end{proposition}

\vspace{-1pt}
\begingroup
\clubpenalties 1 0\relax
Part~\textbf{(a)} shows the covariance reduction achieved by gradient averaging. The gap $\Sigma_{\bg,\mathrm{as}}^{l}-\Sigma_{\bbm,\mathrm{as}}^{l}$ is positive semidefinite and arises from the centered sketching fluctuation $K^\star-\tK^\star$. Part~\textbf{(b)} shows that gradient averaging leaves the last-iterate covariance of exact online Newton in \eqref{nsequ:1} unchanged. Part~\textbf{(c)} compares sketched and exact Newton with $\varphi_t=1/t$, so that all four last-iterate covariances correspond to the same $\sqrt{t}$ normalization. In this regime, the remaining gap $\Sigma_{\bbm,\mathrm{as}}^{l}-\Omega$ stems from the factor $I-K^\star$ in the Lyapunov equation \eqref{eq:last-normality}. Thus, gradient averaging eliminates the covariance contribution from the centered sketching fluctuation $K^\star-\tK^\star$, while the residual dependence on~$I-K^\star$ in \eqref{eq:last-effective-noise} prevents the covariance from further reducing to that of exact Newton method.
\par
\endgroup

Lastly, we turn to the Polyak--Ruppert average $\bar{\bx}_t$ studied in~\eqref{eq:averaged-effective-noise}.

\begin{theorem}\label{thm:averaged-normality}
Under the conditions of Theorem~\ref{thm:sketch-noise-reduction}, suppose that $\varphi<1$. Then, we have
\begin{equation}\label{eq:averaged-normality}
\sqrt{t}\,(\bar{\bx}_t-\bx^\star) \xrightarrow{d} \mN(\zero,\Sigma_{\bbm,\mathrm{as}}^{a}), \qquad \text{where}\quad \Sigma_{\bbm,\mathrm{as}}^{a}=\Omega.
\end{equation}
\end{theorem}

The result~\eqref{eq:averaged-effective-noise} reduces the averaged iterate to a normalized sum of i.i.d. mean-zero random vectors up to an $o_p(1)$ term. Therefore, the multivariate central limit theorem and Slutsky's theorem yield~\eqref{eq:averaged-normality}. 
To quantify the effect of gradient averaging, consider vanilla online sketched Newton in~\eqref{nsequ:2}. By Theorem~\ref{thm:single-normality} and the definition of $\Gamma_{\bg,\mathrm{as}}$ in~\eqref{eq:single-limiting-covariances}, its averaged-iterate covariance~is$\quad$
\begin{multline}\label{eq:single-averaged-covariance}
\Sigma_{\bg,\mathrm{as}}^{a} \coloneqq (I-K^\star)^{-1}\Gamma_{\bg,\mathrm{as}}(I-K^\star)^{-1} \\
=\Omega+(I-K^\star)^{-1}\mE [(\tK^\star-K^\star)\Omega(\tK^\star-K^\star)^\top ](I-K^\star)^{-1}\succeq\Omega.
\end{multline}
Comparing \eqref{eq:averaged-normality} with \eqref{eq:single-averaged-covariance} reveals a sharp statistical distinction. Vanilla online sketched Newton suffers from a positive semidefinite covariance inflation induced by the centered sketching fluctuation, while $\Sigma_{\bbm,\mathrm{as}}^{a}=\Omega$. The covariance $\Omega$ coincides with that of averaged exact online Newton in \eqref{nsequ:1} and that of averaged SGD~\citep{Ruppert1988Efficient,Polyak1992Acceleration}, which is known to be asymptotically minimax optimal~\citep{Duchi2021Asymptotic,Davis2024Asymptotic}. Therefore, for every fixed $\tau$ satisfying the conditions of Theorem~\ref{thm:global-convergence}, accelerated sketching incurs \textbf{no computational--statistical trade-off for the averaged iterate}: it significantly \textit{reduces the cost} of Newton system solve while preserving~the asymptotically \textit{optimal statistical efficiency} achieved by averaged exact Newton and SGD.

\section{Numerical Experiments}\label{sec:experiments}

We compare online Newton methods with and without gradient averaging, using either accelerated sketching or exact solves, and include SGD as a baseline. We consider linear and logistic regression with Gaussian designs, vary $d\in\{20,40\}$, and vary the design covariance structure among identity, equicorrelation, or Toeplitz. Kaczmarz and Gaussian sketches with $\tau\in\{5,10\}$ yield 48 matched settings. Each configuration uses $R=200$ independent runs of $T=10^5$ iterations. All methods share the same data streams, and matched sketching methods additionally share the same sketches.~$\hat{\Sigma}$ denotes the centered empirical covariances, computed across runs, of $(\bx_t-\bx^\star)/\sqrt{\varphi_t}$ and $\sqrt{t}(\bar{\bx}_t-\bx^\star)$. Bias and mean squared error (MSE) are evaluated separately. Appendix~\ref{app:experiments} provides complete experimental setups and additional results{\myred , including estimation-error comparisons with SGD and comparisons with averaged SGD on confidence-interval coverage, length, and computational cost}.

\begin{figure}[t]
\centering
\captionsetup{font=footnotesize,skip=5pt}
\captionsetup[subfigure]{font=footnotesize,skip=1pt}
\begin{subfigure}[t]{0.247\linewidth}
\centering
\includegraphics[width=\linewidth]{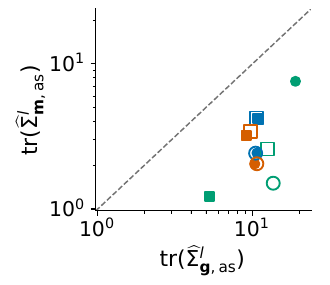}\vskip-0.1cm
\caption{Linear: last}
\end{subfigure}\hfill
\begin{subfigure}[t]{0.247\linewidth}
\centering
\includegraphics[width=\linewidth]{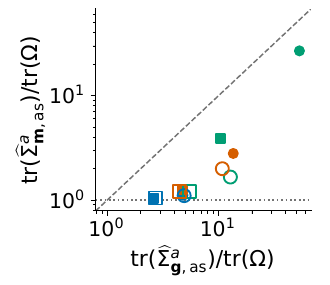}\vskip-0.1cm
\caption{Linear: averaged}
\end{subfigure}\hfill
\begin{subfigure}[t]{0.247\linewidth}
\centering
\includegraphics[width=\linewidth]{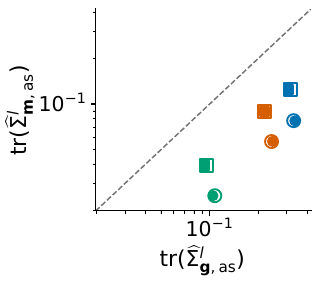}\vskip-0.1cm
\caption{Logistic: last}
\end{subfigure}\hfill
\begin{subfigure}[t]{0.247\linewidth}
\centering
\includegraphics[width=\linewidth]{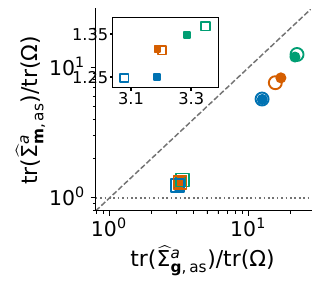}\vskip-0.1cm
\caption{Logistic: averaged}
\end{subfigure}
\par
\includegraphics[width=0.94\linewidth,trim={0pt 4pt 0pt 4pt},clip]{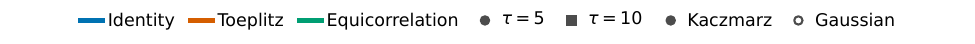}
\par\vspace{-0.1cm}
\begin{subfigure}[t]{0.247\linewidth}
\centering
\includegraphics[width=\linewidth]{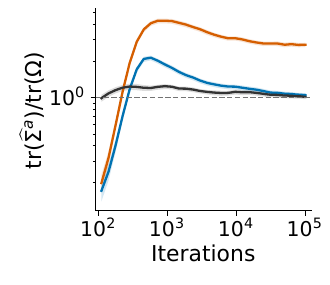}\vspace{-0.2cm}
\caption{Linear: covariance}
\end{subfigure}\hfill
\begin{subfigure}[t]{0.247\linewidth}
\centering
\includegraphics[width=\linewidth]{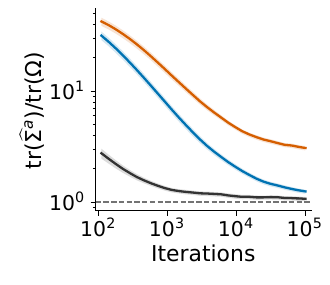}\vspace{-0.2cm}
\caption{Logistic: covariance}
\end{subfigure}\hfill
\begin{subfigure}[t]{0.247\linewidth}
\centering
\includegraphics[width=\linewidth]{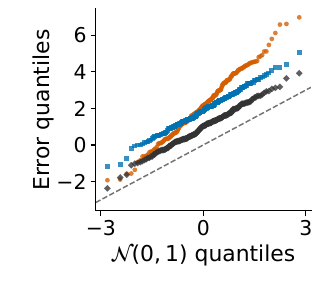}\vspace{-0.2cm}
\caption{Linear: normal QQ}
\end{subfigure}\hfill
\begin{subfigure}[t]{0.247\linewidth}
\centering
\includegraphics[width=\linewidth]{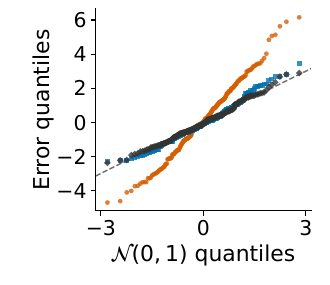}\vspace{-0.2cm}
\caption{Logistic: normal QQ}
\end{subfigure}
\par\vspace{-0.1cm}
\includegraphics[width=0.88\linewidth]{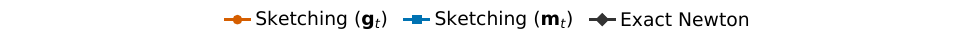}
\vspace{-0.2cm}
\caption{\textit{
Covariance comparisons ($d=20$, $T=10^5$, $R=200$). Panels (a--d) compare covariance traces without (horizontal axis) and with (vertical axis) gradient averaging across 12 settings. All averaged-iterate covariance traces are normalized by $\operatorname{tr}(\Omega)$. For identity design covariance, Gaussian sketches, and $\tau=10$, panels (e,f) show covariance trace trajectories with pointwise 95\% bootstrap intervals, while panels (g,h) show standard normal QQ plots of $\sqrt{T}\,\mathbf{1}^\top(\bar{\bx}_T-\bx^\star)/\sqrt{\mathbf{1}^\top\Omega\mathbf{1}}$ without recentering. Diagonal lines indicate equality, and horizontal lines indicate one. The inset magnifies the six points corresponding to $\tau=10$. Sketching uses $\bg_t$ without gradient averaging and $\bbm_t$ with gradient averaging, while exact Newton uses $\bg_t$.
}}
\label{fig:covariance-comparison}
\label{fig:pr-trajectories}
\end{figure}

\vskip-0.1cm
\noindent$\bullet$ \textbf{Covariance reduction with gradient averaging.}
Figure~\ref{fig:covariance-comparison}(a,c) shows that gradient averaging reduces the last-iterate covariance trace with the same data streams, sketches, inner sketching solve budgets, and outer stepsizes. In particular, each point on the plot corresponds to a specific configuration, and all points lie below the line $y=x$. Thus, the reduction holds across design covariance structures and sketching distributions, including higher-dimensional settings reported in Appendix~\ref{app:experiments}.
By comparison, Table~\ref{tab:experiment-ratios} shows that the covariance trace of exact Newton remains nearly unchanged~under gradient averaging. These observations align with Proposition~\ref{prop:covariance-comparison}: gradient averaging reduces the last-iterate limiting covariance of sketched Newton while leaving that~of exact Newton unchanged.$\quad$

\vskip-0.1cm
\noindent$\bullet$ \textbf{Comparison with exact online Newton.}
Figure~\ref{fig:covariance-comparison}(b,d,e,f) shows that gradient averaging brings the averaged-iterate covariance trace closer to the exact Newton benchmark, since not only do all points lie below $y=x$, but more points also lie close to $y=1$. Note that both vanilla sketched Newton and our method adopt iterate averaging, so the improvement is purely attributable to gradient averaging. Theorem~\ref{thm:averaged-normality} explains this comparison: gradient averaging along with iterate averaging recovers $\Omega$ asymptotically, while vanilla sketched Newton retains an additional sketch-dependent covariance contribution. The trajectories in panels (e,f) further support this conclusion: with gradient averaging, the covariance trace approaches $\operatorname{tr}(\Omega)$ while the number of sketching steps per iteration~remains~fixed.

\vskip-0.1cm
\noindent$\bullet$ \textbf{Estimation accuracy and normal approximation.}
From Table~\ref{tab:experiment-summary}, we observe that gradient averaging reduces the MSE of both the last and averaged iterates across all tested settings. The bias comparisons in Appendix~\ref{app:experiments} show similar improvements, indicating that the covariance reduction is accompanied by improved estimation accuracy. Increasing $\tau$ also brings the averaged-iterate covariance trace closer to $\operatorname{tr}(\Omega)$. Thus, although the limiting covariance is independent of sketching, more accurate inner solves remain beneficial at finite iterations. Figure~\ref{fig:covariance-comparison}(g,h) further complements these comparisons. In particular, with gradient averaging, the logistic regression QQ plot closely follows the standard normal reference. The linear regression QQ plot exhibits a similar pattern but~is~shifted~due~to~finite-sample bias; this shift is still smaller with gradient averaging.

\begin{table}[t]
\centering
\captionsetup{font=footnotesize,skip=3pt}
\vspace{-0.2cm}
\caption{\textit{
Comparisons at $T=10^5$ with $R=200$. Covariance-trace and MSE ratios compare sketched Newton with gradient averaging to its vanilla counterpart. The central columns report $\operatorname{tr}(\widehat\Sigma_{u,\mathrm{as}}^{a})/\operatorname{tr}(\Omega)$ for~$u\in\{\bg,\bbm\}$. Entries are medians of six settingwise ratios across three design covariances and two sketching distributions.}}
\label{tab:experiment-summary}
\footnotesize
\newcommand{\summarylabel}[1]{{\fontsize{8}{9}\selectfont #1}}
\renewcommand{\arraystretch}{0.7}
\setlength{\aboverulesep}{0.15ex}
\setlength{\belowrulesep}{0.2ex}
\begin{tabular}{lccrrrrrr}
\toprule
& & & \multicolumn{2}{c}{\summarylabel{Covariance-trace ratio}}
& \multicolumn{2}{c}{\summarylabel{Averaged trace$/\operatorname{tr}(\Omega)$}}
& \multicolumn{2}{c}{\summarylabel{MSE ratio}}\\
\cmidrule(lr){4-5}\cmidrule(lr){6-7}\cmidrule(lr){8-9}
\summarylabel{Model} & \summarylabel{$d$} & \summarylabel{$\tau$}
& \summarylabel{Last} & \summarylabel{Averaged}
& \summarylabel{$\bg$} & \summarylabel{$\bbm$}
& \summarylabel{Last} & \summarylabel{Averaged}\\
\midrule
\summarylabel{Linear} & 20 & 5  & 0.213 & 0.215 & 11.808  & 1.830   & 0.213 & 0.249\\
& 20 & 10 & 0.352 & 0.323 & 4.619   & 1.201   & 0.352 & 0.349\\
& 40 & 5  & 0.164 & 0.232 & 66.756  & 19.108  & 0.234 & 0.301\\
& 40 & 10 & 0.208 & 0.219 & 15.775  & 3.299   & 0.208 & 0.253\\
\midrule
\summarylabel{Logistic} & 20 & 5  & 0.233 & 0.490 & 16.234  & 7.953   & 0.233 & 0.483\\
& 20 & 10 & 0.411 & 0.409 & 3.195   & 1.314   & 0.412 & 0.408\\
& 40 & 5  & 0.686 & 0.854 & 751.846 & 641.821 & 0.691 & 0.814\\
& 40 & 10 & 0.226 & 0.510 & 18.316  & 9.343   & 0.225 & 0.496\\   
\bottomrule
\end{tabular}
\vskip-0.35cm
\end{table}

\section{Conclusion}\label{sec:conclusion}

We studied statistical efficiency of online Newton methods with Nesterov accelerated sketching and gradient averaging. We established almost-sure convergence and asymptotic normality of the last and averaged iterates. We showed that gradient averaging yields a last-iterate covariance no larger than that of vanilla online sketched Newton, while subsequent averaging of the outer iterates further achieves the minimax optimal limiting covariance of exact online Newton. Thus, for averaged iterates, a fixed number of sketching steps can reduce the cost of an expensive Newton solve without inflating the limiting covariance. Gradient averaging requires no extra oracle evaluations and leaves the contraction bound of the inner sketching solver unchanged. Experiments on linear and logistic regression support our free-lunch finding and further demonstrate improvements in bias and mean squared error.
We should mention that our asymptotic normality only provides the first step toward practical inference. Establishing theoretical guarantees for online covariance estimation or random scaling based on the gradient-averaged iterates remains an important direction for future work.

\subsection*{AI use statement}

We used generative AI tools to assist with writing and polishing the manuscript. We have reviewed all AI-assisted text and take responsibility for the final manuscript, including text, claims, proofs, and experimental results.

\bibliography{ref}
\bibliographystyle{iclr2027_conference}
\clearpage

\appendix
\startcontents[appendix]

\begin{center}
\Large \bf Appendix: ``\papertitle"
\end{center}

\section*{Contents}
\printcontents[appendix]{}{1}[2]{}

\clearpage

{\myred \section{Additional Discussion}\label{app:additional-discussion}}

{\myred In this section, we provide additional discussion of several aspects of the main text: the convergence rates and stepsize conditions for SGD and online Newton methods; Assumption~\ref{ass:1}, particularly the population strong convexity of the objective and the relationship between the conditions on $B_t$ and those used in conditioned SGD; the choice of stepsizes for statistical inference; and the computational complexity, including the required number of sketching steps and the cost per outer iteration.}

{\myred \noindent$\bullet$ \textbf{Rates of SGD and online Newton.} 
In the online setting with vanishing stepsizes, SGD and online Newton methods share the same convergence rates, which are governed by the stepsize: for both families, the last iterate converges at rate $O_p(\sqrt{\varphi_t})$, while the averaged iterate converges at rate $O_p(1/\sqrt{t})$. Their distinction for statistical inference lies instead in the limiting covariance of the iterates and the accuracy of covariance estimation, as discussed in Section~\ref{sec:intro}. However, a difference arises in the stepsize requirements for last-iterate asymptotic normality. When $\varphi_t=C_\varphi/(t+1)$, SGD requires $C_\varphi>1/\{2\lambda_{\min}(B^\star)\}$, for which Assumption~\ref{ass:1} yields the sufficient condition $C_\varphi>1/(2\gamma_H)$. In either case, the requirement depends on the unknown curvature of the objective. In contrast, exact online Newton requires only $C_\varphi>1/2$ due to affine invariance, as Hessian preconditioning removes the dependence on the objective curvature. For accelerated sketched Newton, the corresponding condition is $C_\varphi>1/\{2\lambda_{\min}(I-K^\star)\}$ (cf. Lemma~\ref{lem:local-rate} and Theorem~\ref{thm:sketch-noise-reduction}), which reduces to $C_\varphi>1/2$ when the Newton systems are solved exactly, i.e., $K^\star=\zero$.
}

{\myred 
\noindent$\bullet$ \textbf{Population curvature and sample Hessians.} 
Assumption~\ref{ass:1} requires strong convexity and smoothness only of the \textit{population} objective $f$, without imposing such conditions on the sample functions $F(\cdot;\xi)$. These are standard conditions in SGD-based inference~\citep{Polyak1992Acceleration,Chen2020Statistical,Zhu2021Online}. Although SGD can converge under much weaker conditions, including for nonconvex problems, strong convexity remains a common and convenient assumption for \emph{inference}, for two reasons. First, inference is inherently local: it quantifies the uncertainty of the iterates around a strict local minimizer $\bx^\star$ satisfying $B^\star=\nabla^2f(\bx^\star)\succ0$. By continuity of $\nabla^2f$, strong convexity then holds in a neighborhood of $\bx^\star$. The global lower bound on $\nabla^2f$ is imposed here to establish $\bx_t\stackrel{a.s.}{\rightarrow}\bx^\star$ in Theorem~\ref{thm:global-convergence}, while the essential curvature requirement for inference is local. Second, positive definiteness of the population Hessian at the solution is essential for establishing meaningful statistical lower bounds. At such a minimizer, $\Omega$ is well defined and characterizes the local asymptotic minimax covariance~\citep{Duchi2021Asymptotic,Davis2024Asymptotic}. Recent instance-dependent nonasymptotic (local minimax) lower bounds are likewise established for strongly convex population objectives~\citep{Jiang2026InstanceOptimal}. Thus, our conditions on the objective are no~stronger than those commonly imposed for SGD inference.

For the sample functions, Assumption~\ref{ass:1} requires only twice continuous differentiability, and Assumption~\ref{ass:3} controls the second moment of the Hessian noise. In particular, we do not require the sample Hessians to be positive definite or uniformly bounded \citep{Kuang2025Online, Wang2026Inference, Du2026Inference}. Linear regression illustrates the importance of this relaxation. Consider $F(\bx;(\ba,y))=(\ba^\top\bx-y)^2/2$ with $\ba\sim\mN(\zero,\Sigma_a)$ and $\Sigma_a\succ0$. The sample Hessian $\ba\ba^\top$ is rank one and has no uniform upper bound, so uniform sample-level spectral bounds cannot hold. In contrast, the population Hessian satisfies $\nabla^2f(\bx)=\mE[\ba\ba^\top]=\Sigma_a\succ0$, and the clipped Hessian average $B_t$ in~\eqref{eq:clipped-hessian-average} satisfies the conditions on $B_t$ in Assumption~\ref{ass:1} by construction, without requiring spectral bounds on individual sample Hessians. We also note that, as is standard, a regularization term is included in logistic regression to ensure the global lower bound on $\nabla^2f$. See Appendix~\ref{app:experiment-design}.

}

{\myred \noindent$\bullet$ \textbf{Relation to conditioned SGD.} 
For the Hessian approximation $B_t$, Assumption~\ref{ass:1} adopts \textbf{exactly the same} conditions used in conditioned SGD \citep[Assumption~9]{Leluc2023Asymptotic}: $B_t$ is $\mathcal F_{t-1}$-measurable and has lower and upper bounds of the same order. No additional conditions on $B_t$ are imposed. In particular, we do not require any sample Hessian to be positive definite. To see the correspondence, note that the conditioning matrix applied to the gradient in our Newton update is $B_t^{-1}$, and the bounds in Assumption~\ref{ass:1} give $1/\Upsilon_{B,t}\cdot I\preceq B_t^{-1}\preceq1/\gamma_{B,t}\cdot I$. \citet[Assumption~9]{Leluc2023Asymptotic} states the stepsize conditions through the upper bound of the conditioning matrix, which is $1/\gamma_{B,t}$ here, so the two series conditions become exactly $\sum_t\varphi_t/\gamma_{B,t}=\infty$ and $\sum_t(\varphi_t/\gamma_{B,t})^2<\infty$. Likewise, the ratio of the upper and lower bounds, which is required to be bounded, is $\Upsilon_{B,t}/\gamma_{B,t}$. 
}

{\myred 

\noindent$\bullet$ \textbf{Choice of stepsizes for inference.} 
There is no universally optimal exponent $\varphi$: its choice depends on whether the last or averaged iterate is used for inference and reflects a trade-off between nonasymptotic and asymptotic performance. \textbf{(i)}~For nonasymptotic convergence, $\varphi_t\propto1/\sqrt{t}$, i.e., $\varphi=0.5$, is known to achieve the optimal convergence rate~\citep{Moulines2011Non}. \textbf{(ii)}~For the averaged iterate, any $\varphi\in(0.5,1)$ yields $\sqrt{t}$-consistency with the same optimal limiting covariance $\Omega$~\citep{Polyak1992Acceleration}, as also established for our method in Theorem~\ref{thm:averaged-normality}. Thus, when using the averaged iterate for inference, one would favor $\varphi$ close to $0.5$, which improves nonasymptotic performance and allows the averaged iterate to enter its asymptotic regime sooner, as larger stepsizes diminish the effect of initialization more rapidly. \textbf{(iii)}~For the last iterate, the situation is reversed. Its asymptotic normality is established for $(\bx_t-\bx^\star)/\sqrt{\varphi_t}$, so its variance is of order $\varphi_t=C_\varphi/(t+1)^\varphi$. An exponent close to $0.5$ therefore leads to a larger variance, while $\sqrt{t}$-consistency with the optimal last-iterate covariance is attained only when $\varphi_t\propto1/t$ (Proposition~\ref{prop:covariance-comparison}(c)). Thus, when using the last iterate for inference, one would favor $\varphi$ close to $1$, at the expense of slower nonasymptotic convergence. At $\varphi=1$, the condition on $C_\varphi$ in Theorem~\ref{thm:sketch-noise-reduction} becomes relevant, and Newton methods impose weaker requirements on $C_\varphi$ than SGD, as discussed above. In our experiments, we choose $\varphi=0.8$ as a compromise that allows both estimators to be evaluated using the same runs.

Constant stepsizes are common when optimization is the primary objective, as in training. However, they are generally unsuitable for achieving optimal statistical efficiency in inference, which is the focus of this paper. With a constant stepsize, SGD iterates typically do not converge to $\bx^\star$ but instead approach a stationary distribution whose mean is biased away from $\bx^\star$ by an amount of the order of the stepsize. Iterate averaging reduces the variance but does not eliminate this bias~\citep{Dieuleveut2020Bridging,Yu2021analysis,Huo2026Bias}. Consequently, the resulting estimator is generally not consistent, let alone $\sqrt{t}$-consistent with a limiting covariance comparable to $\Omega$. For this reason, the existing literature on online inference with SGD and Newton methods, upon which we build, adopts decaying stepsizes~\citep{Chen2020Statistical,Zhu2021Online,Leluc2023Asymptotic,Kuang2025Online,Wang2026Inference,Du2026Inference}. Removing the bias induced by constant stepsizes requires additional techniques, such as Richardson--Romberg extrapolation~\citep{Dieuleveut2020Bridging,Huo2026Bias}, which are beyond the scope of the present study.

}

{\myred \noindent$\bullet$ \textbf{Per-iteration computational cost.}
Gradient averaging does not alter the sketching solver, so the discussion below follows the existing literature on sketch-and-project methods~\citep{Gower2015Randomized,Gower2018Accelerated,Derezinski2025Fine}. The computational cost of the sketching solver~is determined by the number of sketching steps and the cost per step.

Theorems~\ref{thm:global-convergence} and~\ref{thm:sketch-noise-reduction} require $\tau$ satisfying $\tau(1-\sqrt{\mu_t/\nu_t})^{\tau-2}\leq\gamma_{B,t}/(8\Upsilon_{B,t})$ for all $t$. Since~$1-\sqrt{\mu_t/\nu_t}\leq1-\gamma_S<1$ by Assumption~\ref{ass:4} and $\sup_{t\geq0}\Upsilon_{B,t}/\gamma_{B,t}<\infty$ almost surely by Assumption~\ref{ass:1}, the required number of sketching steps is uniformly bounded in $t$. In particular, we do not require the approximation error in solving the Newton systems to vanish over time. At each sketching step, the expected squared error contracts by a factor of $1-\mu_t$ without acceleration and $1-\sqrt{\mu_t/\nu_t}\leq1-\mu_t$ with acceleration (Lemma~\ref{lem:accelerated-inner-rate}). For randomized coordinate sketching $S=\be_j$, we have $\mu_t\geq1/(d\kappa_t^2)$ and $\nu_t=d$, where $\kappa_t=\lambda_{\max}(B_t)/\lambda_{\min}(B_t)$. Consequently, acceleration reduces the number of steps required to reach a prescribed accuracy $\epsilon$ from $O(d\kappa_t^2\log(1/\epsilon))$~to $O(d\kappa_t\log(1/\epsilon))$.

The per sketching step cost depends on the sketching distribution and the sparsity of the sketching matrix. For randomized coordinate sketching, each step accesses only one column of $B_t$ and costs~$O(d)$, with or without acceleration.

Finally, by (i) and (ii), the sketching solver costs $O(d)\times O(d)=O(d^2)$ per outer iteration, compared with $O(d^3)$ for an exact Newton solve. The constant hidden in this bound depends on the condition number $\kappa_t$, and acceleration improves the worst-case dependence from $\kappa_t^2$ to $\kappa_t$. The remaining components do not increase the computational order: updating the Hessian average costs $O(d^2)$, spectral clipping, performed once every $d$ iterations, incurs an amortized cost of $O(d^2)$, and gradient averaging adds only an $O(d)$ vector update.	

}

\section{Preparation Lemmas}

\begin{lemma}[Kronecker's lemma]\label{lem:kronecker}
Suppose $\{A_t\}_{t\geq0}$ satisfies $\sum_{t=0}^{\infty}A_t<\infty$. Then, for any positive, non-decreasing sequence $\{a_t\}_{t\geq0}$ satisfying $a_t\rightarrow\infty$, we have
$\frac{1}{a_T}\sum_{t=1}^{T}a_tA_t\rightarrow0$ as $T\rightarrow\infty$.
\end{lemma}

\begin{lemma}[Stolz--Ces\`aro theorem]\label{lem:stolz-cesaro}
Let $\{a_t\}_{t\geq1}$ and $\{b_t\}_{t\geq1}$ be two sequences of real numbers. Suppose that $\{b_t\}_{t\geq1}$ is strictly monotone and divergent, and that the following limit exists:
\begin{equation*}
\lim_{t\rightarrow\infty}\frac{a_{t+1}-a_t}{b_{t+1}-b_t}=\ell.
\end{equation*}
Then,
\begin{equation*}
\lim_{t\rightarrow\infty}\frac{a_t}{b_t}=\ell.
\end{equation*}
\end{lemma}

\begin{lemma}\citep[Lemma~3]{Leluc2023Asymptotic}\label{lem:polynomial_product_rate}
Let $\{\beta_t\}_{t\geq1}$ be a nonnegative sequence converging to zero, and let $\lambda>0$, $m\geq1$, and $q\geq0$. Consider two nonnegative sequences $\{A_t\}_{t\geq0}$ and $\{\varepsilon_t\}_{t\geq1}$ such that, for all sufficiently large $t$,
\begin{equation*}
A_t=(1-\lambda\beta_t)^mA_{t-1}+\beta_t^{q+1}\varepsilon_t.
\end{equation*}
If $\beta_t=\beta_0/(t+1)^a$ with $a\in(0,1)$, then
\begin{equation*}
\limsup_{t\rightarrow\infty}\frac{A_t}{\beta_t^q}
\leq\frac{1}{m\lambda}\limsup_{t\rightarrow\infty}\varepsilon_t.
\end{equation*}
If $\beta_t=\beta_0/(t+1)$ and $q<m\lambda\beta_0$, then
\begin{equation*}
\limsup_{t\rightarrow\infty}\frac{A_t}{\beta_t^q}
\leq\frac{\beta_0}{m\lambda\beta_0-q}\limsup_{t\rightarrow\infty}\varepsilon_t.
\end{equation*}
\end{lemma}

\begin{lemma}\citep[Corollary~3.1]{Hall2014Martingale}\label{lemma:hall}
For every $t\geq1$, let $\{W_{t,i}\}_{i=0}^{t-1}$ be a triangular array of $\mR^d$-valued random vectors such that $W_{t,i}$ is $\mF_i$-measurable and
\begin{align}
\mE[W_{t,i}\mid\mF_{i-1}] &= \zero,\label{hh1}\\
\sum_{i=0}^{t-1}\mE[W_{t,i}W_{t,i}^\top\mid\mF_{i-1}] &\xrightarrow{p} V^\star\succeq\zero,\label{hh2}\\
\sum_{i=0}^{t-1}\mE[\|W_{t,i}\|^2\1_{\{\|W_{t,i}\|>\epsilon\}}\mid\mF_{i-1}] &\xrightarrow{p}0,\qquad\forall\,\epsilon>0.\label{hh3}
\end{align}
Then, we have
\begin{equation*}
\sum_{i=0}^{t-1}W_{t,i}\xrightarrow{d}\mN(\zero,V^\star).
\end{equation*}
\end{lemma}

\section{Proofs of Section~\ref{sec:convergence}}\label{app:global-convergence}

Throughout this section, we write $f_t\coloneqq f(\bx_t)$ and $f^\star\coloneqq f(\bx^\star)$ and use $\1_{\{\cdot\}}$ to denote the indicator function.

\subsection{Proof of Theorem~\ref{thm:global-convergence}}\label{subapp:global-convergence}

We first introduce the inner-solver quantities used in the proof. For every $t\geq0$, we define
\begin{equation*}
\tZ_{t,j}\coloneqq B_tS_{t,j}(S_{t,j}^\top B_t^2S_{t,j})^\dagger S_{t,j}^\top B_t \in \mR^{d \times d}
\end{equation*}
and
\begin{align}\label{equ:Cttilde}
\tC_{t,j}
&\coloneqq
\begin{pmatrix}
1-\alpha_t & \alpha_t\\
(1-\alpha_t)(1-\beta_t) & \alpha_t+\beta_t-\alpha_t\beta_t
\end{pmatrix}\otimes I_d
-\begin{pmatrix}
1-\alpha_t & \alpha_t\\
(1-\alpha_t)\gamma_t & \alpha_t\gamma_t
\end{pmatrix}\otimes\tZ_{t,j}.
\end{align}
Let $\Tilde C_t\coloneqq\tC_{t,\tau-1}\cdots\tC_{t,0}$ and write $[\Tilde C_t]_{a,b}$ for its $(a,b)$-th $d\times d$ block. The realized residual matrix introduced in~\eqref{eq:sketching-residual-matrix} has the explicit representation
\begin{equation}\label{def:tilde_K_t}
\tK_t\coloneqq[\Tilde C_t]_{1,1}+[\Tilde C_t]_{1,2}
=\begin{pmatrix}I&\zero\end{pmatrix}\Tilde C_t\begin{pmatrix}I\\I\end{pmatrix}.
\end{equation}
Moreover, we define
\begin{equation}\label{eq:mean-inner-transition}
M_t\coloneqq
\begin{pmatrix}
1-\alpha_t & \alpha_t\\
(1-\alpha_t)(1-\beta_t) & \alpha_t+\beta_t-\alpha_t\beta_t
\end{pmatrix}\otimes I_d
-\begin{pmatrix}
1-\alpha_t & \alpha_t\\
(1-\alpha_t)\gamma_t & \alpha_t\gamma_t
\end{pmatrix}\otimes Z_t.
\end{equation}
The conditional independence of the inner sketches gives
\begin{equation*}
C_t\coloneqq\mE[\Tilde C_t\mid\mF_{t-1}]=M_t^\tau,
\qquad
K_t\coloneqq\mE[\tK_t\mid\mF_{t-1}]
=\begin{pmatrix}I&\zero\end{pmatrix}C_t\begin{pmatrix}I\\I\end{pmatrix}.
\end{equation*}

The matrices $\tK_t$ and $K_t$ are precisely the realized and mean residual matrices introduced in Section~\ref{sec:preliminaries}. Indeed, the block matrices in~\eqref{equ:Cttilde} and their conditional means depend on $B_t$, the acceleration parameters, and the current sketches or their distribution, but not on the right-hand side of the Newton system. Therefore, replacing $\bg_t$ with $\bbm_t$ changes neither $\tK_t$ nor $K_t$. Consequently, the bound $1-\sqrt{\mu_t/\nu_t}\leq1-\gamma_S<1$ from \citet[Lemma~3.6]{Wang2026Inference} and the mean-residual bound from \citet[Lemma~3.7]{Wang2026Inference} continue to hold without modification. For the moving-average gradient estimates, the exact direction $\Delta\bx_t=-B_t^{-1}\bbm_t$, and the conditional-mean of the approximate direction $\bz_{t,\tau}$ must be formulated with respect to $\mF_{t-0.5}$.

The following auxiliary lemma records these two relations. Its proof is deferred to Appendix~\ref{app:inner-transition}.

\begin{lemma}\label{lem:inner_transition}
Define $\Delta\bx_t\coloneqq-B_t^{-1}\bbm_t$. Then, for every $t\geq0$, we have
\begin{equation*}
\bz_{t,\tau}-\Delta\bx_t=\tK_t(\bz_{t,0}-\Delta\bx_t),
\qquad
\mE[\bz_{t,\tau}\mid\mF_{t-0.5}]=-(I-K_t)B_t^{-1}\bbm_t.
\end{equation*}
Consequently, $(\bz_{t,\tau}-(I-K_t)\Delta\bx_t,\mF_t)_{t\geq0}$ forms a martingale difference sequence. Furthermore, if $\gamma_t=1$, then $\tK_t$ has a simple form of $\tK_t=(I-\tZ_{t,\tau-1})\cdots(I-\tZ_{t,0})$.
\end{lemma}

The realized block matrices in~\eqref{equ:Cttilde} and $\tK_t$ depend on $B_t$, the acceleration parameters, and the inner sketches, while their conditional means depend on $B_t$, the acceleration parameters, and the sketch distribution. None of these matrices depends on the right-hand side of the Newton system. Therefore, replacing $\bg_t$ with $\bbm_t$ does not affect the bound $1-\sqrt{\mu_t/\nu_t}\leq1-\gamma_S<1$ from \citet[Lemma~3.6]{Wang2026Inference} or the mean-residual bound from \citet[Lemma~3.7]{Wang2026Inference}; only the conditional-mean relation in Lemma~\ref{lem:inner_transition} must be stated with respect to $\mF_{t-0.5}$.

By Assumption~\ref{ass:1}, $f(\bx)$ is strongly convex. By the basic property of strong convexity \citep{Nesterov2018Lectures}, we have the following:
\begin{equation}\label{eq:global-strong-convexity}
\frac{\gamma_H}{2}\|\bx_t-\bx^{\star}\|^2\leq f_t-f^{\star}\leq\frac{1}{2\gamma_H}\|\nabla f_t\|^2.
\end{equation}
Still, by the $\Upsilon_H$-Lipschitz continuity of $\nabla f(\bx)$ implied by Assumption~\ref{ass:1}, we have
\begin{equation}\label{eq:global-smoothness}
\frac{1}{2\Upsilon_H}\|\nabla f_t\|^2\leq f_t-f^{\star}\leq\frac{\Upsilon_H}{2}\|\bx_t-\bx^{\star}\|^2.
\end{equation}
Furthermore, the $\Upsilon_H$-Lipschitz continuity of $\nabla f(\bx)$ yields
\begin{equation*}
f_{t+1}-f^{\star}\leq f_t-f^{\star}+\varphi_t\nabla f_t^{\top}\bz_{t,\tau}+\frac{\Upsilon_H}{2}\varphi_t^2\|\bz_{t,\tau}\|^2.
\end{equation*}
Taking the conditional expectation $\mE[\cdot\mid\mF_{t-0.5}]$ on both sides yields
\begin{equation}\label{eq:global-conditional-taylor}
\mE[f_{t+1}-f^{\star}\mid\mF_{t-0.5}]\leq f_t-f^{\star}+\varphi_t\mE[\nabla f_t^{\top}\bz_{t,\tau}\mid\mF_{t-0.5}]+\frac{\Upsilon_H}{2}\varphi_t^2\mE[\|\bz_{t,\tau}\|^2\mid\mF_{t-0.5}].
\end{equation}
{\red In addition, by Assumption~\ref{ass:1}, $\gamma_{B,t} I \preceq B_t \preceq \Upsilon_{B,t} I$ and hence we obtain $\Upsilon_{B,t}^{-1} I \preceq B_t^{-1} \preceq \gamma_{B,t}^{-1}I$ almost surely for every $t \geq 0$.}

We first control the conditional mean of the search direction. If $\mu_t=1$, the eigendirection representation in the proof of \citet[Lemma~3.7]{Wang2026Inference} gives $K_t=\zero$. If $\mu_t<1$, \citet[Lemma~3.7]{Wang2026Inference} and the condition on $\tau$ give
\begin{equation*}
{\red \|K_t\| \leq 2 \tau \left(1 - \sqrt{\frac{\mu_t}{\nu_t}}\right)^{\tau - 2} \leq 2 \cdot \frac{\gamma_{B,t}}{8 \Upsilon_{B,t}} = \frac{\gamma_{B,t}}{4 \Upsilon_{B,t}}.}
\end{equation*}
{\red Therefore, in either case, it holds almost surely that $\|K_t\| \leq \gamma_{B,t} / (4 \Upsilon_{B,t})$ for every $t \geq 0$.}
By Lemma~\ref{lem:inner_transition}, we obtain
\begin{align}\label{eq:global-mean-descent}
\varphi_t\mE[\nabla f_t^{\top}\bz_{t,\tau}\mid\mF_{t-0.5}] &= -\varphi_t\nabla f_t^{\top}(I-K_t)B_t^{-1}(\nabla f_t+\be_t) \nonumber\\
&\leq {\red -\frac{\varphi_t}{\Upsilon_{B,t}}\|\nabla f_t\|^2+\frac{\varphi_t}{\gamma_{B,t}}\|K_t\|\|\nabla f_t\|^2+\frac{\varphi_t}{\gamma_{B,t}}(1+\|K_t\|)\|\nabla f_t\|\|\be_t\|} \nonumber\\
&\leq {\red -\frac{\varphi_t}{2\Upsilon_{B,t}}\|\nabla f_t\|^2+\frac{\varphi_t}{\gamma_{B,t}}\left(1+\frac{\gamma_{B,t}}{4\Upsilon_{B,t}}\right)\|\nabla f_t\|\|\be_t\|} \nonumber\\
&\leq {\red -\frac{\varphi_t}{4\Upsilon_{B,t}}\|\nabla f_t\|^2+\frac{\Upsilon_{B,t}\varphi_t}{\gamma_{B,t}^2}\left(1+\frac{\gamma_{B,t}}{4\Upsilon_{B,t}}\right)^2\|\be_t\|^2},
\end{align}
where the last inequality follows from the Young's inequality
\begin{equation*}
{\red \frac{1}{\gamma_{B,t}}\left(1+\frac{\gamma_{B,t}}{4\Upsilon_{B,t}}\right)ab \leq \frac{1}{4\Upsilon_{B,t}}a^2+\frac{\Upsilon_{B,t}}{\gamma_{B,t}^2}\left(1+\frac{\gamma_{B,t}}{4\Upsilon_{B,t}}\right)^2b^2.}
\end{equation*}

To control the second moment of the search direction, we first establish a uniform bound for $\|\tK_t\|$. Each $\tZ_{t,j}$ is an orthogonal projection, and the definitions of the acceleration parameters together with $1\leq\nu_t\leq1/\mu_t$ and $\mu_t\geq\gamma_S$, imply
\begin{equation*}
\|\tZ_{t,j}\|\leq1,\qquad\|I-\tZ_{t,j}\|\leq1,\qquad1\leq\gamma_t=\frac{1}{\sqrt{\mu_t\nu_t}}\leq\frac{1}{\sqrt{\mu_t}}\leq\frac{1}{\sqrt{\gamma_S}}.
\end{equation*}
Consequently,
\begin{align}\label{eq:global-uniform-tK}
\|\tK_t\| &= \left\|\begin{pmatrix}I&\b0\end{pmatrix}\Tilde C_t\begin{pmatrix}I\\I\end{pmatrix}\right\| \leq \sqrt{2}\|\Tilde C_t\| \leq \sqrt{2}\prod_{j=0}^{\tau-1}\|\tC_{t,j}\| \nonumber\\
&\leq \sqrt{2}\prod_{j=0}^{\tau-1}\Big\{\|I-\tZ_{t,j}\|+(1-\alpha_t)(1-\beta_t)+(1-\alpha_t)\gamma_t\|\tZ_{t,j}\|+(\alpha_t+\beta_t-\alpha_t\beta_t)+\alpha_t\gamma_t\|\tZ_{t,j}\|\Big\} \nonumber\\
&\leq \sqrt{2}\prod_{j=0}^{\tau-1}\left(\|I-\tZ_{t,j}\|+1+\gamma_t\|\tZ_{t,j}\|\right) \nonumber\\
&\leq \sqrt{2}\left(2+\frac{1}{\sqrt{\gamma_S}}\right)^\tau\eqqcolon C_K.
\end{align}
By Lemma~\ref{lem:inner_transition} and~\eqref{eq:global-uniform-tK}, we immediately have
\begin{equation}\label{eq:global-direction-second-moment}
\mE[\|\bz_{t,\tau}\|^2\mid\mF_{t-0.5}]\leq{\red \frac{2(1+C_K)^2}{\gamma_{B,t}^2}}\left(\|\nabla f_t\|^2+\|\be_t\|^2\right).
\end{equation}

Substituting~\eqref{eq:global-mean-descent} and~\eqref{eq:global-direction-second-moment} into~\eqref{eq:global-conditional-taylor} yields
\begin{multline*}
\mE[f_{t+1}-f^{\star}\mid\mF_{t-0.5}] \leq f_t-f^{\star}-{\red \left\{\frac{1}{4\Upsilon_{B,t}}-\frac{\Upsilon_H(1+C_K)^2}{\gamma_{B,t}^2}\varphi_t\right\}}\varphi_t\|\nabla f_t\|^2 \\
+ {\red \left\{\frac{\Upsilon_{B,t}}{\gamma_{B,t}^2}\left(1+\frac{\gamma_{B,t}}{4\Upsilon_{B,t}}\right)^2+\frac{\Upsilon_H(1+C_K)^2}{\gamma_{B,t}^2}\varphi_t\right\}}\varphi_t\|\be_t\|^2.
\end{multline*}
{\red Since $\sum_t(\varphi_t/\gamma_{B,t})^2<\infty$ and $\limsup_{t\to\infty}\Upsilon_{B,t}/\gamma_{B,t}<\infty$ almost surely, we have $\varphi_t/\gamma_{B,t}\to0$ and $\varphi_t\Upsilon_{B,t}/\gamma_{B,t}^2\to0$ almost surely. Thus, almost surely for all sufficiently large $t$, $\varphi_t/\gamma_{B,t}\leq1$ and $\Upsilon_H(1+C_K)^2\varphi_t\Upsilon_{B,t}/\gamma_{B,t}^2\leq1/8$.}
For such sufficiently large $t$, we have
\begin{multline}\label{eq:global-objective-recursion}
\mE[f_{t+1}-f^{\star}\mid\mF_{t-0.5}] \leq f_t-f^{\star}-{\red \frac{\varphi_t}{8\Upsilon_{B,t}}}\|\nabla f_t\|^2 \\
+ {\red \left\{\frac{25}{16}+\Upsilon_H(1+C_K)^2\right\}\frac{\varphi_t\Upsilon_{B,t}}{\gamma_{B,t}^2}}\|\be_t\|^2.
\end{multline}

We next control the moving-average error $\be_t$, which admits the recursion
\begin{equation}\label{eq:global-tracking-identity}
\be_{t+1}=(1-\eta_{t+1})(\be_t+\nabla f_t-\nabla f_{t+1})+\eta_{t+1}(\bg_{t+1}-\nabla f_{t+1}).
\end{equation}
Conditional on $\mF_t$, the first term on the right-hand side is fixed. By Assumption~\ref{ass:2}, we have
\begin{equation}\label{eq:global-tracking-second-moment}
\mE[\|\be_{t+1}\|^2\mid\mF_t]=(1-\eta_{t+1})^2\|\be_t+\nabla f_t-\nabla f_{t+1}\|^2+\eta_{t+1}^2\mE[\|\bg_{t+1}-\nabla f_{t+1}\|^2\mid\mF_t].
\end{equation}
For $\eta_{t+1}\in(0,1)$, applying Young's inequality to the cross term gives
\begin{align*}
& (1-\eta_{t+1})^2\|\be_t+\nabla f_t-\nabla f_{t+1}\|^2 \\
& = (1-\eta_{t+1})^2\left\{\|\be_t\|^2+2\be_t^\top(\nabla f_t-\nabla f_{t+1})+\|\nabla f_t-\nabla f_{t+1}\|^2\right\} \\
& \leq (1-\eta_{t+1})^2\left\{\left(1+\frac{\eta_{t+1}}{1-\eta_{t+1}}\right)\|\be_t\|^2+\left(1+\frac{1-\eta_{t+1}}{\eta_{t+1}}\right)\|\nabla f_t-\nabla f_{t+1}\|^2\right\} \\
& = (1-\eta_{t+1})\|\be_t\|^2+\frac{(1-\eta_{t+1})^2}{\eta_{t+1}}\|\nabla f_t-\nabla f_{t+1}\|^2 \\
& \leq (1-\eta_{t+1})\|\be_t\|^2+\eta_{t+1}^{-1}\|\nabla f_t-\nabla f_{t+1}\|^2.
\end{align*}
When $\eta_{t+1}=1$, the left-hand side is zero, so the same inequality holds trivially. Moreover, the $\Upsilon_H$-Lipschitz continuity of $\nabla f(\bx)$ and the outer update imply
\begin{equation*}
\|\nabla f_{t+1}-\nabla f_t\|\leq\Upsilon_H\|\bx_{t+1}-\bx_t\|=\Upsilon_H\varphi_t\|\bz_{t,\tau}\|.
\end{equation*}
Applying these inequalities to~\eqref{eq:global-tracking-second-moment}, taking the conditional expectation with respect to $\mF_{t-0.5}$, and using the tower property and~\eqref{eq:global-direction-second-moment}, we obtain
\begin{multline*}
\mE[\|\be_{t+1}\|^2\mid\mF_{t-0.5}] \leq (1-\eta_{t+1})\|\be_t\|^2+{\red \frac{2\Upsilon_H^2(1+C_K)^2\varphi_t^2}{\gamma_{B,t}^2\eta_{t+1}}}\left(\|\nabla f_t\|^2+\|\be_t\|^2\right) \\
+ \eta_{t+1}^2\mE[\|\bg_{t+1}-\nabla f_{t+1}\|^2\mid\mF_{t-0.5}].
\end{multline*}
By Assumption~\ref{ass:2} with $q_g=2$, we have
\begin{equation*}
\mE[\|\bg_{t+1}-\nabla f_{t+1}\|^2\mid\mF_t]\leq C_{g,1}\|\bx_{t+1}-\bx^{\star}\|^2+C_{g,2}.
\end{equation*}
{\red Since $\varphi_t/\gamma_{B,t}\leq1$ almost surely for all sufficiently large $t$,}~\eqref{eq:global-strong-convexity},~\eqref{eq:global-smoothness}, and~\eqref{eq:global-direction-second-moment} yield
\begin{multline*}
\mE[\|\bx_{t+1}-\bx^{\star}\|^2\mid\mF_{t-0.5}] \leq 2\|\bx_t-\bx^{\star}\|^2+2\varphi_t^2\mE[\|\bz_{t,\tau}\|^2\mid\mF_{t-0.5}] \\
\leq \left\{\frac{4}{\gamma_H}+{\red 8\Upsilon_H(1+C_K)^2}\right\}(f_t-f^{\star})+{\red 4(1+C_K)^2}\|\be_t\|^2.
\end{multline*}
Therefore, by the tower property,
\begin{multline*}
\mE[\|\bg_{t+1}-\nabla f_{t+1}\|^2\mid\mF_{t-0.5}]
\leq C_{g,1}\left\{\frac{4}{\gamma_H}+{\red 8\Upsilon_H(1+C_K)^2}\right\}(f_t-f^{\star}) \\
+{\red 4C_{g,1}(1+C_K)^2}\|\be_t\|^2+C_{g,2}.
\end{multline*}
Substituting this bound into the preliminary tracking recursion yields
\begin{multline}\label{eq:global-tracking-recursion}
\mE[\|\be_{t+1}\|^2\mid\mF_{t-0.5}] \leq (1-\eta_{t+1})\|\be_t\|^2+{\red \frac{2\Upsilon_H^2(1+C_K)^2\varphi_t^2}{\gamma_{B,t}^2\eta_{t+1}}}\left(\|\nabla f_t\|^2+\|\be_t\|^2\right) \\
+ {\red C_{g,1}\left\{\frac{4}{\gamma_H}+8\Upsilon_H(1+C_K)^2\right\}}\eta_{t+1}^2(f_t-f^{\star})+{\red 4C_{g,1}(1+C_K)^2}\eta_{t+1}^2\|\be_t\|^2+C_{g,2}\eta_{t+1}^2.
\end{multline}

{\red Since $\eta_t/\eta_{t+1}\to1$, the theorem's condition gives us $\varphi_t/(\gamma_{B,t}\eta_{t+1})\to0$ almost surely. Together with $\limsup_{t\to\infty}\Upsilon_{B,t}/\gamma_{B,t}<\infty$, this implies that almost surely for all sufficiently large $t$,}
\begin{align*}
{\red 2\Upsilon_H^2(1+C_K)^2\frac{\varphi_t\Upsilon_{B,t}}{\gamma_{B,t}^2\eta_{t+1}}} &{\red \leq \frac1{16}}, \\
{\red \left\{\frac{25}{16}+\Upsilon_H(1+C_K)^2\right\}\frac{\varphi_t\Upsilon_{B,t}}{\gamma_{B,t}^2\eta_{t+1}}} &{\red \leq \frac14}, \\
{\red 2\Upsilon_H^2(1+C_K)^2\left(\frac{\varphi_t}{\gamma_{B,t}\eta_{t+1}}\right)^2} &{\red \leq \frac14}.
\end{align*}
Adding~\eqref{eq:global-objective-recursion} and~\eqref{eq:global-tracking-recursion} and using these three inequalities gives us
\begin{multline}\label{eq:global-joint-recursion}
\mE[f_{t+1}-f^{\star}+\|\be_{t+1}\|^2\mid\mF_{t-0.5}] \\
\leq \left[1+{\red C_{g,1}\left\{\frac{4}{\gamma_H}+8\Upsilon_H(1+C_K)^2+4(1+C_K)^2\right\}}\eta_{t+1}^2\right]\left(f_t-f^{\star}+\|\be_t\|^2\right) \\
{\red -\frac{\varphi_t}{16\Upsilon_{B,t}}\|\nabla f_t\|^2}-\frac{\eta_{t+1}}{2}\|\be_t\|^2+C_{g,2}\eta_{t+1}^2.
\end{multline}
Since $2\eta>1$, we have $\sum_{t=0}^{\infty}\eta_{t+1}^2<\infty$. {\red Let $\mathcal L_t\coloneqq f_t-f^{\star}+\|\be_t\|^2$. For each integer $m\geq1$, choose a deterministic $\delta_m\in(0,1]$ sufficiently small that~\eqref{eq:global-objective-recursion} and the three inequalities preceding~\eqref{eq:global-joint-recursion} hold whenever $\Upsilon_{B,t}/\gamma_{B,t}\leq m$ and $\varphi_t/(\gamma_{B,t}\eta_{t+1})\leq\delta_m$. Such a choice is possible since $\eta_{t+1}\leq1$ and these two bounds imply}
\begin{equation*}
{\red \frac{\varphi_t}{\gamma_{B,t}}\leq\delta_m,\qquad \frac{\varphi_t\Upsilon_{B,t}}{\gamma_{B,t}^2}\leq m\delta_m,\qquad \frac{\varphi_t\Upsilon_{B,t}}{\gamma_{B,t}^2\eta_{t+1}}\leq m\delta_m,\qquad \left(\frac{\varphi_t}{\gamma_{B,t}\eta_{t+1}}\right)^2\leq\delta_m^2.}
\end{equation*}
{\red For any $k\geq0$, we define}
\begin{equation}\label{eq:global-stopping-time}
{\red \sigma_{k,m}\coloneqq\inf\left\{t\geq k:\frac{\Upsilon_{B,t}}{\gamma_{B,t}}>m\;\text{ OR }\;\frac{\varphi_t}{\gamma_{B,t}\eta_{t+1}}>\delta_m\right\},\qquad \inf\varnothing=\infty.}
\end{equation}
{\red Since $\Upsilon_{B,t}/\gamma_{B,t}$ and $\varphi_t/(\gamma_{B,t}\eta_{t+1})$ are $\mF_{t-1}$-measurable, $\sigma_{k,m}$ is a stopping time and $\{\sigma_{k,m}>t\}\in\mF_{t-1}$. Hence,~\eqref{eq:global-joint-recursion} holds on $\{\sigma_{k,m}>t\}$. For any $\ell \geq 1$, $t\geq k$, the stopped process satisfies}
\begin{multline}\label{eq:global-stopped-recursion}
{\red \mE[\mathcal L_{(t+1)\wedge\sigma_{k,m}}\1_{\{\mathcal L_k\leq\ell\}}\mid\mF_{t-0.5}]} \\
{\red \leq\left[1+C_{g,1}\left\{\frac{4}{\gamma_H}+8\Upsilon_H(1+C_K)^2+4(1+C_K)^2\right\}\eta_{t+1}^2\right]\mathcal L_{t\wedge\sigma_{k,m}}\1_{\{\mathcal L_k\leq\ell\}}} \\
{\red -\left\{\frac{\varphi_t}{16\Upsilon_{B,t}}\|\nabla f_t\|^2+\frac{\eta_{t+1}}{2}\|\be_t\|^2\right\}\1_{\{\sigma_{k,m}>t\}}\1_{\{\mathcal L_k\leq\ell\}}+C_{g,2}\eta_{t+1}^2\1_{\{\mathcal L_k\leq\ell\}},}
\end{multline}
{\red where $a\wedge b\coloneqq\min\{a,b\}$. Since $\mathcal L_k\1_{\{\mathcal L_k\leq\ell\}}\leq\ell$, applying the Robbins--Siegmund theorem~\citep[Theorem~1.3.12]{Duflo2013Random} to~\eqref{eq:global-stopped-recursion} gives $\lim_{t\to\infty} \mathcal L_{t\wedge\sigma_{k,m}} \1_{\{\mathcal L_k\leq\ell\}} \in [0,\infty)$,
\begin{equation*}
\sum_{t=k}^{\infty} \frac{\varphi_t}{\Upsilon_{B,t}}\|\nabla f_t\|^2\1_{\{\sigma_{k,m}>t\}} \1_{\{\mathcal L_k\leq\ell\}} < \infty,
\quad
\sum_{t=k}^{\infty} \eta_{t+1}\|\be_t\|^2\1_{\{\sigma_{k,m}>t\}}\1_{\{\mathcal L_k\leq\ell\}} <\infty
\end{equation*}
almost surely. By Assumption~\ref{ass:1}, $\limsup_{t\to\infty}\Upsilon_{B,t}/\gamma_{B,t}<\infty$ almost surely. For each such realization, choose an integer $m>\limsup_{t\to\infty}\Upsilon_{B,t}/\gamma_{B,t}$. Since $\varphi_t/(\gamma_{B,t}\eta_{t+1})\to0$ and $\delta_m>0$, we can then choose $k$ such that $\sigma_{k,m}=\infty$. Finally, $\mathcal L_k<\infty$ allows us to choose an integer $\ell\geq\mathcal L_k$. As a result, we have}
\begin{equation*}
{\red \PP\left(\bigcup_{k\geq0}\bigcup_{m,\ell\geq1}\{\sigma_{k,m}=\infty,\ \mathcal L_k\leq\ell\}\right)=1.}
\end{equation*}
{\red Since $(k,m,\ell)$ ranges over a countable set, the three conclusions obtained from the Robbins--Siegmund theorem hold almost surely for all $(k,m,\ell)$. For almost every realization, choose $(k,m,\ell)$ such that $\sigma_{k,m}=\infty$ and $\mathcal L_k\leq\ell$. For this choice, $t\wedge\sigma_{k,m}=t$ and $\1_{\{\sigma_{k,m}>t\}}=\1_{\{\mathcal L_k\leq\ell\}}=1$ for every $t\geq k$. Therefore, $\mathcal L_t$ converges to a finite random variable, and both sums over $t \geq k$ are finite. Adding the finitely many terms with $t<k$, we obtain}
\begin{equation}\label{eq:global-weighted-sums}
{\red \sum_{t=0}^{\infty}\frac{\varphi_t}{\Upsilon_{B,t}}\|\nabla f_t\|^2<\infty,}
\qquad
\sum_{t=0}^{\infty}\eta_{t+1}\|\be_t\|^2<\infty
\end{equation}
almost surely.
In particular, $f_t-f^{\star}$ and $\|\be_t\|^2$ are bounded almost surely. Moreover,
\begin{equation*}
{\red 2\Upsilon_H^2(1+C_K)^2\left(\frac{\varphi_t}{\gamma_{B,t}\eta_{t+1}}\right)^2+4C_{g,1}(1+C_K)^2\eta_{t+1}\rightarrow0.}
\end{equation*}
Hence, for all sufficiently large $t$,~\eqref{eq:global-tracking-recursion} implies
\begin{multline}\label{eq:global-refined-tracking}
\mE[\|\be_{t+1}\|^2\mid\mF_{t-0.5}] \leq \left(1-\frac{\eta_{t+1}}{2}\right)\|\be_t\|^2+{\red \frac{2\Upsilon_H^2(1+C_K)^2\varphi_t^2}{\gamma_{B,t}^2\eta_{t+1}}}\|\nabla f_t\|^2 \\
+ {\red C_{g,1}\left\{\frac{4}{\gamma_H}+8\Upsilon_H(1+C_K)^2\right\}}\eta_{t+1}^2(f_t-f^{\star})+C_{g,2}\eta_{t+1}^2.
\end{multline}
The positive terms on the right-hand side of~\eqref{eq:global-refined-tracking} are almost surely summable. Indeed, by~\eqref{eq:global-weighted-sums},
\begin{equation*}
{\red 2\Upsilon_H^2(1+C_K)^2\sum_{t=0}^{\infty}\frac{\varphi_t^2}{\gamma_{B,t}^2\eta_{t+1}}\|\nabla f_t\|^2 = 2\Upsilon_H^2(1+C_K)^2\sum_{t=0}^{\infty}\left(\frac{\Upsilon_{B,t}}{\gamma_{B,t}}\frac{\varphi_t}{\gamma_{B,t}\eta_{t+1}}\right)\left(\frac{\varphi_t}{\Upsilon_{B,t}}\|\nabla f_t\|^2\right)<\infty}
\end{equation*}
almost surely. Furthermore, the almost-sure boundedness of $f_t-f^{\star}$ and $\sum_t\eta_{t+1}^2<\infty$ also imply
\begin{equation*}
\sum_{t=0}^{\infty} \eta_{t+1}^2\left[C_{g,1}\left\{\frac{4}{\gamma_H}+{\red 8\Upsilon_H(1+C_K)^2}\right\}(f_t-f^{\star})+C_{g,2} \right] < \infty
\end{equation*}
almost surely. {\red Applying Robbins--Siegmund again to the same stopped processes and then removing the localization, we conclude that $\|\be_t\|^2$ converges almost surely; the positive terms are summable by the preceding two bounds. Moreover, $\sum_{t=0}^{\infty}\eta_{t+1}=\infty$. Otherwise, $\varphi_t/(\gamma_{B,t}\eta_{t+1})\to0$ would give $\varphi_t/\gamma_{B,t}\leq\eta_{t+1}$ eventually, contradicting $\sum_t\varphi_t/\gamma_{B,t}=\infty$ in Assumption~\ref{ass:1}. Together with $\sum_{t=0}^{\infty}\eta_{t+1}\|\be_t\|^2<\infty$, this gives us}
\begin{equation*}
\be_t=\bbm_t-\nabla f_t\stackrel{a.s.}{\rightarrow}\b0.
\end{equation*}

Finally, the almost-sure convergence of $f_t-f^{\star}+\|\be_t\|^2$ and $\|\be_t\|^2 \rightarrow 0$ imply that $f_t-f^{\star}$ also converges almost surely. Since ${\red \sum_{t=0}^{\infty}\varphi_t/\Upsilon_{B,t}=\infty}$ and ${\red \sum_{t=0}^{\infty}(\varphi_t/\Upsilon_{B,t})\|\nabla f_t\|^2<\infty}$, we have
\begin{equation*}
\liminf_{t\rightarrow\infty}\|\nabla f_t\|=0
\qquad\text{almost surely}.
\end{equation*}
It follows from~\eqref{eq:global-strong-convexity} that $\liminf_{t\rightarrow\infty}(f_t-f^{\star})=0$ almost surely. Given that $f_t-f^{\star}$ converges almost surely, we conclude that $f_t-f^{\star} \rightarrow 0$ almost surely. Again, we can apply the lower bound in~\eqref{eq:global-strong-convexity} and obtain
\begin{equation*}
\bx_t\stackrel{a.s.}{\rightarrow}\bx^{\star},
\end{equation*}
which completes the proof.

\subsection{Proof of Lemma~\ref{lem:inner_transition}}\label{app:inner-transition}

Fix an outer iteration $t$. In Section~\ref{sec:preliminaries}, we temporarily suppress the outer iteration index. Below, we write $\{\bz_{t,j},\bv_{t,j},\by_{t,j},\bw_{t,j}\}_{j=0}^{\tau}$ to make the dependence on $t$ explicit. Recall that $B_t\Delta\bx_t=-\bbm_t$. It follows from the definition of $\bw_{t,j}$ in Algorithm~\ref{alg:nasketch} that
\begin{equation}\label{eq:inner-correction-error}
\bw_{t,j}
= B_tS_{t,j}(S_{t,j}^\top B_t^2S_{t,j})^\dagger
S_{t,j}^\top B_t(\by_{t,j}-\Delta\bx_t)
= \tZ_{t,j}(\by_{t,j}-\Delta\bx_t).
\end{equation}
Consequently, by Algorithm~\ref{alg:nasketch},
\begin{align*}
\bz_{t,j+1}-\Delta\bx_t
&= \by_{t,j}-\Delta\bx_t-\bw_{t,j}\\
&= (I-\tZ_{t,j}) \left\{(1-\alpha_t)(\bz_{t,j}-\Delta\bx_t) + \alpha_t(\bv_{t,j}-\Delta\bx_t)\right\},
\end{align*}
and
\begin{align*}
\bv_{t,j+1}-\Delta\bx_t
&= \beta_t(\bv_{t,j}-\Delta\bx_t) + (1-\beta_t)(\by_{t,j}-\Delta\bx_t) - \gamma_t\bw_{t,j} \\
&= \left\{(1-\alpha_t)(1-\beta_t)I - (1-\alpha_t)\gamma_t\tZ_{t,j}\right\}
 (\bz_{t,j}-\Delta\bx_t) \\
& \quad + \left\{(\alpha_t+\beta_t-\alpha_t\beta_t)I - \alpha_t\gamma_t\tZ_{t,j}\right\} (\bv_{t,j}-\Delta\bx_t).
\end{align*}
Combining the two displays, we obtain the block recursion
\begin{equation}\label{eq:inner-block-transition}
\begin{pmatrix}
\bz_{t,j+1}-\Delta\bx_t \\
\bv_{t,j+1}-\Delta\bx_t
\end{pmatrix}
=
\tC_{t,j}
\begin{pmatrix}
\bz_{t,j}-\Delta\bx_t \\
\bv_{t,j}-\Delta\bx_t
\end{pmatrix},
\end{equation}
where $\tC_{t,j}$ is defined in~\eqref{equ:Cttilde}. Iterating~\eqref{eq:inner-block-transition} yields
\begin{equation}\label{eq:inner-block-product}
\begin{pmatrix}
\bz_{t,\tau}-\Delta\bx_t \\
\bv_{t,\tau}-\Delta\bx_t
\end{pmatrix}
=
\tC_{t,\tau-1}\cdots\tC_{t,0}
\begin{pmatrix}
\bz_{t,0}-\Delta\bx_t \\
\bv_{t,0}-\Delta\bx_t
\end{pmatrix}.
\end{equation}
Since $\bz_{t,0}=\bv_{t,0}=\b0$, $\Tilde C_t = \prod_{j=0}^{\tau-1} \tC_{t,j}$, taking the first block row in~\eqref{eq:inner-block-product} and using~\eqref{def:tilde_K_t} gives
\begin{equation*}
\bz_{t,\tau}-\Delta\bx_t = -\tK_t\Delta\bx_t = \tK_t(\bz_{t,0}-\Delta\bx_t),
\end{equation*}
or, equivalently,
\begin{equation}\label{eq:inner-direction-representation}
\bz_{t,\tau} = (I-\tK_t)\Delta\bx_t = -(I-\tK_t)B_t^{-1}\bbm_t.
\end{equation}
Conditional on $\mF_{t-0.5}$, the quantities $(B_t,\bbm_t,\alpha_t,\beta_t,\gamma_t)$ are fixed, and the only randomness in~\eqref{eq:inner-direction-representation} comes from the sketching matrices. By Assumption~\ref{ass:4}, we have
\begin{equation*}
\mE[\tK_t\mid\mF_{t-0.5}] = \mE[\tK_t\mid\mF_{t-1}] = K_t.  
\end{equation*}
Consequently,
\begin{equation}\label{eq:inner-conditional-response}
\mE[\bz_{t,\tau}\mid\mF_{t-0.5}] = -(I-K_t)B_t^{-1}\bbm_t.
\end{equation}
Moreover,
\begin{equation*}
\mE[\bz_{t,\tau}-(I-K_t)\Delta\bx_t\mid\mF_{t-0.5}]=\b0.
\end{equation*}
By the tower property,
\begin{equation*}
\mE[\bz_{t,\tau}-(I-K_t)\Delta\bx_t\mid\mF_{t-1}] = \mE[\mE[\bz_{t,\tau}-(I-K_t)\Delta\bx_t\mid\mF_{t-0.5}] \mid \mF_{t-1}] = \b0,
\end{equation*}
which proves the martingale-difference statement with respect to $\{\mF_t\}_{t\geq0}$.

Finally, suppose that $\gamma_t=1$. For any $Y\in\mR^{d\times d}$, direct multiplication in~\eqref{equ:Cttilde} gives
\begin{equation}\label{eq:unaccelerated-block-action}
\tC_{t,j} \left\{\begin{pmatrix}1\\1\end{pmatrix}\otimes Y\right\} = \begin{pmatrix}1\\1\end{pmatrix} \otimes \{(I-\tZ_{t,j})Y\}.
\end{equation}
By the above property, we then obtain 
\begin{align*}
\Tilde C_t \begin{pmatrix}
I_d \\
I_d
\end{pmatrix} &= \tC_{t, \tau-1} \tC_{t, \tau-2} \cdots \tC_{t, 0} \rbr{\begin{pmatrix}
1 \\ 1
\end{pmatrix} \otimes I_d} = \begin{pmatrix}
1 \\ 1
\end{pmatrix} \otimes \rbr{\prod_{j = 0}^{\tau - 1} (I_d - \tZ_{t, j})}.
\end{align*}
Applying~\eqref{eq:unaccelerated-block-action} successively and then using~\eqref{def:tilde_K_t}, we obtain
\begin{align*}
\tK_t = \begin{pmatrix}I&\zero\end{pmatrix} \Tilde C_t \begin{pmatrix}I\\I\end{pmatrix} = (I-\tZ_{t,\tau-1})\cdots(I-\tZ_{t,0}).
\end{align*}
This completes the proof.

\section{Proofs of Section~\ref{sec:normality}}\label{app:normality}

Throughout this section, let $\bDelta_t\coloneqq\bx_t-\bx^\star$, $\mathcal R\coloneqq I-K^\star$ and $\lambda_m\coloneqq\lambda_{\min}(\mathcal R)$. We use $a_t=O(b_t)$ and $a_t\lesssim b_t$ interchangeably to simplify notation, where the implicit constants are deterministic and independent of $t$. We also use $\1_{\{\cdot\}}$ to denote the indicator function.
{\red In the proofs below, $B_t$ is clipped at every iteration as in~\eqref{eq:clipped-hessian-average}. The periodically refreshed implementation can be treated by the same arguments: for fixed $d$, its latest refresh differs from $t$ by at most $d-1$ iterations, so the corresponding Hessian average has the same almost-sure limit and $O_p(\sqrt{\eta_t})$ local rate.}

\subsection{Proof of Lemma~\ref{lem:operator-stabilization}}\label{app:operator-stabilization}

\noindent $\bullet$ \textbf{Convergence of $B_t$.}
{\red By the construction of $\tilde{B}_t$, we have}
\begin{align}\label{eq:operator-B-decomposition}
\|{\red \tilde{B}_t}-B^\star\| &= \left\|\frac{1}{t}\sum_{i=0}^{t-1}(H_i-\nabla^2f_i)+\frac{1}{t}\sum_{i=0}^{t-1}(\nabla^2f_i-B^\star)\right\| \nonumber\\
&\leq \left\|\frac{1}{t}\sum_{i=0}^{t-1}(H_i-\nabla^2f_i)\right\|+\left\|\frac{1}{t}\sum_{i=0}^{t-1}(\nabla^2f_i-B^\star)\right\| \nonumber\\
&\leq \left\|\frac{1}{t}\sum_{i=0}^{t-1}(H_i-\nabla^2f_i)\right\|+\frac{\Upsilon_L}{t}\sum_{i=0}^{t-1}\|\bx_i-\bx^\star\|.
\end{align}
By Theorem~\ref{thm:global-convergence}, $\bx_i\stackrel{a.s.}{\rightarrow}\bx^\star$. Hence, the second term on the right-hand side of~\eqref{eq:operator-B-decomposition} converges to zero almost surely by Lemma~\ref{lem:stolz-cesaro}.
For the first term, $\{H_i-\nabla^2f_i\}_{i\geq0}$ is a matrix-valued martingale difference sequence since $\mE[H_i\mid\mF_{i-1}]=\nabla^2f_i$.
{\red For each integer $m\geq1$, $\1_{\{\|\bx_i-\bx^\star\|\leq m\}}$ is $\mF_{i-1}$-measurable. By Assumption~\ref{ass:3}, $\|A\|_{\mathrm F}^2\leq d\|A\|^2$ and martingale orthogonality, we have
\begin{multline*}
\mE\left[\left\|\sum_{i=0}^{t-1}\frac{(H_i-\nabla^2f_i)\1_{\{\|\bx_i-\bx^\star\|\leq m\}}}{i+1}\right\|_{\mathrm F}^2\right] = \sum_{i=0}^{t-1}\frac{\mE[\|H_i-\nabla^2f_i\|_{\mathrm F}^2\1_{\{\|\bx_i-\bx^\star\|\leq m\}}]}{(i+1)^2} \\
\leq d(C_{H,1}m^2+C_{H,2})\sum_{i=0}^{\infty}\frac{1}{(i+1)^2}<\infty.    
\end{multline*}
where the equality follows from the orthogonality of the martingale differences under the Frobenius inner product. Hence, for every $m\geq1$, $\{\sum_{i=0}^{t-1}(H_i-\nabla^2f_i)\1_{\{\|\bx_i-\bx^\star\|\leq m\}}/(i+1)\}_{t\geq1}$ is an $L^2$-bounded martingale. By Doob's martingale convergence theorem \citep[Corollary~2.2]{Hall2014Martingale}, $\sum_{i=0}^{t-1}(H_i-\nabla^2f_i)\1_{\{\|\bx_i-\bx^\star\|\leq m\}}/(i+1)$ converges almost surely.

Since $\bx_i\stackrel{a.s.}{\rightarrow}\bx^\star$ and the positive integers are countable, almost surely $\sup_{i\geq0}\|\bx_i-\bx^\star\|<\infty$ and the truncated weighted sum converges for every $m\geq1$. For each such realization, we could choose an integer $m>\sup_{i\geq0}\|\bx_i-\bx^\star\|$. Then $\1_{\{\|\bx_i-\bx^\star\|\leq m\}}=1$ for every $i\geq0$, so $\sum_{i=0}^{t-1}(H_i-\nabla^2f_i)/(i+1)$ converges. Applying Lemma~\ref{lem:kronecker} further yields}
\begin{equation*}
\frac{1}{t}\sum_{i=0}^{t-1}(H_i-\nabla^2f_i)\stackrel{a.s.}{\rightarrow}\zero.
\end{equation*}
{\red Combining the preceding conclusion with~\eqref{eq:operator-B-decomposition}, we obtain $\tilde{B}_t\stackrel{a.s.}{\rightarrow}B^\star$. Since $B^\star$ belongs to the clipping interval for all sufficiently large $t$ almost surely, the projection property of $\operatorname{Clip}(\cdot)$ gives us}
\begin{equation*}
{\red \|B_t-B^\star\|\leq\|B_t-B^\star\|_{\mathrm F}\leq\|\tilde{B}_t-B^\star\|_{\mathrm F}\leq\sqrt d\,\|\tilde{B}_t-B^\star\|\stackrel{a.s.}{\rightarrow}0.}
\end{equation*}
{\red Therefore,}
\begin{equation}\label{eq:operator-B-convergence}
B_t\stackrel{a.s.}{\rightarrow}B^\star.
\end{equation}

\noindent $\bullet$ \textbf{Convergence of $(\alpha_t,\beta_t,\gamma_t)$.}
Recall from~\eqref{eq:acceleration-parameters} that the acceleration parameters $(\alpha_t, \beta_t, \gamma_t)$ (the distinction between $(\alpha_t, \beta_t, \gamma_t)$ and $(\alpha^{\star}, \beta^{\star}, \gamma^{\star})$ is solely due to the evaluation at $(\mu_t,\nu_t)$ v.s. $(\mu^{\star},\nu^{\star})$) can be regarded as functions of $(\mu_t,\nu_t)$ as follows:
\begin{equation*}
\alpha(\mu_t,\nu_t) = \frac{1}{1+\sqrt{\nu_t/\mu_t}},
\qquad
\beta(\mu_t,\nu_t)=1-\sqrt{\frac{\mu_t}{\nu_t}},
\qquad
\gamma(\mu_t,\nu_t)=\frac{1}{\sqrt{\mu_t\nu_t}}.
\end{equation*}
Each of these functions is continuously differentiable on $(0,\infty)^2$. Since $(\mu_t,\nu_t)\in[\gamma_S,1]\times[1,1/\gamma_S]$ for all $t\geq0$, their gradients are uniformly bounded on this compact and convex set. Therefore, by the mean-value theorem, we further have
\begin{equation}\label{eq:operator-parameter-MVT}
|\alpha_t-\alpha^\star|+|\beta_t-\beta^\star|+|\gamma_t-\gamma^\star| \lesssim |\mu_t-\mu^\star|+|\nu_t-\nu^\star|.
\end{equation}
Thus, it suffices to study the convergence of $(\mu_t,\nu_t)$.
For a generic sketching matrix $S$, define
\begin{equation}
\begin{aligned}\label{eq:operator-limiting-projections}
\tZ_t(S) &\coloneqq B_tS(S^\top B_t^2S)^\dagger S^\top B_t,
&\qquad
Z_t &\coloneqq \mE[\tZ_t(S)\mid\mF_{t-1}],\\
\tZ^\star(S) &\coloneqq B^\star S(S^\top(B^\star)^2S)^\dagger S^\top B^\star,
&\qquad
Z^\star &\coloneqq \mE[\tZ^\star(S)].
\end{aligned}
\end{equation}
We first recall the following perturbation result for the projection matrix.

\begin{lemma}\citep[Lemma~5.2]{Na2025Statistical}
\label{lem:continuity-projection-matrix}
Suppose $B_t, B^\star\in\mR^{d\times d}$ are both nonsingular. Then, for any sketching matrix $S\in\mR^{d\times s}$, we have
\begin{equation*}
\|\tZ_t(S)-\tZ^\star(S)\| \leq \frac{2\|B_t-B^\star\|}{\sigma_{\min}(B^\star)}\|S\|\|S^\dagger\|,
\end{equation*}
where $\sigma_{\min}(\cdot)$ denotes the smallest singular value.
\end{lemma}

By Lemma~\ref{lem:continuity-projection-matrix}, Assumption~\ref{ass:4}, and the independence between $S$ and $\mF_{t-1}$, we have
\begin{equation}\label{eq:operator-projection-bound}
\mE[\|\tZ_t(S)-\tZ^\star(S)\|\mid\mF_{t-1}] \leq \frac{2\|B_t-B^\star\|}{\gamma_H}\mE[\|S\|\|S^\dagger\|] \lesssim \|B_t-B^\star\|.
\end{equation}

\noindent $\bullet\bullet$ \textbf{Convergence of $\mu_t$.}
By the definitions of $\mu_t$ and $\mu^\star$, we have $\mu_t=\lambda_{\min}(Z_t)$ and $\mu^\star=\lambda_{\min}(Z^\star)$. Applying Weyl's perturbation theorem \citep[Corollary~III.2.6]{Bhatia1997Matrix}, we obtain
\begin{multline}\label{eq:operator-mu-bound}
|\mu_t-\mu^\star| \leq \|Z_t-Z^\star\| = \left\|\mE[\tZ_t(S)-\tZ^\star(S)\mid\mF_{t-1}]\right\| \\
\leq \mE[\|\tZ_t(S)-\tZ^\star(S)\|\mid\mF_{t-1}] \lesssim \|B_t-B^\star\|.
\end{multline}
Since we have shown $B_t\stackrel{a.s.}{\rightarrow}B^\star$, it follows that $\mu_t\stackrel{a.s.}{\rightarrow}\mu^\star$.

\noindent $\bullet\bullet$ \textbf{Convergence of $\nu_t$.}
Define
\begin{equation*}
A_t \coloneqq \mE[\tZ_t(S)Z_t^{-1}\tZ_t(S)\mid\mF_{t-1}],
\qquad
A^\star \coloneqq \mE[\tZ^\star(S)(Z^\star)^{-1}\tZ^\star(S)].
\end{equation*}
The matrices $A_t$ and $A^\star$ are symmetric positive semidefinite. Letting $\bb=Z_t^{1/2}\ba$ in~\eqref{eq:mu-nu}, we obtain
\begin{multline}\label{eq:operator-nu-variational}
\nu_t = \sup_{\ba\neq\zero}\frac{\ba^\top A_t\ba}{\ba^\top Z_t\ba} = \sup_{\bb\neq\zero}\frac{\bb^\top Z_t^{-1/2}A_tZ_t^{-1/2}\bb}{\bb^\top\bb} \\
= \lambda_{\max}(Z_t^{-1/2}A_tZ_t^{-1/2}) = \|Z_t^{-1/2}A_tZ_t^{-1/2}\|.
\end{multline}
Similarly, we have $\nu^\star = \|(Z^\star)^{-1/2}A^\star(Z^\star)^{-1/2}\|$.
Consequently,
\begin{multline}\label{eq:operator-nu-decomposition}
|\nu_t-\nu^\star| = \left|\|Z_t^{-1/2} A_t Z_t^{-1/2}\| - \|(Z^{\star})^{-1/2} A^{\star} (Z^{\star})^{-1/2}\|\right| \\
\leq \|Z_t^{-1/2}A_tZ_t^{-1/2}-(Z^\star)^{-1/2}A^\star(Z^\star)^{-1/2}\| \leq \|(Z_t^{-1/2}-(Z^\star)^{-1/2})A_tZ_t^{-1/2}\| \\
+ \|(Z^\star)^{-1/2}(A_t-A^\star)Z_t^{-1/2}\| + \|(Z^\star)^{-1/2}A^\star(Z_t^{-1/2}-(Z^\star)^{-1/2})\|.
\end{multline}
To control the three terms on the right-hand side, we use the following preparation lemma.

\begin{lemma}\label{lem:matrix-squareroot-lipschitz}
Under Assumption~\ref{ass:4}, we have
\begin{equation*}
\|Z_t^{-1}-(Z^\star)^{-1}\| \leq \frac{1}{\gamma_S^2}\|Z_t-Z^\star\|,
\qquad
\|Z_t^{-1/2}-(Z^\star)^{-1/2}\| \leq \frac{1}{2\gamma_S^{3/2}}\|Z_t-Z^\star\|.
\end{equation*}
\end{lemma}

\noindent $\bullet\bullet\bullet$ \textbf{The first term in~\eqref{eq:operator-nu-decomposition}.}
Since $\tZ_t(S)$ is an orthogonal projection and $Z_t\succeq\gamma_SI$, we have $\|A_t\|\leq1/\gamma_S$ and $\|Z_t^{-1/2}\|\leq1/\sqrt{\gamma_S}$. Therefore, Lemma~\ref{lem:matrix-squareroot-lipschitz} implies that
\begin{equation}\label{eq:operator-nu-first-term}
\|(Z_t^{-1/2}-(Z^\star)^{-1/2})A_tZ_t^{-1/2}\| \lesssim \|Z_t-Z^\star\|.
\end{equation}

\noindent $\bullet\bullet\bullet$ \textbf{The third term in~\eqref{eq:operator-nu-decomposition}.}
Similarly, $\|A^\star\|\leq1/\gamma_S$ and $\|(Z^\star)^{-1/2}\|\leq1/\sqrt{\gamma_S}$. Therefore,
\begin{equation}\label{eq:operator-nu-third-term}
\|(Z^\star)^{-1/2}A^\star(Z_t^{-1/2}-(Z^\star)^{-1/2})\| \lesssim \|Z_t-Z^\star\|.
\end{equation}

\noindent $\bullet\bullet\bullet$ \textbf{The second term in~\eqref{eq:operator-nu-decomposition}.}
Using $\|(Z^\star)^{-1/2}\|\leq\gamma_S^{-1/2}$ and $\|Z_t^{-1/2}\|\leq\gamma_S^{-1/2}$, we have
\begin{align}\label{eq:operator-nu-second-term}
&\|(Z^\star)^{-1/2}(A_t-A^\star)Z_t^{-1/2}\| \leq \frac{1}{\gamma_S}\|A_t-A^\star\| \nonumber\\
&= \frac{1}{\gamma_S}\left\|\mE[\tZ_t(S)Z_t^{-1}\tZ_t(S)\mid\mF_{t-1}]-\mE[\tZ^\star(S)(Z^\star)^{-1}\tZ^\star(S)]\right\| \nonumber\\
&= \frac{1}{\gamma_S}\left\|\mE[\tZ_t(S)Z_t^{-1}\tZ_t(S)-\tZ^\star(S)(Z^\star)^{-1}\tZ^\star(S)\mid\mF_{t-1}]\right\| \nonumber\\
&= \frac{1}{\gamma_S}\left\|\mE\left[
(\tZ_t(S)-\tZ^\star(S))Z_t^{-1}\tZ_t(S)
+\tZ^\star(S)\{Z_t^{-1}-(Z^\star)^{-1}\}\tZ_t(S)\right.\right. \nonumber\\
&\hspace{5.3cm}\left.\left. +\tZ^\star(S)(Z^\star)^{-1}(\tZ_t(S)-\tZ^\star(S))\mid\mF_{t-1}\right]\right\| \nonumber\\
&\leq \frac{1}{\gamma_S}\left\{\frac{1}{\gamma_S}\mE[\|\tZ_t(S)-\tZ^\star(S)\|\mid\mF_{t-1}]+\|Z_t^{-1}-(Z^\star)^{-1}\|\right. \nonumber\\
&\hspace{5.3cm}\left. +\frac{1}{\gamma_S}\mE[\|\tZ_t(S)-\tZ^\star(S)\|\mid\mF_{t-1}]\right\} \nonumber\\
&\leq \frac{1}{\gamma_S^3}\|Z_t-Z^\star\| + \frac{2}{\gamma_S^2}\mE[\|\tZ_t(S)-\tZ^\star(S)\|\mid\mF_{t-1}].
\end{align}
Substituting~\eqref{eq:operator-nu-first-term}, \eqref{eq:operator-nu-third-term}, and~\eqref{eq:operator-nu-second-term} into~\eqref{eq:operator-nu-decomposition}, we obtain
\begin{equation}\label{eq:operator-nu-bound}
|\nu_t-\nu^\star| \lesssim \|Z_t-Z^\star\| + \mE[\|\tZ_t(S)-\tZ^\star(S)\|\mid\mF_{t-1}] \lesssim \|B_t-B^\star\|,
\end{equation}
where the last inequality follows from~\eqref{eq:operator-projection-bound}. Since $B_t\stackrel{a.s.}{\rightarrow}B^\star$, we can conclude that $\nu_t\stackrel{a.s.}{\rightarrow}\nu^\star$.

Combining~\eqref{eq:operator-parameter-MVT}, \eqref{eq:operator-mu-bound}, and~\eqref{eq:operator-nu-bound}, we obtain
\begin{equation}\label{eq:operator-parameter-bound}
|\alpha_t-\alpha^\star|+|\beta_t-\beta^\star|+|\gamma_t-\gamma^\star| \lesssim \|B_t-B^\star\|.
\end{equation}
Together with~\eqref{eq:operator-B-convergence}, this further implies
\begin{equation*}
(\alpha_t,\beta_t,\gamma_t) \stackrel{a.s.}{\rightarrow} (\alpha^\star,\beta^\star,\gamma^\star).
\end{equation*}

\noindent $\bullet$ \textbf{Convergence of $K_t$.}
Recall the constructions of $K_t$ and $K^\star$ from $C_t$ and $C^\star$ in~\eqref{def:tilde_K_t}. By the conditional independence of the sketching matrices, we have
\begin{equation*}
C_t=M_t^\tau, \qquad C^\star=(M^\star)^\tau,
\end{equation*}
where $M_t$ is defined in~\eqref{eq:mean-inner-transition}, and $M^\star$ is defined analogously by replacing $(\alpha_t,\beta_t,\gamma_t,Z_t)$ with $(\alpha^\star,\beta^\star,\gamma^\star,Z^\star)$. From the block representation of $M_t$, we have
\begin{align}
\|M_t-M^\star\| &\leq \|(1-\alpha_t)(I-Z_t)-(1-\alpha^\star)(I-Z^\star)\| + \|\alpha_t(I-Z_t)-\alpha^\star(I-Z^\star)\| \nonumber\\
&\quad + \|(1-\alpha_t)(1-\beta_t)I-(1-\alpha_t)\gamma_tZ_t-(1-\alpha^\star)(1-\beta^\star)I+(1-\alpha^\star)\gamma^\star Z^\star\| \nonumber\\
&\quad + \|(\alpha_t+\beta_t-\alpha_t\beta_t)I-\alpha_t\gamma_tZ_t \nonumber-(\alpha^\star+\beta^\star-\alpha^\star\beta^\star)I+\alpha^\star\gamma^\star Z^\star\| \nonumber\\
&\lesssim|\alpha_t-\alpha^\star|+|\beta_t-\beta^\star|+|\gamma_t-\gamma^\star|+\|Z_t-Z^\star\| \lesssim\|B_t-B^\star\|,
\label{eq:operator-M-bound}
\end{align}
where the last inequality follows from~\eqref{eq:operator-projection-bound} and~\eqref{eq:operator-parameter-bound}.
Furthermore, using $\|I-Z_t\|\leq1$, $\|Z_t\|\leq1$, $0<\alpha_t<1$, $0\leq\beta_t<1$, and $\gamma_t\leq1/\sqrt{\gamma_S}$, we obtain
\begin{multline*}
\|M_t\| \leq \|I-Z_t\|+(1-\alpha_t)(1-\beta_t)+(1-\alpha_t)\gamma_t\|Z_t\| \\
+ (\alpha_t+\beta_t-\alpha_t\beta_t)+\alpha_t\gamma_t\|Z_t\| \leq 2+\frac{1}{\sqrt{\gamma_S}}.
\end{multline*}
The same bound also holds for $\|M^\star\|$. Therefore, a telescoping expansion gives us
\begin{align}\label{eq:operator-C-bound}
\|C_t-C^\star\| &= \|M_t^\tau-(M^\star)^\tau\| \nonumber\\
&= \left\|\sum_{k=0}^{\tau-1}M_t^{\tau-1-k}(M_t-M^\star)(M^\star)^k\right\| \nonumber\\
&\leq \left(\sum_{k=0}^{\tau-1}\|M_t\|^{\tau-1-k}\|M^\star\|^k\right)\|M_t-M^\star\| \nonumber\\
&\lesssim \|M_t-M^\star\| \lesssim \|B_t-B^\star\|.
\end{align}
Finally, by the definitions of $K_t$ and $K^\star$, there exists a deterministic constant $L_K>0$ such that
\begin{equation}\label{eq:operator-K-bound}
\|K_t-K^\star\| = \left\|\begin{pmatrix}I&\zero\end{pmatrix}(C_t-C^\star)\begin{pmatrix}I\\I\end{pmatrix}\right\| \leq \sqrt2\|C_t-C^\star\| \leq L_K\|B_t-B^\star\|.
\end{equation}
Since $B_t\stackrel{a.s.}{\rightarrow}B^\star$, we conclude that
\begin{equation*}
K_t\stackrel{a.s.}{\rightarrow}K^\star.
\end{equation*}

\noindent $\bullet$ \textbf{Symmetry and bound of $K^\star$.}
Recall from the proof of \citet[Lemma~3.7]{Wang2026Inference} that $U_t^\top K_t U_t=\operatorname{diag}\left(p_{t,\tau}(z_{t,1}),\ldots,p_{t,\tau}(z_{t,d})\right)$, implying $K_t$ is symmetric. Since $K_t\stackrel{a.s.}{\rightarrow}K^\star$, $K^\star$ is symmetric.

If $\mu_t=1$, \citet[Lemma~3.7]{Wang2026Inference} gives $K_t=\zero$. If $\mu_t<1$, \citet[Lemma~3.7]{Wang2026Inference} and the condition on $\tau$ give
\begin{equation*}
\|K_t\| \leq 2\tau\left(1-\sqrt{\frac{\mu_t}{\nu_t}}\right)^{\tau-2} \leq {\red \frac{\gamma_{B,t}}{4\Upsilon_{B,t}}}.
\end{equation*}
Therefore, $\|K_t\|\leq{\red 1/4}$ for every $t \geq 0$. Taking the almost-sure limit yields
\begin{equation*}
\|K^\star\| \leq {\red \frac{1}{4}} < 1.
\end{equation*}
This completes the proof.

\subsection{Proof of Lemma~\ref{lem:matrix-squareroot-lipschitz}}\label{app:matrix-squareroot-lipschitz}

For the first result, we have
\begin{equation*}
\|Z_t^{-1}-(Z^\star)^{-1}\| = \|Z_t^{-1}(Z^\star-Z_t)(Z^\star)^{-1}\| \leq \|Z_t^{-1}\|\|Z_t-Z^\star\|\|(Z^\star)^{-1}\| \leq \frac{1}{\gamma_S^2}\|Z_t-Z^\star\|,
\end{equation*}
where the last inequality follows from $Z_t\succeq\gamma_SI$ in Assumption~\ref{ass:4} and $Z^\star\succeq\gamma_SI$, which is due to $Z_t\stackrel{a.s.}{\rightarrow}Z^\star$ established above.

For the second result, we proceed in two steps.

\noindent $\bullet$ \textbf{Step 1.}
We show that, for any positive definite matrix $X$,
\begin{equation}\label{equ:represent_step1}
X^{-1/2}=\frac{1}{\pi}\int_0^\infty s^{-1/2}(X+sI)^{-1}\,ds.
\end{equation}
Indeed, for any scalar $x>0$, the following identity holds:
\begin{equation*}
x^{-1/2}=\frac{1}{\pi}\int_0^\infty s^{-1/2}(x+s)^{-1}\,ds.
\end{equation*}
To verify this identity, let $s=xu^2$, so that $ds=2xu\,du$. Then,
\begin{align}\label{eq:inverse-squareroot-scalar}
\frac{1}{\pi}\int_0^\infty s^{-1/2}(x+s)^{-1}\,ds &= \frac{1}{\pi}\int_0^\infty(xu^2)^{-1/2}(x+xu^2)^{-1}2xu\,du \nonumber\\
&= \frac{1}{\pi}\int_0^\infty\frac{2}{x^{1/2}(1+u^2)}\,du \nonumber\\
&= \frac{2}{\pi x^{1/2}}\left.\operatorname{arctan}(u)\right|_0^\infty = x^{-1/2}.
\end{align}
We now extend~\eqref{eq:inverse-squareroot-scalar} to matrices. Since $X$ is positive definite, it admits an eigendecomposition $X=UDU^\top$, where $D$ is diagonal and $U$ is orthogonal. Then, $X^{-1/2}=UD^{-1/2}U^\top$, and
\begin{align}\label{eq:inverse-squareroot-matrix}
\frac{1}{\pi}\int_0^\infty s^{-1/2}(X+sI)^{-1}\,ds &= \frac{1}{\pi}\int_0^\infty s^{-1/2}(UDU^\top+sUIU^\top)^{-1}\,ds \nonumber\\
&= \frac{1}{\pi}\int_0^\infty s^{-1/2}\{U(D+sI)U^\top\}^{-1}\,ds \nonumber\\
&= \frac{1}{\pi}\int_0^\infty s^{-1/2}U(D+sI)^{-1}U^\top\,ds \nonumber\\
&= U\left\{\frac{1}{\pi}\int_0^\infty s^{-1/2}(D+sI)^{-1}\,ds\right\}U^\top.
\end{align}
Since $(D+sI)^{-1}$ is diagonal, the integral is evaluated entry-wise. By~\eqref{eq:inverse-squareroot-scalar}, its $i$th diagonal entry equals $\lambda_i^{-1/2}$, where $\lambda_i$ is the $i$th eigenvalue of $X$. Therefore, applying~\eqref{eq:inverse-squareroot-matrix} gives us
\begin{equation*}
\frac{1}{\pi}\int_0^\infty s^{-1/2}(X+sI)^{-1}\,ds = U\left\{\frac{1}{\pi}\int_0^\infty s^{-1/2}(D+sI)^{-1}\,ds\right\}U^\top =UD^{-1/2}U^\top = X^{-1/2}.
\end{equation*}

\noindent $\bullet$ \textbf{Step 2.}
By the representation~\eqref{equ:represent_step1} established in \textbf{Step~1}, we can write
\begin{align*}
\|Z_t^{-1/2}-(Z^\star)^{-1/2}\| &= \left\|\frac{1}{\pi}\int_0^\infty y^{-1/2}\{(Z_t+yI)^{-1}-(Z^\star+yI)^{-1}\}\,dy\right\| \\
&= \left\|\frac{1}{\pi}\int_0^\infty y^{-1/2}(Z_t+yI)^{-1}(Z^\star-Z_t)(Z^\star+yI)^{-1}\,dy\right\| \\
&\leq \frac{\|Z^\star-Z_t\|}{\pi}\int_0^\infty y^{-1/2}\|(Z_t+yI)^{-1}\|\|(Z^\star+yI)^{-1}\|\,dy \\
&\leq \frac{\|Z^\star-Z_t\|}{\pi}\int_0^\infty\frac{y^{-1/2}}{(\gamma_S+y)^2}\,dy,
\end{align*}
where the last inequality follows from $Z_t\succeq\gamma_SI$ and $Z^\star\succeq\gamma_SI$. To evaluate the integral, we let $y=\gamma_Su^2$ and hence obtain the following:
\begin{equation*}
\int_0^\infty\frac{y^{-1/2}}{(\gamma_S+y)^2}\,dy = \int_0^\infty\frac{(\gamma_Su^2)^{-1/2}}{(\gamma_S+\gamma_Su^2)^2}(2\gamma_Su)\,du = \frac{2}{\gamma_S^{3/2}}\int_0^\infty\frac{1}{(1+u^2)^2}\,du.
\end{equation*}
Letting $u=\tan\theta$ further yields
\begin{multline*}
\int_0^\infty\frac{1}{(1+u^2)^2}\,du = \int_0^{\pi/2}\frac{\sec^2\theta}{(1+\tan^2\theta)^2}\,d\theta = \int_0^{\pi/2}\frac{1}{\sec^2\theta}\,d\theta \\
= \int_0^{\pi/2}\cos^2\theta\,d\theta = \int_0^{\pi/2}\frac{1+\cos(2\theta)}{2}\,d\theta = \frac{\pi}{4}.
\end{multline*}
Combining the preceding displays, we conclude that
\begin{equation*}
\|Z_t^{-1/2}-(Z^\star)^{-1/2}\| \leq \frac{\|Z^\star-Z_t\|}{\pi}\int_0^\infty\frac{y^{-1/2}}{(\gamma_S+y)^2}\,dy = \frac{1}{2\gamma_S^{3/2}}\|Z^\star-Z_t\|.
\end{equation*}
This completes the proof.

\subsection{Proof of Lemma~\ref{lem:local-rate}}\label{app:local-rate}

Recall that $\bDelta_t = \bx_t-\bx^\star$, $\mathcal R = I-K^\star$ and $\lambda_m = \lambda_{\min}(\mathcal R)$. By Lemma~\ref{lem:operator-stabilization}, $K^\star$ is symmetric and $\|K^\star\|<1$. Hence, $\mathcal R$ is symmetric positive definite and $\lambda_m>0$.

Also, define the realized and conditional-mean inner-solver operators as
\begin{equation*}
\tQ_t\coloneqq(I-\tK_t)B_t^{-1},
\qquad
Q_t\coloneqq(I-K_t)B_t^{-1}.
\end{equation*}
Conditional on $\mF_{t-0.5}$, $B_t$ is fixed. Therefore, by Lemma~\ref{lem:inner_transition}, we obtain
\begin{multline}\label{eq:local-Q-conditional-mean}
\mE[\tQ_t\mid\mF_{t-0.5}] = \mE[(I-\tK_t)B_t^{-1}\mid\mF_{t-0.5}] \\
= \left\{I-\mE[\tK_t\mid\mF_{t-0.5}]\right\}B_t^{-1} = (I-K_t)B_t^{-1} = Q_t.
\end{multline}

{\red Let $r\in(0,\gamma_H/2)$ be chosen below. On the event $\{\|B_t-B^\star\|\leq r\}$, Assumption~\ref{ass:1} gives $B^\star\succeq\gamma_HI$ and hence $B_t\succeq(\gamma_H-r)I\succeq(\gamma_H/2)I$. Therefore, $\|B_t^{-1}\|\leq2/\gamma_H$, and the following bounds hold on this event.}
By~\eqref{eq:global-uniform-tK}, we have $\|\tK_t\|\leq C_K$. Moreover, \citet[Lemma~3.7]{Wang2026Inference} and the condition on $\tau$ ensure that $\|K_t\| \leq 2\tau\left(1-\sqrt{\mu_t/ \nu_t}\right)^{\tau-2} \leq {\red \gamma_{B,t}/(4\Upsilon_{B,t})\leq1/4}$.
{\red On $\{\|B_t-B^\star\|\leq r\}$, it follows that}
\begin{align}\label{eq:local-Q-uniform-bound}
\|\tQ_t\| &= \|(I-\tK_t)B_t^{-1}\| \leq \big(1+\|\tK_t\|\big)\|B_t^{-1}\| \leq {\red \frac{2(1+C_K)}{\gamma_H}} \lesssim 1, \nonumber\\
\|Q_t\| &= \|(I-K_t)B_t^{-1}\| \leq \big(1+\|K_t\|\big)\|B_t^{-1}\| \leq {\red \frac{5}{2\gamma_H}} \lesssim 1.
\end{align}

{\red On the event $\{\|B_t-B^\star\|\leq r\}$, we further bound the deviation of the conditional-mean operator from its limit as follows:}
\begin{align}\label{eq:local-Q-perturbation-bound}
\|Q_tB^\star-\mathcal R\| &= \left\|(I-K_t)B_t^{-1}B^\star-(I-K^\star)\right\| \nonumber\\
&= \left\|(I-K_t)B_t^{-1}(B^\star-B_t)+(K^\star-K_t)\right\| \nonumber\\
&\leq \|I-K_t\|\|B_t^{-1}\|\|B^\star-B_t\|+\|K_t-K^\star\| \nonumber\\
&\leq \left\{{\red \frac{2(1+\|K_t\|)}{\gamma_H}}+L_K\right\}\|B_t-B^\star\| \nonumber\\
&\leq \left\{{\red \frac{5}{2\gamma_H}}+L_K\right\}\|B_t-B^\star\|.
\end{align}
{\red Here, the first equality follows from the definitions of $Q_t$ and $\mathcal R$; the fourth inequality uses $\|B_t^{-1}\|\leq2/\gamma_H$ on $\{\|B_t-B^\star\|\leq r\}$ and~\eqref{eq:operator-K-bound}; and the last inequality follows from $\|K_t\|\leq1/4$.}

Also, by the $\Upsilon_L$-Lipschitz continuity of $\nabla^2f$, we obtain
\begin{align}\label{eq:local-gradient-linearization}
\|\nabla f_t-B^\star\bDelta_t\|
&= \left\|\int_0^1\left\{\nabla^2f(\bx^\star+s\bDelta_t)-B^\star\right\}\bDelta_t\,ds\right\| \nonumber\\
&\leq \int_0^1\left\|\nabla^2f(\bx^\star+s\bDelta_t)-\nabla^2f(\bx^\star)\right\|\|\bDelta_t\|\,ds \nonumber\\
&\leq \Upsilon_L\int_0^1s\|\bDelta_t\|^2\,ds
= \frac{\Upsilon_L}{2}\|\bDelta_t\|^2.
\end{align}

Combining~\eqref{eq:local-Q-uniform-bound}, \eqref{eq:local-Q-perturbation-bound}, and~\eqref{eq:local-gradient-linearization}, we further obtain {\red the following bound on $\{\|B_t-B^\star\|\leq r\}$:}
\begin{align}\label{eq:local-Q-gradient-bound}
&\|Q_t\nabla f_t-\mathcal R\bDelta_t\|=\left\|Q_t(\nabla f_t-B^\star\bDelta_t)+(Q_tB^\star-\mathcal R)\bDelta_t\right\| \nonumber\\
&\leq \|Q_t\|\|\nabla f_t-B^\star\bDelta_t\|+\|Q_tB^\star-\mathcal R\|\|\bDelta_t\| \nonumber\\
&\leq {\red \frac{5\Upsilon_L}{4\gamma_H}}\|\bDelta_t\|^2+\left\{{\red \frac{5}{2\gamma_H}}+L_K\right\}\|B_t-B^\star\|\|\bDelta_t\| \nonumber\\
&\lesssim \big\{\|\bDelta_t\|+\|B_t-B^\star\|\big\}\|\bDelta_t\|.
\end{align}

Now we localize the analysis by introducing the stopping time: {\red for $k\geq1$ and the radius $r\in(0,\gamma_H/2)$ which will be specified below,}
\begin{equation}\label{eq:local-stopping-time}
{\red \tau_{k,r}\coloneqq\inf\left\{t\geq k:
\begin{aligned}
&\|\bDelta_t\|>r\;\text{ OR }\;\|B_t-B^\star\|>r\;\text{ OR }\;\|\be_{t-1}\|>r\\
&\text{ OR }\;\|\bDelta_{t-1}\|>r\;\text{ OR }\;\|\tilde{B}_t-B^\star\|_{\mathrm F}>r\\
&\text{ OR }\;\gamma_{B,t}>\lambda_{\min}(B^\star)\;\text{ OR }\;\Upsilon_{B,t}<\lambda_{\max}(B^\star)
\end{aligned}\right\}.}
\end{equation}
{\red Since $\bDelta_{t-1}$, $\bDelta_t$, $\tilde{B}_t$, $B_t$, $\be_{t-1}$, $\gamma_{B,t}$, and $\Upsilon_{B,t}$ are $\mF_{t-1}$-measurable, $\tau_{k,r}$ is a stopping time and $\{\tau_{k,r}>t\}\in\mF_{t-1}$. On $\{\tau_{k,r}>t\}$, we have $\|B_t-B^\star\|\leq r$, so the local bounds above hold. The clipping interval also contains $B^\star$, and hence $\|B_t-B^\star\|_{\mathrm F}\leq\|\tilde{B}_t-B^\star\|_{\mathrm F}$.} Moreover, Theorem~\ref{thm:global-convergence} and Lemma~\ref{lem:operator-stabilization} together ensure
\begin{equation*}
\bDelta_t\stackrel{a.s.}{\rightarrow}\zero,
\qquad
B_t\stackrel{a.s.}{\rightarrow}B^\star,
\qquad
\be_t\stackrel{a.s.}{\rightarrow}\zero.
\end{equation*}
{\red The proof of Lemma~\ref{lem:operator-stabilization} also gives $\|\tilde{B}_t-B^\star\|_{\mathrm F}\stackrel{a.s.}{\rightarrow}0$, and the clipping interval contains $B^\star$ for all sufficiently large $t$ almost surely. Thus, for each fixed $r\in(0,\gamma_H/2)$, $\tau_{k,r}=\infty$ for all sufficiently large $k$ almost surely, which implies}
\begin{equation}\label{eq:local-stopping-probability}
\mathbb{P}(\tau_{k,r}<\infty)\rightarrow 0 \qquad\text{as}\qquad k\rightarrow\infty.
\end{equation}
{\red Once $r$ is chosen below, for any $\epsilon>0$ we can choose $k=k(\epsilon,r)$ sufficiently large such that $\PP(\tau_{k,r}<\infty)<\epsilon/2$. For every $t\geq k$, we then have}
\begin{equation}\label{equ:tau}
\mathbb{P}(\tau_{k,r} \leq t) \leq \mathbb{P}(\tau_{k, r} < \infty) < \frac{\epsilon}{2}.
\end{equation}

Next we determine the radius $r$ and index $k$ step by step. We first choose a margin $\varepsilon_0 > 0$. If $\varphi<1$, we could choose any $\varepsilon_0\in(0,1)$. If $\varphi=1$, we choose $\varepsilon_0\in(0,1)$ sufficiently small such that
\begin{equation}\label{eq:local-harmonic-margin}
2C_\varphi(1-\varepsilon_0)\lambda_m>\eta,
\end{equation}
which is possible since $C_{\varphi} > \eta / (2\lambda_m)$. The precise choice of the localization radius $r$ will be specified after deriving the preliminary recursion for the iterate error.

By the outer update, Lemma~\ref{lem:inner_transition}, and the definition of $\tQ_t$, we have
\begin{equation*}
\bDelta_{t+1}=\bDelta_t-\varphi_t\tQ_t(\nabla f_t+\be_t).
\end{equation*}
Conditional on $\mF_{t-0.5}$, the quantities $(\bDelta_t,\nabla f_t,\be_t,Q_t)$ are fixed. Therefore, we obtain
\begin{multline}\label{eq:local-direct-conditional-iterate}
\mE[\|\bDelta_{t+1}\|^2\mid\mF_{t-0.5}] = \|\bDelta_t-\varphi_tQ_t(\nabla f_t+\be_t)\|^2 \\
+\varphi_t^2\mE[\|(\tQ_t-Q_t)(\nabla f_t+\be_t)\|^2\mid\mF_{t-0.5}].
\end{multline}
Here, the cross term vanishes since $\mE[\tQ_t-Q_t\mid\mF_{t-0.5}]=\zero$ given in~\eqref{eq:local-Q-conditional-mean}. Moreover,
\begin{equation*}
\bDelta_t-\varphi_tQ_t(\nabla f_t+\be_t)
=(I-\varphi_t\mathcal R)\bDelta_t-\varphi_t(Q_t\nabla f_t-\mathcal R\bDelta_t)-\varphi_tQ_t\be_t.
\end{equation*}

{\red On the event $\{\tau_{k,r}>t\}$, since} $\mathcal R$ is symmetric positive definite, for all sufficiently large $t$ such that $\|I-\varphi_t\mathcal R\|\leq1$, we have
\begin{align}\label{equ:delta_recursion_term_1}
&\|\bDelta_t-\varphi_tQ_t(\nabla f_t+\be_t)\|^2 = \left\|(I-\varphi_t\mathcal R)\bDelta_t-\varphi_t(Q_t\nabla f_t-\mathcal R\bDelta_t)-\varphi_tQ_t\be_t\right\|^2 \nonumber\\
&\leq \|(I-\varphi_t\mathcal R)\bDelta_t\|^2+2\varphi_t\|\bDelta_t\|\|Q_t\nabla f_t-\mathcal R\bDelta_t\| \nonumber\\
&\quad + {\red \frac{5\varphi_t}{\gamma_H}}\|\bDelta_t\|\|\be_t\|+2\varphi_t^2\|Q_t\nabla f_t-\mathcal R\bDelta_t\|^2+{\red \frac{25\varphi_t^2}{2\gamma_H^2}}\|\be_t\|^2 \nonumber\\
&\leq \left(1-2\lambda_m\varphi_t+\|\mathcal R\|^2\varphi_t^2\right)\|\bDelta_t\|^2+2\varphi_t\|\bDelta_t\|\|Q_t\nabla f_t-\mathcal R\bDelta_t\| \nonumber\\
&\quad + {\red \frac{5\varphi_t}{\gamma_H}}\|\bDelta_t\|\|\be_t\|+2\varphi_t^2\|Q_t\nabla f_t-\mathcal R\bDelta_t\|^2+{\red \frac{25\varphi_t^2}{2\gamma_H^2}}\|\be_t\|^2,
\end{align}
{\red where the second inequality uses the Cauchy--Schwarz inequality, $\|\ba+\bb\|^2\leq2\|\ba\|^2+2\|\bb\|^2$, and the local bound $\|Q_t\|\leq5/(2\gamma_H)$ from~\eqref{eq:local-Q-uniform-bound}, and the last inequality holds due to the fact that}
\begin{equation*}
\|(I-\varphi_t\mathcal R)\ba\|^2 \leq \left(1-2\lambda_m\varphi_t+\|\mathcal R\|^2\varphi_t^2\right)\|\ba\|^2.
\end{equation*}

On $\{\tau_{k,r}>t\}$, by the definition of $\tau_{k,r}$ and~\eqref{eq:local-Q-gradient-bound}, we have
\begin{equation*}
\|Q_t\nabla f_t-\mathcal R\bDelta_t\| \leq \left\{{\red \frac{5\Upsilon_L}{4\gamma_H}+\frac{5}{2\gamma_H}+L_K}\right\}r\|\bDelta_t\|.
\end{equation*}
{\red Moreover, on $\{\tau_{k,r}>t\}$, Young's inequality directly implies}
\begin{equation*}
{\red \frac{5\varphi_t}{\gamma_H}}\|\bDelta_t\|\|\be_t\| \leq \varepsilon_0\lambda_m\varphi_t\|\bDelta_t\|^2+{\red \frac{25\varphi_t}{4\varepsilon_0\lambda_m\gamma_H^2}}\|\be_t\|^2.
\end{equation*}
On $\{\tau_{k,r}>t\}$, substituting these two bounds into~\eqref{equ:delta_recursion_term_1} yields
\begin{multline}\label{eq:local-conditional-mean-bound}
    \|\bDelta_t-\varphi_tQ_t(\nabla f_t+\be_t)\|^2 \leq \Bigg[1-(2-\varepsilon_0)\lambda_m\varphi_t+2\left\{{\red \frac{5\Upsilon_L}{4\gamma_H}+\frac{5}{2\gamma_H}+L_K}\right\}r\varphi_t \\
    + \left(\|\mathcal R\|^2+2\left\{{\red \frac{5\Upsilon_L}{4\gamma_H}+\frac{5}{2\gamma_H}+L_K}\right\}^2r^2\right)\varphi_t^2\Bigg]\|\bDelta_t\|^2 + {\red \frac{25}{4\gamma_H^2}} \left(\frac{\varphi_t}{\varepsilon_0\lambda_m}+2\varphi_t^2\right)\|\be_t\|^2.
\end{multline}

We next control the second term in~\eqref{eq:local-direct-conditional-iterate}. {\red On the event $\{\tau_{k,r}>t\}$, since $(\nabla f_t,\be_t)$ is $\mF_{t-0.5}$-measurable,~\eqref{eq:local-Q-uniform-bound} gives}
\begin{align}\label{equ:centered-sketching_bound}
&\mE[\|(\tQ_t-Q_t)(\nabla f_t+\be_t)\|^2\mid\mF_{t-0.5}] \leq \mE[\|\tQ_t-Q_t\|^2\mid\mF_{t-0.5}]\|\nabla f_t+\be_t\|^2 \nonumber\\
&\leq 2\left\{{\red \frac{2(1+C_K)}{\gamma_H}+\frac{5}{2\gamma_H}}\right\}^2\left(\|\nabla f_t\|^2+\|\be_t\|^2\right) \nonumber\\
&\leq 2\left\{{\red \frac{2(1+C_K)}{\gamma_H}+\frac{5}{2\gamma_H}}\right\}^2\left(\Upsilon_H^2\|\bDelta_t\|^2+\|\be_t\|^2\right).
\end{align}

Combining~\eqref{eq:local-direct-conditional-iterate},~\eqref{eq:local-conditional-mean-bound} and~\eqref{equ:centered-sketching_bound}, we obtain
\begin{align}\label{eq:local-direct-iterate-preliminary}
&\mE[\|\bDelta_{t+1}\|^2\mid\mF_{t-0.5}]\1_{\{\tau_{k,r}>t\}} \nonumber\\
&\leq \Bigg[1-(2-\varepsilon_0)\lambda_m\varphi_t+2\left\{{\red \frac{5\Upsilon_L}{4\gamma_H}+\frac{5}{2\gamma_H}+L_K}\right\}r\varphi_t \nonumber\\
&\quad+\Bigg(\|\mathcal R\|^2+2\left\{{\red \frac{5\Upsilon_L}{4\gamma_H}+\frac{5}{2\gamma_H}+L_K}\right\}^2r^2+2\Upsilon_H^2\left\{{\red \frac{9+4C_K}{2\gamma_H}}\right\}^2\Bigg)\varphi_t^2\Bigg]\|\bDelta_t\|^2\1_{\{\tau_{k,r}>t\}} \nonumber\\
&\quad+\Bigg[{\red \frac{25}{4\varepsilon_0\lambda_m\gamma_H^2}}\varphi_t+2\left\{{\red \frac{25}{4\gamma_H^2}}+\left\{{\red \frac{9+4C_K}{2\gamma_H}}\right\}^2\right\}\varphi_t^2\Bigg]\|\be_t\|^2\1_{\{\tau_{k,r}>t\}}.
\end{align}
For notational conciseness, define the following deterministic constants:
\begin{alignat*}{2}
L_\Delta &\coloneqq {\red \frac{5\Upsilon_L}{4\gamma_H}+\frac{5}{2\gamma_H}+L_K},\qquad
&L_S &\coloneqq {\red \frac{2(1+C_K)}{\gamma_H}+\frac{5}{2\gamma_H}},\\
D_\Delta(r) &\coloneqq \|\mathcal R\|^2+2L_\Delta^2r^2+2\Upsilon_H^2L_S^2,\qquad
&D_e &\coloneqq {\red \frac{25}{4\varepsilon_0\lambda_m\gamma_H^2}+2\left\{\frac{25}{4\gamma_H^2}+L_S^2\right\}}.
\end{alignat*}
Therefore, for all sufficiently large $t$ such that $\varphi_t\leq1$,~\eqref{eq:local-direct-iterate-preliminary} implies
\begin{multline}\label{eq:local-direct-iterate-condensed}
\mE[\|\bDelta_{t+1}\|^2\mid\mF_{t-0.5}]\1_{\{\tau_{k,r}>t\}} \\
\leq \left\{\left[1-(2-\varepsilon_0)\lambda_m\varphi_t+2L_\Delta r\varphi_t+D_\Delta(r)\varphi_t^2\right]\|\bDelta_t\|^2+D_e\varphi_t\|\be_t\|^2\right\}\1_{\{\tau_{k,r}>t\}}.
\end{multline}

We now choose {\red $r\in(0,\min\{1,\gamma_H/2\})$} sufficiently small such that
\begin{equation}\label{eq:local-r-condition}
2L_\Delta r\leq\frac{\varepsilon_0\lambda_m}{2}.
\end{equation}
For this fixed $r$ and any given $\epsilon>0$, we first choose $k=k(\epsilon,r)$ sufficiently large so that $\PP(\tau_{k,r}<\infty)<\epsilon/2$, which guarantees~\eqref{equ:tau} for every $t\geq k$. Additionally, for every $t\geq k$, we require
\begin{align}\label{eq:local-k-conditions}
\varphi_t \leq 1,
\qquad
\varphi_t\|\mathcal R\| \leq 1,
\qquad
D_\Delta(r)\varphi_t \leq \frac{\varepsilon_0\lambda_m}{2},
\qquad
0<2(1-\varepsilon_0)\lambda_m\varphi_t < 1.
\end{align}
Specifically, the third condition allows the second-order term $D_\Delta(r)\varphi_t^2$ in~\eqref{eq:local-direct-iterate-condensed} to be absorbed into the first-order contraction margin, thereby preserving the desired contraction coefficient.

For these fixed $(k,r)$, we define the following localized second moments
\begin{align}\label{eq:local-uvw-definitions}
u_t \coloneqq \mE[\|\bDelta_t\|^2\1_{\{\tau_{k,r}>t\}}],
\quad
v_t\coloneqq\mE[\|\be_t\|^2\1_{\{\tau_{k,r}>t\}}],
\quad
w_t \coloneqq \mE[\|B_t-B^\star\|^2\1_{\{\tau_{k,r}>t\}}].
\end{align}

\noindent $\bullet$ \textbf{Convergence rate of $B_t$.}
For any $M>0$ and $t\geq k$, by Markov's inequality and~\eqref{equ:tau}, we have
\begin{align}\label{eq:local-B-probability}
&\PP(\|B_t-B^\star\|>M\sqrt{\eta_t}) = \PP(\|B_t-B^\star\|^2>M^2\eta_t) \nonumber\\
&= \mathbb{P}\!\left(\{\|B_t-B^{\star}\|^2 > M^2 \eta_t\} \cap \{\tau_{k,r} > t\}\right) + \mathbb{P}\!\left(\{\|B_t-B^{\star}\|^2 > M^2 \eta_t\} \cap \{\tau_{k,r} \leq t\}\right) \nonumber\\
&\leq \PP(\tau_{k,r}\leq t)+\PP(\|B_t-B^\star\|^2\1_{\{\tau_{k,r}>t\}}>M^2\eta_t) \leq \frac{\epsilon}{2}+\frac{w_t}{M^2\eta_t}.
\end{align}
Therefore, in order to obtain $O_p(\sqrt{\eta_t})$, it suffices to establish $w_t=O(\eta_t)$. Recall that $k=k(\epsilon,r)$ is fixed. {\red On $\{\tau_{k,r}>t\}$, we have $\|\tilde{B}_k-B^\star\|_{\mathrm F}\leq r$. By the projection property of $\operatorname{Clip}$ and the decomposition of $\tilde{B}_t$ at index $k$, we obtain
\begin{align}\label{eq:local-w-comparison}
w_t &\leq \mE[\|\tilde{B}_t-B^\star\|_{\mathrm F}^2\1_{\{\tau_{k,r}>t\}}] \nonumber\\
&= \mE\left[\left\|\frac{k}{t}(\tilde{B}_k-B^\star)+\frac{1}{t}\sum_{i=k}^{t-1}(H_i-\nabla^2f_i)+\frac{1}{t}\sum_{i=k}^{t-1}(\nabla^2f_i-B^\star)\right\|_{\mathrm F}^2\1_{\{\tau_{k,r}>t\}}\right] \nonumber\\
&\lesssim \frac{k^2r^2}{t^2}+\frac{1}{t^2}\mE\left[\left\|\sum_{i=k}^{t-1}(H_i-\nabla^2f_i)\1_{\{\tau_{k,r}>i\}}\right\|_{\mathrm F}^2\right]+\frac{1}{t^2}\mE\left[\left(\sum_{i=k}^{t-1}\|\bDelta_i\|\1_{\{\tau_{k,r}>i\}}\right)^2\right] \nonumber\\
&\lesssim \frac{k^2r^2}{t^2}+\frac{1}{t^2}\sum_{i=k}^{t-1}\mE[\|H_i-\nabla^2f_i\|_{\mathrm F}^2\1_{\{\tau_{k,r}>i\}}]+\frac{1}{t^2}\left(\sum_{i=k}^{t-1}\sqrt{u_i}\right)^2 \nonumber\\
&\lesssim \frac{k^2r^2}{t^2}+\frac{1}{t}+\frac{1}{t^2}\left(\sum_{i=k}^{t-1}\sqrt{u_i}\right)^2.
\end{align}
}
Here, $(H_i-\nabla^2f_i)\1_{\{\tau_{k,r}>i\}}$ is a matrix-valued martingale difference sequence because $\1_{\{\tau_{k,r}>i\}}$ is $\mF_{i-1}$-measurable. {\red Therefore, the third inequality follows from the $\Upsilon_L$-Lipschitz continuity of $\nabla^2f$, $\|\tilde{B}_k-B^\star\|_{\mathrm F}\leq r$ on $\{\tau_{k,r}>t\}$, and $\{\tau_{k,r}>t\}\subseteq\{\tau_{k,r}>i\}$ for $i\leq t$; the fourth inequality follows from martingale orthogonality under the Frobenius inner product and Minkowski's inequality; and the last inequality follows from Assumption~\ref{ass:3} and $\|\bDelta_i\|\leq r$ on $\{\tau_{k,r}>i\}$.} Then, it remains to establish the rates of $u_t$ and $v_t$.

To complete the proof, we present the following two auxiliary lemmas. And their proofs will be deferred to Appendices~\ref{app:local-coupled-bounds} and~\ref{app:local-finite-step-rate}, respectively.

\begin{lemma}\label{lem:local-coupled-bounds}
For $(k,r)$ chosen above and all sufficiently large $t$, we have
\begin{equation}\label{eq:local-coupled-iterate-bound}
u_{t+1}\leq\left\{1-2(1-\varepsilon_0)\lambda_m\varphi_t\right\}u_t+D_e\varphi_tv_t.
\end{equation}
and
\begin{multline}\label{eq:local-coupled-tracking-bound}
v_t\lesssim\left\{\prod_{j=k+1}^{t}(1-\eta_j)\right\}^2{\red v_k}+\sum_{i=k+1}^{t}\left\{\prod_{j=i+1}^{t}(1-\eta_j)^2\right\}\eta_i^2 \\
+\left\{\sum_{i=k+1}^{t}\left\{\prod_{j=i+1}^{t}(1-\eta_j)\right\}\varphi_{i-1}\big(\sqrt{u_{i-1}}+\sqrt{v_{i-1}}\big)\right\}^2.
\end{multline}
Note that $\prod_{j=m}^{n}(1-\eta_j)\coloneqq1$ when $m > n$.
\end{lemma}

\begin{lemma}\label{lem:local-finite-step-rate}
Under the conditions of Lemma~\ref{lem:local-rate}, for every integer $q\geq0$, there exists a deterministic constant $C_q(k,r)>0$ such that
\begin{equation}\label{eq:local-finite-step-rate}
\max\{u_t,v_t\}\leq C_q(k,r)\left\{\eta_t+\left(\frac{\varphi_t}{\eta_t}\right)^{2q}\right\}
\end{equation}
for all sufficiently large $t$.
\end{lemma}

Since $\varphi>\eta$, we can choose the finite integer $q_\star\coloneqq\left\lfloor \eta/ [2(\varphi-\eta)]\right\rfloor+1,$ which satisfies $2q_\star(\varphi-\eta)>\eta$. Consequently, $\left(\varphi_t/\eta_t\right)^{2q_\star}=o(\eta_t)$. Applying Lemma~\ref{lem:local-finite-step-rate} with $q=q_\star$ gives
\begin{equation}\label{eq:local-stopped-primary-rates}
u_t=O(\eta_t), \qquad v_t=O(\eta_t).
\end{equation}
Returning to~\eqref{eq:local-w-comparison} and using~\eqref{eq:local-stopped-primary-rates}, we obtain
\begin{align}\label{eq:local-stopped-Hessian-rate}
w_t \lesssim {\red \frac{k^2r^2}{t^2}+}\frac{1}{t}+\frac{1}{t^2}\left(\sum_{i=k}^{t-1}\sqrt{\eta_i}\right)^2 \lesssim \frac{1}{t}+\eta_t \lesssim \eta_t,
\end{align}
Therefore, for all sufficiently large $t$, there exists $C_{B,k,r} > 0$ such that
\begin{equation*}
\PP(\|B_t-B^\star\|>M\sqrt{\eta_t}) \leq \frac{\epsilon}{2}+{\red \frac{C_{B,k,r}}{M^2}}.
\end{equation*}
Choosing $M \geq \sqrt{2C_{B,k,r} / \epsilon}$ ensures $\PP(\|B_t-B^\star\|>M\sqrt{\eta_t}) < \epsilon$ for $t$ large enough. Therefore, 
\begin{equation}\label{eq:local-B-rate}
\|B_t-B^\star\|=O_p(\sqrt{\eta_t}).
\end{equation}

\noindent $\bullet$ \textbf{Convergence rates of $\bbm_t$ and $\bDelta_t$.}
By~\eqref{eq:local-stopped-primary-rates} and the same stopping-time decomposition as above, for the fixed $(k,r)$ and $t$ large enough, there exist constants $C_{\Delta,k,r},C_{e,k,r}>0$ such that
\begin{align*}
\PP(\|\bDelta_t\|>M\sqrt{\eta_t}) \leq \frac{\epsilon}{2}+\frac{C_{\Delta,k,r}}{M^2},
\qquad
\PP(\|\be_t\|>M\sqrt{\eta_t}) \leq \frac{\epsilon}{2}+\frac{C_{e,k,r}}{M^2}.
\end{align*}
Again, choosing $M \geq \sqrt{2\max\{C_{\Delta,k,r},C_{e,k,r}\} / \epsilon}$ ensures that
\begin{equation}\label{eq:local-primary-rates}
\|\bDelta_t\|=\|\bx_t-\bx^\star\|=O_p(\sqrt{\eta_t}),
\qquad
\|\be_t\|=O_p(\sqrt{\eta_t}).
\end{equation}
Since $\bbm_t=\be_t+\nabla f_t$ and $\|\nabla f_t\|\leq\Upsilon_H\|\bDelta_t\|$,~\eqref{eq:local-primary-rates} further yields
\begin{equation*}
\|\bbm_t\|\leq\|\be_t\|+\Upsilon_H\|\bDelta_t\|=O_p(\sqrt{\eta_t}).
\end{equation*}

\noindent $\bullet$ \textbf{Convergence rates of $(\alpha_t,\beta_t,\gamma_t)$.}
Combining~\eqref{eq:operator-parameter-bound} and~\eqref{eq:local-B-rate}, we obtain
\begin{equation*}
\max\left\{|\alpha_t-\alpha^\star|,\;|\beta_t-\beta^\star|,\;|\gamma_t-\gamma^\star|\right\}=O_p(\sqrt{\eta_t}).
\end{equation*}

\noindent $\bullet$ \textbf{Convergence rate of $K_t$.}
Combining~\eqref{eq:operator-K-bound} and~\eqref{eq:local-B-rate} gives
\begin{equation*}
\|K_t-K^\star\|\leq L_K\|B_t-B^\star\|=O_p(\sqrt{\eta_t}).
\end{equation*}

Together with~\eqref{eq:local-B-rate} and~\eqref{eq:local-primary-rates}, these bounds yield
\begin{multline*}
\max\big\{\|\bbm_t-\nabla f_t\|,\;\|\bbm_t\|,\;\|\bx_t-\bx^\star\|,\;\|B_t-B^\star\|, \\
|\alpha_t-\alpha^\star|,\;|\beta_t-\beta^\star|,\;|\gamma_t-\gamma^\star|,\;\|K_t-K^\star\|\big\}=O_p(\sqrt{\eta_t}),
\end{multline*}
which completes the proof of Lemma~\ref{lem:local-rate}.

\subsection{Proof of Lemma~\ref{lem:local-coupled-bounds}}\label{app:local-coupled-bounds}

Since $\1_{\{\tau_{k,r}>t+1\}}\leq\1_{\{\tau_{k,r}>t\}}$ and $\1_{\{\tau_{k,r}>t\}}$ is $\mF_{t-0.5}$-measurable,~\eqref{eq:local-direct-iterate-condensed}--\eqref{eq:local-k-conditions} yield
\begin{multline}\label{eq:local-direct-iterate-recursion}
u_{t+1} = \mE[\|\bDelta_{t+1}\|^2\1_{\{\tau_{k,r}>t+1\}}] \leq \mE\left[\mE[\|\bDelta_{t+1}\|^2\mid\mF_{t-0.5}]\1_{\{\tau_{k,r}>t\}}\right] \\
\leq \left\{1-2(1-\varepsilon_0)\lambda_m\varphi_t\right\}u_t+D_e\varphi_tv_t,
\qquad t\geq k.
\end{multline}
This proves the first result. It remains to establish~\eqref{eq:local-coupled-tracking-bound}. By the moving-average update,
\begin{equation*}
\be_t=(1-\eta_t)\be_{t-1}+\eta_t(\bg_t-\nabla f_t)+(1-\eta_t)(\nabla f_{t-1}-\nabla f_t).
\end{equation*}
{\red On $\{\tau_{k,r}>k\}$, we have $\|\be_{k-1}\|\leq r$ and $\|\bDelta_{k-1}\|,\|\bDelta_k\|\leq r$. Thus, $\|\nabla f_{k-1}-\nabla f_k\|\leq2\Upsilon_Hr$, and the preceding recursion and Assumption~\ref{ass:2} give $v_k\lesssim r^2+\eta_k^2<\infty$.}
Iterating this identity from the fixed index $k$ gives
\begin{multline}\label{eq:local-tracking-unrolling}
\be_t = \left\{\prod_{j=k+1}^{t}(1-\eta_j)\right\}\be_k+\sum_{i=k+1}^{t}\left\{\prod_{j=i+1}^{t}(1-\eta_j)\right\}\eta_i(\bg_i-\nabla f_i) \\
+\sum_{i=k+1}^{t}\left\{\prod_{j=i+1}^{t}(1-\eta_j)\right\}(1-\eta_i)(\nabla f_{i-1}-\nabla f_i).
\end{multline}
Since $\1_{\{\tau_{k,r}>i\}}$ is $\mF_{i-1}$-measurable and $\mE[\bg_i-\nabla f_i\mid\mF_{i-1}]=\zero$, $\left\{(\bg_i-\nabla f_i)\1_{\{\tau_{k,r}>i\}}\right\}_{i\geq k+1}$ forms a martingale difference sequence. Moreover, $\1_{\{\tau_{k,r}>t\}}\leq\1_{\{\tau_{k,r}>i\}}$ for every $i\leq t$. As a result, by Jensen's inequality and Assumption~\ref{ass:2}, we obtain
\begin{align}\label{eq:local-stopped-gradient-noise}
&\hskip-0.22cm \mE\left[\left\|\sum_{i=k+1}^{t}\left\{\prod_{j=i+1}^{t}(1-\eta_j)\right\}\eta_i(\bg_i-\nabla f_i)\right\|^2\1_{\{\tau_{k,r}>t\}}\right] \nonumber\\
&\hskip-0.22cm\leq \mE\left[\left\|\sum_{i=k+1}^{t}\left\{\prod_{j=i+1}^{t}(1-\eta_j)\right\}\eta_i(\bg_i-\nabla f_i)\1_{\{\tau_{k,r}>i\}}\right\|^2\right] \nonumber\\
&\hskip-0.22cm = \sum_{i=k+1}^{t}\left\{\prod_{j=i+1}^{t}(1-\eta_j)^2\right\}\eta_i^2\mE\left[\|\bg_i-\nabla f_i\|^2\1_{\{\tau_{k,r}>i\}}\right] \lesssim \sum_{i=k+1}^{t}\left\{\prod_{j=i+1}^{t}(1-\eta_j)^2\right\}\eta_i^2.
\end{align}

For the last term in~\eqref{eq:local-tracking-unrolling}, by the $\Upsilon_H$-Lipschitz continuity of $\nabla f$ and Lemma~\ref{lem:inner_transition}, we have
\begin{equation*}
\|\nabla f_i-\nabla f_{i-1}\|\leq \Upsilon_H\varphi_{i-1}\|\bz_{i-1,\tau}\|=\Upsilon_H\varphi_{i-1}\|\tQ_{i-1}(\nabla f_{i-1}+\be_{i-1})\|.
\end{equation*}
Then, by~\eqref{eq:local-Q-uniform-bound} and $\1_{\{\tau_{k,r}>i\}}\leq\1_{\{\tau_{k,r}>i-1\}}$, we obtain
\begin{align}\label{eq:local-stopped-gradient-drift}
&\mE[\|\nabla f_i-\nabla f_{i-1}\|^2\1_{\{\tau_{k,r}>i\}}] \leq \Upsilon_H^2\varphi_{i-1}^2\mE[\|\bz_{i-1,\tau}\|^2\1_{\{\tau_{k,r}>i-1\}}] \nonumber\\
&= \Upsilon_H^2\varphi_{i-1}^2\mE[\|\tQ_{i-1}(\nabla f_{i-1}+\be_{i-1})\|^2\1_{\{\tau_{k,r}>i-1\}}] \nonumber\\
&\leq {\red \frac{4 \Upsilon_H^2(1+C_K)^2}{\gamma_H^2}} \varphi_{i-1}^2\mE[\|\nabla f_{i-1}+\be_{i-1}\|^2\1_{\{\tau_{k,r}>i-1\}}] \nonumber\\
&\leq {\red \frac{8\Upsilon_H^2(1+C_K)^2}{\gamma_H^2}}\varphi_{i-1}^2\left\{\mE[\|\nabla f_{i-1}\|^2\1_{\{\tau_{k,r}>i-1\}}]+v_{i-1}\right\} \nonumber\\
&\leq {\red \frac{8\Upsilon_H^2(1+C_K)^2}{\gamma_H^2}} \varphi_{i-1}^2\left(\Upsilon_H^2u_{i-1}+v_{i-1}\right) \lesssim \varphi_{i-1}^2(u_{i-1}+v_{i-1}),
\end{align}
where the third inequality follows from {\red $\|\tQ_{i-1}\|\leq2(1+C_K)/\gamma_H$ on $\{\tau_{k,r}>i-1\}$} in~\eqref{eq:local-Q-uniform-bound}, and the last inequality uses $\|\nabla f_{i-1}\|\leq\Upsilon_H\|\bDelta_{i-1}\|$ together with the definitions of $u_{i-1}$ and $v_{i-1}$.

Applying Minkowski's inequality to~\eqref{eq:local-tracking-unrolling} {\red and using~\eqref{eq:local-stopped-gradient-noise} and~\eqref{eq:local-stopped-gradient-drift}}, we obtain
\begin{multline*}
\sqrt{v_t} \lesssim \left\{\prod_{j=k+1}^{t}(1-\eta_j)\right\}\sqrt{v_k}
+\left\{\sum_{i=k+1}^{t}\left\{\prod_{j=i+1}^{t}(1-\eta_j)^2\right\}\eta_i^2\right\}^{1/2} \\
\quad + \sum_{i=k+1}^{t}\left\{\prod_{j=i+1}^{t}(1-\eta_j)\right\}\varphi_{i-1}\big(\sqrt{u_{i-1}}+\sqrt{v_{i-1}}\big).
\end{multline*}
Squaring both sides and using $(a+b+c)^2\leq3(a^2+b^2+c^2)$ yields
\begin{multline*}
v_t\lesssim \left\{\prod_{j=k+1}^{t}(1-\eta_j)^2\right\} v_k
+\sum_{i=k+1}^{t}\left\{\prod_{j=i+1}^{t}(1-\eta_j)^2\right\}\eta_i^2 \\
+\left\{\sum_{i=k+1}^{t}\left\{\prod_{j=i+1}^{t}(1-\eta_j)\right\}\varphi_{i-1}\big(\sqrt{u_{i-1}}+\sqrt{v_{i-1}}\big)\right\}^2,
\end{multline*}
which proves~\eqref{eq:local-coupled-tracking-bound} and completes the proof.

\subsection{Proof of Lemma~\ref{lem:local-finite-step-rate}}\label{app:local-finite-step-rate}

We proceed by induction on $q$. For $q=0$, the right-hand side of~\eqref{eq:local-finite-step-rate} is of constant order. By the definition of the stopping time, $u_t=\mE[\|\bDelta_t\|^2\1_{\{\tau_{k,r}>t\}}]\leq r^2$.
Moreover, on $\{\tau_{k,r}>t\}$,
\begin{equation*}
\|\be_{t-1}\|\leq r,
\qquad
\|\bDelta_{t-1}\|\leq r,
\qquad
\|\bDelta_t\|\leq r.
\end{equation*}
Using the recursion for $\be_t$, Assumption~\ref{ass:2} {\red and the bound that $\|\nabla f_{t-1}-\nabla f_t\|\leq2\Upsilon_Hr$ on the event $\{\tau_{k,r}>t\}$}, we obtain
\begin{align*}
v_t &= \mE[\|\be_t\|^2\1_{\{\tau_{k,r}>t\}}] \\
&\lesssim \mE\left[\left\{\|\be_{t-1}\|^2+\eta_t^2\|\bg_t-\nabla f_t\|^2+\|\nabla f_{t-1}-\nabla f_t\|^2\right\}\1_{\{\tau_{k,r}>t\}}\right] \\
&\lesssim {\red r^2+\eta_t^2} \lesssim 1.
\end{align*}
Therefore,~\eqref{eq:local-finite-step-rate} holds for $q=0$.

Suppose that~\eqref{eq:local-finite-step-rate} holds for some integer $q\geq0$. Then, we have
\begin{equation}\label{eq:local-q-square-root}
\sqrt{u_i}+\sqrt{v_i}\lesssim\sqrt{\eta_i}+\left(\frac{\varphi_i}{\eta_i}\right)^q.
\end{equation}
Since $\eta<\varphi\leq1$, by Lemma~\ref{lem:polynomial_product_rate}, we obtain
\begin{align}\label{eq:local-fast-convolution-bounds}
\left\{\prod_{j=k+1}^{t}(1-\eta_j)\right\}^2&=o(\eta_t),
&
\sum_{i=k+1}^{t}\left\{\prod_{j=i+1}^{t}(1-\eta_j)\right\}\varphi_{i-1}\sqrt{\eta_{i-1}}&\lesssim\frac{\varphi_t}{\eta_t}\sqrt{\eta_t},
\nonumber\\
\sum_{i=k+1}^{t}\left\{\prod_{j=i+1}^{t}(1-\eta_j)^2\right\}\eta_i^2&\lesssim\eta_t,
&
\sum_{i=k+1}^{t}\left\{\prod_{j=i+1}^{t}(1-\eta_j)\right\}\varphi_{i-1}\left(\frac{\varphi_{i-1}}{\eta_{i-1}}\right)^q&\lesssim\left(\frac{\varphi_t}{\eta_t}\right)^{q+1}.
\end{align}
Substituting~\eqref{eq:local-q-square-root} and~\eqref{eq:local-fast-convolution-bounds} into~\eqref{eq:local-coupled-tracking-bound}, we obtain
\begin{equation}\label{eq:local-v-q-improvement}
v_t \lesssim \eta_t+\eta_t\left(\frac{\varphi_t}{\eta_t}\right)^2+\left(\frac{\varphi_t}{\eta_t}\right)^{2(q+1)} \lesssim \eta_t+\left(\frac{\varphi_t}{\eta_t}\right)^{2(q+1)},
\end{equation}
where the last inequality uses the fact that $\varphi_t/\eta_t\rightarrow0$. Consequently, there exists a deterministic constant $C_{v,q+1}(k,r)>0$ such that
\begin{equation}\label{eq:local-v-q-bound}
v_t\leq C_{v,q+1}(k,r)\left\{\eta_t+\left(\frac{\varphi_t}{\eta_t}\right)^{2(q+1)}\right\}
\end{equation}
for all sufficiently large $t$.

Combining~\eqref{eq:local-v-q-bound} with~\eqref{eq:local-coupled-iterate-bound}, we obtain a deterministic constant $C_{u,q+1}(k,r)>0$ such that
\begin{equation}\label{eq:local-u-q-recursion}
u_{t+1}\leq\left\{1-\frac{2C_\varphi(1-\varepsilon_0)\lambda_m}{(t+1)^\varphi}\right\}u_t+\frac{C_{u,q+1}(k,r)}{(t+1)^{\varphi+\min\{\eta,\,2(q+1)(\varphi-\eta)\}}}.
\end{equation}
By Lemma~\ref{lem:polynomial_product_rate}, we obtain for $\varphi\in(0.5,1]$
\begin{align}\label{eq:local-u-q-improvement}
&\limsup_{t\rightarrow\infty}\frac{u_t}{\eta_t+\left(\varphi_t/\eta_t\right)^{2(q+1)}}\lesssim
\begin{cases}
\displaystyle\frac{C_{u,q+1}(k,r)}{2C_\varphi(1-\varepsilon_0)\lambda_m}, & \varphi\in(0.5,1),\\[10pt]
\displaystyle\frac{C_{u,q+1}(k,r)}{2C_\varphi(1-\varepsilon_0)\lambda_m-\min\{\eta,\,2(q+1)(\varphi-\eta)\}}, & \varphi=1,
\end{cases} \nonumber\\
& \Longrightarrow\qquad u_t\lesssim\eta_t+\left(\frac{\varphi_t}{\eta_t}\right)^{2(q+1)}.
\end{align}
When $\varphi=1$, the denominator is positive since
\begin{equation*}
\min\{\eta,\,2(q+1)(\varphi-\eta)\}\leq\eta<2C_\varphi(1-\varepsilon_0)\lambda_m,
\end{equation*}
which is ensured by the choice of $\varepsilon_0$ in~\eqref{eq:local-harmonic-margin}.

Combining the above bound with~\eqref{eq:local-v-q-improvement}, we obtain~\eqref{eq:local-finite-step-rate} with $q$ replaced by $q+1$. This completes the induction and hence the proof.

\subsection{Proof of Theorem~\ref{thm:sketch-noise-reduction}}\label{app:sketch-noise-reduction}

Recall that $\mathcal R\coloneqq I-K^\star$, which is symmetric positive definite. Fix $r>0$ {\red as chosen in the proof of Lemma~\ref{lem:local-rate}} and choose $k=k(\epsilon,r)$. The localized rates established in~\eqref{eq:local-stopped-primary-rates} and~\eqref{eq:local-stopped-Hessian-rate} give
\begin{equation}\label{eq:effective-localized-moments}
\max\{\mE[\|\bDelta_t\|^2\1_{\{\tau_{k,r}>t\}}],\; \mE[\|\be_t\|^2\1_{\{\tau_{k,r}>t\}}], \; \mE[\|B_t-B^\star\|^2\1_{\{\tau_{k,r}>t\}}]\}\lesssim\eta_t.
\end{equation}
Since $\bbm_t=\be_t+\nabla f_t$ and $\|\nabla f_t\|\leq\Upsilon_H\|\bDelta_t\|$, it follows that
\begin{align}\label{eq:effective-localized-m-moment}
\mE[\|\bbm_t\|^2\1_{\{\tau_{k,r}>t\}}] &\leq 2\mE[\|\be_t\|^2\1_{\{\tau_{k,r}>t\}}]+2\Upsilon_H^2\mE[\|\bDelta_t\|^2\1_{\{\tau_{k,r}>t\}}]\lesssim\eta_t.
\end{align}

Since $\bx_t\stackrel{a.s.}{\rightarrow}\bx^\star$ and $\xi_t$ is independent of $\mF_{t-1}$, Assumptions~\ref{ass:1} and~\ref{ass:2} as well as Fatou's lemma together imply that $\mE[\|\nabla F(\bx^\star;\xi)\|^{q_g}]\leq C_{g,2}$. The $q_g$-moment bound also gives uniform integrability, so the unbiasedness extends to $\bx^\star$. Using $\nabla f(\bx^\star)=\zero$, we obtain
\begin{align}\label{eq:effective-gradient-remainders}
\mE[\nabla F(\bx^\star;\xi_t)\mid\mF_{t-1}] = \zero,
\qquad
\mE[(\bg_t-\nabla f_t)-\nabla F(\bx^\star;\xi_t)\mid\mF_{t-1}]=\zero.
\end{align}
{\red Moreover, since $\bx_t\stackrel{a.s.}{\rightarrow}\bx^\star$, the continuity of $\nabla F(\cdot;\xi)$ and $\nabla f$ gives pointwise convergence of the gradient difference to zero. The $q_g$-moment bounds above with $q_g>2$ imply uniform integrability. Since $\xi_t$ is independent of $\mF_{t-1}$, we obtain
\begin{equation*}
\mE[\|(\bg_t-\nabla f_t)-\nabla F(\bx^\star;\xi_t)\|^2\mid\mF_{t-1}]\stackrel{a.s.}{\rightarrow}0.
\end{equation*}
On $\{\tau_{k,r}>t\}$, Assumption~\ref{ass:2} and $\|\bDelta_t\|\leq r$ bound this conditional second moment by $2(C_{g,1}r^{q_g}+C_{g,2})^{2/q_g}+2C_{g,2}^{2/q_g}$. Since $\{\tau_{k,r}>t\}\in\mF_{t-1}$, dominated convergence yields}
\begin{equation}\label{eq:effective-sample-gradient-remainder}
{\red \mE[\|(\bg_t-\nabla f_t)-\nabla F(\bx^\star;\xi_t)\|^2\1_{\{\tau_{k,r}>t\}}]=o(1).}
\end{equation}

By~\eqref{eq:local-Q-conditional-mean}, $B_t^{-1}\bbm_t$ is $\mF_{t-0.5}$-measurable{\red , while $\1_{\{\tau_{k,r}>t\}}$ is $\mF_{t-1}$-measurable,} and hence
\begin{equation*}
\mE[(\tK_t-K_t)B_t^{-1}\bbm_t{\red \1_{\{\tau_{k,r}>t\}}}\mid\mF_{t-0.5}]
=\mE[\tK_t-K_t\mid\mF_{t-0.5}]B_t^{-1}\bbm_t{\red \1_{\{\tau_{k,r}>t\}}}=\zero.
\end{equation*}
Then, by the tower property and~\eqref{eq:effective-gradient-remainders}, we have
\begin{multline}\label{eq:effective-centered-mds}
\mE\left[(\tK_t-K_t)B_t^{-1}\bbm_t{\red \1_{\{\tau_{k,r}>t\}}}
-\mathcal R(B^\star)^{-1}\{(\bg_t-\nabla f_t)-\nabla F(\bx^\star;\xi_t)\}{\red \1_{\{\tau_{k,r}>t\}}}\mid\mF_{t-1}\right] \\
=\mE\left[\mE[(\tK_t-K_t)B_t^{-1}\bbm_t{\red \1_{\{\tau_{k,r}>t\}}}\mid\mF_{t-0.5}]\mid\mF_{t-1}\right] \\
-\mathcal R(B^\star)^{-1}{\red \1_{\{\tau_{k,r}>t\}}}
\mE[(\bg_t-\nabla f_t)-\nabla F(\bx^\star;\xi_t)\mid\mF_{t-1}]=\zero.
\end{multline}
As a result, by~\eqref{eq:effective-centered-mds}, we could conclude that the expression inside the conditional expectation forms a martingale difference sequence.

We next control its localized second moment. By \eqref{eq:global-uniform-tK},~\eqref{eq:effective-sample-gradient-remainder}, the Cauchy--Schwarz inequality {\red and the bounds $\|K_t\|\leq1/4$, $\|(B^\star)^{-1}\|\leq1/\gamma_H$, and $\|B_t^{-1}\|\leq2/\gamma_H$ on $\{\tau_{k,r}>t\}$}, we obtain
\begin{align}\label{eq:effective-centered-moment}
&\mE\left[\left\|(\tK_t-K_t)B_t^{-1}\bbm_t-\mathcal R(B^\star)^{-1}\{(\bg_t-\nabla f_t)-\nabla F(\bx^\star;\xi_t)\}\right\|^2\1_{\{\tau_{k,r}>t\}}\right] \nonumber\\
&\leq 2\mE[\|(\tK_t-K_t)B_t^{-1}\bbm_t\|^2\1_{\{\tau_{k,r}>t\}}]+2\mE[\|\mathcal R(B^\star)^{-1}\{(\bg_t-\nabla f_t)-\nabla F(\bx^\star;\xi_t)\}\|^2\1_{\{\tau_{k,r}>t\}}] \nonumber\\
&\leq {\red \frac{8}{\gamma_H^2}\left(C_K+\frac{1}{4}\right)^2}\mE[\|\bbm_t\|^2\1_{\{\tau_{k,r}>t\}}]+{\red \frac{2\|\mathcal R\|^2}{\gamma_H^2}\mE[\|(\bg_t-\nabla f_t)-\nabla F(\bx^\star;\xi_t)\|^2\1_{\{\tau_{k,r}>t\}}]=o(1),}
\end{align}
where the last inequality follows from~\eqref{eq:effective-localized-m-moment} {\red and~\eqref{eq:effective-sample-gradient-remainder}}.

Moreover, {\red on $\{\tau_{k,r}>t\}$,} by the definition of $Q_t$,~\eqref{eq:local-Q-perturbation-bound}, and $\|(B^\star)^{-1}\|\leq1/\gamma_H$, we obtain
\begin{align}\label{eq:effective-operator-perturbation}
&\left\|(I-K_t)B_t^{-1}-\mathcal R(B^\star)^{-1}\right\| = \left\|\left\{(I-K_t)B_t^{-1}B^\star-\mathcal R\right\}(B^\star)^{-1}\right\| \nonumber\\
&= \left\|(Q_tB^\star-\mathcal R)(B^\star)^{-1}\right\| \leq \|Q_tB^\star-\mathcal R\|\|(B^\star)^{-1}\| \nonumber\\
&\leq \frac{1}{\gamma_H}\left\{{\red \frac{5}{2\gamma_H}}+L_K\right\}\|B_t-B^\star\| \lesssim\|B_t-B^\star\|.
\end{align}
Additionally, by the triangle inequality,~\eqref{eq:local-gradient-linearization}, and~\eqref{eq:effective-operator-perturbation}, we have
\begin{align}\label{eq:effective-noncentered-moment}
&\mE\left[\left\|\mathcal R(B^\star)^{-1}(\nabla f_t-B^\star\bDelta_t)+\left\{(I-K_t)B_t^{-1}-\mathcal R(B^\star)^{-1}\right\}\bbm_t\right\|\1_{\{\tau_{k,r}>t\}}\right] \nonumber\\
&\leq \|\mathcal R\|\|(B^\star)^{-1}\|\mE[\|\nabla f_t-B^\star\bDelta_t\|\1_{\{\tau_{k,r}>t\}}] \nonumber\\
&\quad+\mE\left[\left\|(I-K_t)B_t^{-1}-\mathcal R(B^\star)^{-1}\right\|\|\bbm_t\|\1_{\{\tau_{k,r}>t\}}\right] \nonumber\\
&\lesssim \mE[\|\bDelta_t\|^2\1_{\{\tau_{k,r}>t\}}]+\mE[\|B_t-B^\star\|\|\bbm_t\|\1_{\{\tau_{k,r}>t\}}] \nonumber\\
&\leq \mE[\|\bDelta_t\|^2\1_{\{\tau_{k,r}>t\}}]+\{\mE[\|B_t-B^\star\|^2\1_{\{\tau_{k,r}>t\}}]\}^{1/2}\{\mE[\|\bbm_t\|^2\1_{\{\tau_{k,r}>t\}}]\}^{1/2} \lesssim \eta_t,
\end{align}
where the second-to-last inequality is due to the Cauchy--Schwarz inequality, and the last inequality directly follows from~\eqref{eq:effective-localized-moments} and~\eqref{eq:effective-localized-m-moment}.

\noindent $\bullet$ \textbf{Proof of~\eqref{eq:last-effective-noise}.}
The moving-average update implies the following identity
\begin{equation*}
\bbm_t=\bg_t-\frac{1-\eta_t}{\eta_t}(\bbm_t-\bbm_{t-1}).
\end{equation*}
Moreover, we have
\begin{equation*}
\bg_t=B^\star\bDelta_t+\nabla F(\bx^\star;\xi_t)+\{(\bg_t-\nabla f_t)-\nabla F(\bx^\star;\xi_t)\}+(\nabla f_t-B^\star\bDelta_t).
\end{equation*}

Substituting the above two identities into the outer update and rearranging terms, we obtain
\begin{align}\label{eq:effective-direct-recursion}
\bDelta_{t+1} &= \bDelta_t-\varphi_t(I-\tK_t)B_t^{-1}\bbm_t \nonumber\\
&= \bDelta_t-\varphi_t\mathcal R(B^\star)^{-1}\bbm_t+\varphi_t(\tK_t-K_t)B_t^{-1}\bbm_t-\varphi_t\left\{(I-K_t)B_t^{-1}-\mathcal R(B^\star)^{-1}\right\}\bbm_t \nonumber\\
&= \bDelta_t-\varphi_t\mathcal R(B^\star)^{-1}\bg_t+\varphi_t\frac{1-\eta_t}{\eta_t}\mathcal R(B^\star)^{-1}(\bbm_t-\bbm_{t-1})+\varphi_t(\tK_t-K_t)B_t^{-1}\bbm_t \nonumber\\
&\quad-\varphi_t\left\{(I-K_t)B_t^{-1}-\mathcal R(B^\star)^{-1}\right\}\bbm_t \nonumber\\
&= (I-\varphi_t\mathcal R)\bDelta_t-\varphi_t\mathcal R(B^\star)^{-1}\nabla F(\bx^\star;\xi_t)+\varphi_t\frac{1-\eta_t}{\eta_t}\mathcal R(B^\star)^{-1}(\bbm_t-\bbm_{t-1}) \nonumber\\
&\quad+\varphi_t\left[(\tK_t-K_t)B_t^{-1}\bbm_t-\mathcal R(B^\star)^{-1}\{(\bg_t-\nabla f_t)-\nabla F(\bx^\star;\xi_t)\}\right] \nonumber\\
&\quad-\varphi_t\left[\mathcal R(B^\star)^{-1}(\nabla f_t-B^\star\bDelta_t)+\left\{(I-K_t)B_t^{-1}-\mathcal R(B^\star)^{-1}\right\}\bbm_t\right].
\end{align}
Iterating~\eqref{eq:effective-direct-recursion} from the fixed index $k$ yields
\begin{align}\label{eq:effective-direct-unrolling}
\bDelta_t &= \Pi_{t,k}\bDelta_k-\sum_{i=k}^{t-1}\Pi_{t,i+1}\varphi_i\mathcal R(B^\star)^{-1}\nabla F(\bx^\star;\xi_i) \nonumber\\
&\quad+\sum_{i=k}^{t-1}\Pi_{t,i+1}\varphi_i\frac{1-\eta_i}{\eta_i}\mathcal R(B^\star)^{-1}(\bbm_i-\bbm_{i-1}) \nonumber\\
&\quad+\sum_{i=k}^{t-1}\Pi_{t,i+1}\varphi_i\left[(\tK_i-K_i)B_i^{-1}\bbm_i-\mathcal R(B^\star)^{-1}\{(\bg_i-\nabla f_i)-\nabla F(\bx^\star;\xi_i)\}\right] \nonumber\\
&\quad-\sum_{i=k}^{t-1}\Pi_{t,i+1}\varphi_i\left[\mathcal R(B^\star)^{-1}(\nabla f_i-B^\star\bDelta_i)+\left\{(I-K_i)B_i^{-1}-\mathcal R(B^\star)^{-1}\right\}\bbm_i\right].
\end{align}

We next derive bounds for $\Pi_{t,i}$ and the weighted sums appearing in~\eqref{eq:effective-direct-unrolling}.

Without loss of generality, we may assume that $\varphi_j\|\mathcal R\|\leq1$ for every $j\geq k$. Since $\mathcal R$ is symmetric positive definite, for every $i\geq k$, we obtain
\begin{equation}\label{eq:effective-transition-contraction}
\|I-\varphi_j\mathcal R\| = 1-\lambda_m\varphi_j,
\qquad
\|\Pi_{n,i}\| \leq \prod_{j=i}^{n-1}(1-\lambda_m\varphi_j).
\end{equation}

We first consider the case where $\varphi<1$.
For every fixed $i_0\geq k$, by~\eqref{eq:effective-transition-contraction}, we obtain
\begin{align*}
\|\Pi_{t,i_0}\| \leq \exp\left\{-\lambda_m\sum_{j=i_0}^{t-1}\varphi_j\right\} \leq \exp\left[-C\left\{(t+1)^{1-\varphi}-(i_0+1)^{1-\varphi}\right\}\right]=o(\sqrt{\varphi_t}),
\end{align*}
 where $C>0$ is some constant independent of $t$. Since $\eta>\varphi/2$, we have $\eta_t=o(\sqrt{\varphi_t})$. In addition, $\varphi_t/\eta_t\rightarrow0$ implies that
\begin{equation*}
\frac{\varphi_t}{\sqrt{\eta_t}} = \sqrt{\varphi_t}\sqrt{\frac{\varphi_t}{\eta_t}} = o(\sqrt{\varphi_t}).
\end{equation*}
Therefore, by Lemma~\ref{lem:polynomial_product_rate}, we have
\begin{alignat}{2}\label{eq:effective-kernel-nonharmonic}
\sum_{i=k}^{t-1}\|\Pi_{t,i+1}\|^2\varphi_i^2
&\lesssim \varphi_t,\qquad
& \sum_{i=k}^{t-1}\|\Pi_{t,i+1}\|^2\varphi_i^2\eta_i
&\lesssim \varphi_t\eta_t=o(\varphi_t), \nonumber\\
\sum_{i=k}^{t-1}\|\Pi_{t,i+1}\|\varphi_i\eta_i
&\lesssim \eta_t=o(\sqrt{\varphi_t}),\qquad
& \sum_{i=k}^{t-2}\|\Pi_{t,i+2}\|\frac{\varphi_i^2}{\sqrt{\eta_i}}
&\lesssim \frac{\varphi_t}{\sqrt{\eta_t}}=o(\sqrt{\varphi_t}).
\end{alignat}

Then we consider the case where $\varphi=1$. For every fixed $i_0\geq k$, by~\eqref{eq:effective-transition-contraction}, we obtain
\begin{equation*}
\|\Pi_{t,i_0}\|
\leq \exp\left\{-\lambda_m\sum_{j=i_0}^{t-1}\varphi_j\right\}
\lesssim \left(\frac{i_0+1}{t+1}\right)^{C_\varphi\lambda_m}
\lesssim (t+1)^{-C_\varphi\lambda_m}
=o(t^{-1/2})
=o(\sqrt{\varphi_t}),
\end{equation*}
where the last two relations follow from $C_\varphi\lambda_m>1/2$. Moreover, since $\varphi_t\asymp t^{-1}$, we obtain
\begin{align}\label{eq:effective-kernel-harmonic}
\sum_{i=k}^{t-1}\|\Pi_{t,i+1}\|^2\varphi_i^2
&\lesssim (t+1)^{-2C_\varphi\lambda_m}\sum_{i=k}^{t-1}(i+1)^{2C_\varphi\lambda_m-2}
\lesssim t^{-1}\lesssim\varphi_t, \nonumber\\
\sum_{i=k}^{t-1}\|\Pi_{t,i+1}\|^2\varphi_i^2\eta_i
&\lesssim (t+1)^{-2C_\varphi\lambda_m}\sum_{i=k}^{t-1}(i+1)^{2C_\varphi\lambda_m-2-\eta} \nonumber\\
&\lesssim t^{-\min\{2C_\varphi\lambda_m,\,1+\eta\}}(1+\log t)
=o(t^{-1})=o(\varphi_t), \nonumber\\
\sum_{i=k}^{t-1}\|\Pi_{t,i+1}\|\varphi_i\eta_i
&\lesssim (t+1)^{-C_\varphi\lambda_m}\sum_{i=k}^{t-1}(i+1)^{C_\varphi\lambda_m-1-\eta} \nonumber\\
&\lesssim t^{-\min\{C_\varphi\lambda_m,\,\eta\}}(1+\log t)
=o(t^{-1/2})=o(\sqrt{\varphi_t}), \nonumber\\
\sum_{i=k}^{t-2}\|\Pi_{t,i+2}\|\frac{\varphi_i^2}{\sqrt{\eta_i}}
&\lesssim (t+1)^{-C_\varphi\lambda_m}\sum_{i=k}^{t-2}(i+1)^{C_\varphi\lambda_m-2+\eta/2} \nonumber\\
&\lesssim t^{-\min\{C_\varphi\lambda_m,\,1-\eta/2\}}(1+\log t)
=o(t^{-1/2})=o(\sqrt{\varphi_t}).
\end{align}
Here, $1+\log t$ is included to cover the situation in which the exponent of a corresponding power sum equals $-1$; the little-$o$ relations follow from $C_\varphi\lambda_m>1/2$, $\eta>1/2$ and $1-\eta/2>1/2$.

Combining the cases $\varphi<1$ and $\varphi=1$, for every fixed $i_0\geq k$, we obtain $\|\Pi_{n,i_0}\| = o(\sqrt{\varphi_n})$ and
\begin{alignat}{2}\label{eq:effective-kernel-summary}
\sum_{i=k}^{t-1}\|\Pi_{t,i+1}\|^2\varphi_i^2 & \lesssim \varphi_t,
\qquad
& \sum_{i=k}^{t-1}\|\Pi_{t,i+1}\|^2\varphi_i^2\eta_i &= o(\varphi_t), \nonumber\\
\sum_{i=k}^{t-1}\|\Pi_{t,i+1}\|\varphi_i\eta_i &= o(\sqrt{\varphi_t}),\qquad
& \sum_{i=k}^{t-2}\|\Pi_{t,i+2}\|\frac{\varphi_i^2}{\sqrt{\eta_i}} &= o(\sqrt{\varphi_t}).
\end{alignat}

In~\eqref{eq:effective-direct-unrolling}, the second term on the right-hand side is the leading stochastic term. It remains to show the other three sums are $o_p(\sqrt{\varphi_t})$. We control these sums in the same order as they appear in~\eqref{eq:effective-direct-unrolling}.

We first consider the moving-average difference term. For the prescribed sequences, we have
\begin{equation*}
\varphi_t\frac{1-\eta_t}{\eta_t}
=\frac{C_\varphi}{C_\eta}(t+1)^{-(\varphi-\eta)}
-C_\varphi(t+1)^{-\varphi}
\lesssim \frac{\varphi_t}{\eta_t}.
\end{equation*}
Furthermore, by the mean-value theorem and $\varphi\leq1$, we have
\begin{equation}\label{eq:effective-ema-coefficient-bounds}
\left|\varphi_{t+1}\frac{1-\eta_{t+1}}{\eta_{t+1}}-\varphi_t\frac{1-\eta_t}{\eta_t}\right| \lesssim (t+1)^{-(\varphi-\eta)-1}+(t+1)^{-\varphi-1}\lesssim\frac{\varphi_t^2}{\eta_t}.
\end{equation}
Applying discrete summation by parts to the third term on the right-hand side of~\eqref{eq:effective-direct-unrolling}, we obtain
\begin{multline}\label{eq:effective-ema-abel}
\sum_{i=k}^{t-1}\Pi_{t,i+1}\varphi_i\frac{1-\eta_i}{\eta_i}\mathcal R(B^\star)^{-1}(\bbm_i-\bbm_{i-1}) \\
= \frac{\varphi_{t-1}(1-\eta_{t-1})}{\eta_{t-1}}\mathcal R(B^\star)^{-1}\bbm_{t-1}-\Pi_{t,k+1}\frac{\varphi_k(1-\eta_k)}{\eta_k}\mathcal R(B^\star)^{-1}\bbm_{k-1} \\
+\sum_{i=k}^{t-2}\left\{\Pi_{t,i+1}\frac{\varphi_i(1-\eta_i)}{\eta_i}-\Pi_{t,i+2}\frac{\varphi_{i+1}(1-\eta_{i+1})}{\eta_{i+1}}\right\}\mathcal R(B^\star)^{-1}\bbm_i.
\end{multline}
Since $\Pi_{t,i+1}=\Pi_{t,i+2}(I-\varphi_{i+1}\mathcal R)$, for the last line of~\eqref{eq:effective-ema-abel}, we have
\begin{multline}\label{eq:effective-ema-coefficient-difference}
\left\|\Pi_{t,i+1}\frac{\varphi_i(1-\eta_i)}{\eta_i}-\Pi_{t,i+2}\frac{\varphi_{i+1}(1-\eta_{i+1})}{\eta_{i+1}}\right\| \\
\leq \|\Pi_{t,i+2}\|\left\{\left|\frac{\varphi_i(1-\eta_i)}{\eta_i}-\frac{\varphi_{i+1}(1-\eta_{i+1})}{\eta_{i+1}}\right|+\varphi_{i+1}\frac{\varphi_i(1-\eta_i)}{\eta_i}\|\mathcal R\|\right\} \lesssim \|\Pi_{t,i+2}\|\frac{\varphi_i^2}{\eta_i}.
\end{multline}
where the last inequality follows from~\eqref{eq:effective-ema-coefficient-bounds}.

Since $\1_{\{\tau_{k,r}>t\}}\leq\1_{\{\tau_{k,r}>i\}}$ for every $i\leq t$, Jensen's inequality direclty yields
\begin{equation*}
\mE[\|\bbm_i\|\1_{\{\tau_{k,r}>t\}}]
\leq \left\{\mE[\|\bbm_i\|^2\1_{\{\tau_{k,r}>i\}}]\right\}^{1/2}.
\end{equation*}
Therefore, applying the triangle inequality to~\eqref{eq:effective-ema-abel} and then using~\eqref{eq:effective-localized-m-moment} and~\eqref{eq:effective-ema-coefficient-difference}, we obtain
\begin{align}\label{eq:effective-ema-negligible}
&\mE\left[\left\|\sum_{i=k}^{t-1}\Pi_{t,i+1}\varphi_i\frac{1-\eta_i}{\eta_i}\mathcal R(B^\star)^{-1}(\bbm_i-\bbm_{i-1})\right\|\1_{\{\tau_{k,r}>t\}}\right] \nonumber\\
&\leq \frac{\varphi_{t-1}(1-\eta_{t-1})}{\eta_{t-1}}\|\mathcal R(B^\star)^{-1}\|\mE[\|\bbm_{t-1}\|\1_{\{\tau_{k,r}>t\}}] \nonumber\\
&\quad+\|\Pi_{t,k+1}\|\frac{\varphi_k(1-\eta_k)}{\eta_k}\|\mathcal R(B^\star)^{-1}\|\mE[\|\bbm_{k-1}\|\1_{\{\tau_{k,r}>t\}}] \nonumber\\
&\quad+\sum_{i=k}^{t-2}\left\|\Pi_{t,i+1}\frac{\varphi_i(1-\eta_i)}{\eta_i}-\Pi_{t,i+2}\frac{\varphi_{i+1}(1-\eta_{i+1})}{\eta_{i+1}}\right\|\|\mathcal R(B^\star)^{-1}\|\mE[\|\bbm_i\|\1_{\{\tau_{k,r}>t\}}] \nonumber\\
&\lesssim \frac{\varphi_{t-1}}{\eta_{t-1}}\left\{\mE[\|\bbm_{t-1}\|^2\1_{\{\tau_{k,r}>t-1\}}]\right\}^{1/2}+\|\Pi_{t,k+1}\|{\red \mE[\|\bbm_{k-1}\|\1_{\{\tau_{k,r}>k\}}]} \nonumber\\
&\quad+\sum_{i=k}^{t-2}\|\Pi_{t,i+2}\|\frac{\varphi_i^2}{\eta_i}\left\{\mE[\|\bbm_i\|^2\1_{\{\tau_{k,r}>i\}}]\right\}^{1/2} \nonumber\\
&\lesssim \frac{\varphi_{t-1}}{\sqrt{\eta_{t-1}}}+\|\Pi_{t,k+1}\|+\sum_{i=k}^{t-2}\|\Pi_{t,i+2}\|\frac{\varphi_i^2}{\sqrt{\eta_i}}=o(\sqrt{\varphi_t}).
\end{align}
where the last two steps follow from~\eqref{eq:effective-localized-m-moment},~\eqref{eq:effective-kernel-summary} and $\varphi_t/\eta_t \rightarrow 0$.
{\red On the event $\{\tau_{k,r}>k\}$, $\|\bbm_{k-1}\|\leq\|\be_{k-1}\|+\Upsilon_H\|\bDelta_{k-1}\|\leq(1+\Upsilon_H)r$.}

We next control the centered sum in~\eqref{eq:effective-direct-unrolling}. By~\eqref{eq:effective-centered-mds}, the bracketed expressions {\red multiplied by $\1_{\{\tau_{k,r}>i\}}$} form a martingale difference sequence. Therefore, by~\eqref{eq:effective-centered-moment} and~\eqref{eq:effective-kernel-summary}, we have
\begin{align}\label{eq:effective-martingale-negligible}
&\mE\left[\left\|\sum_{i=k}^{t-1}\Pi_{t,i+1}\varphi_i\left[(\tK_i-K_i)B_i^{-1}\bbm_i-\mathcal R(B^\star)^{-1}\{(\bg_i-\nabla f_i)-\nabla F(\bx^\star;\xi_i)\}\right]\right\|^2\1_{\{\tau_{k,r}>t\}}\right] \nonumber\\
&\leq \mE\left[\left\|\sum_{i=k}^{t-1}\Pi_{t,i+1}\varphi_i\left[(\tK_i-K_i)B_i^{-1}\bbm_i-\mathcal R(B^\star)^{-1}\{(\bg_i-\nabla f_i)-\nabla F(\bx^\star;\xi_i)\}\right]\1_{\{\tau_{k,r}>i\}}\right\|^2\right] \nonumber\\
&= \sum_{i=k}^{t-1}\mE\left[\left\|\Pi_{t,i+1}\varphi_i\left[(\tK_i-K_i)B_i^{-1}\bbm_i-\mathcal R(B^\star)^{-1}\{(\bg_i-\nabla f_i)-\nabla F(\bx^\star;\xi_i)\}\right]\right\|^2\1_{\{\tau_{k,r}>i\}}\right] \nonumber\\
&\leq \sum_{i=k}^{t-1}\|\Pi_{t,i+1}\|^2\varphi_i^2\mE\left[\left\|(\tK_i-K_i)B_i^{-1}\bbm_i-\mathcal R(B^\star)^{-1}\{(\bg_i-\nabla f_i)-\nabla F(\bx^\star;\xi_i)\}\right\|^2\1_{\{\tau_{k,r}>i\}}\right] \nonumber\\
&\lesssim \sum_{i=k}^{t-1}\|\Pi_{t,i+1}\|^2\varphi_i^2\eta_i {\red +\sum_{i=k}^{t-1}\|\Pi_{t,i+1}\|^2\varphi_i^2
\mE[\|(\bg_i-\nabla f_i)-\nabla F(\bx^\star;\xi_i)\|^2\1_{\{\tau_{k,r}>i\}}]}=o(\varphi_t).
\end{align}
{\red The first sum is $o(\varphi_t)$ by~\eqref{eq:effective-kernel-summary}. As for the second one, \eqref{eq:effective-sample-gradient-remainder} implies that, for any $\varepsilon>0$, there exists $N>k$ such that $\mE[\|(\bg_i-\nabla f_i)-\nabla F(\bx^\star;\xi_i)\|^2\1_{\{\tau_{k,r}>i\}}]\leq\varepsilon$, $i\geq N$.
Therefore,
\begin{equation*}
\sum_{i=N}^{t-1}\|\Pi_{t,i+1}\|^2\varphi_i^2
\mE[\|(\bg_i-\nabla f_i)-\nabla F(\bx^\star;\xi_i)\|^2\1_{\{\tau_{k,r}>i\}}] \leq \varepsilon\sum_{i=N}^{t-1}\|\Pi_{t,i+1}\|^2\varphi_i^2 \lesssim \varepsilon\varphi_t.
\end{equation*}
For the finitely many terms with $i<N$, $\|\Pi_{t,i+1}\|=o(\sqrt{\varphi_t})$, so their sum is $o(\varphi_t)$. Then, since $\varepsilon$ is arbitrary, the second sum is also $o(\varphi_t)$.}

For the last sum in~\eqref{eq:effective-direct-unrolling}, by the triangle inequality, ~\eqref{eq:effective-noncentered-moment} and~\eqref{eq:effective-kernel-summary}, we obtain
\begin{align}\label{eq:effective-remainder-negligible}
&\mE\left[\left\|\sum_{i=k}^{t-1}\Pi_{t,i+1}\varphi_i\left[\mathcal R(B^\star)^{-1}(\nabla f_i-B^\star\bDelta_i)+\left\{(I-K_i)B_i^{-1}-\mathcal R(B^\star)^{-1}\right\}\bbm_i\right]\right\|\1_{\{\tau_{k,r}>t\}}\right] \nonumber\\
&\leq \sum_{i=k}^{t-1}\|\Pi_{t,i+1}\|\varphi_i\mE\left[\left\|\mathcal R(B^\star)^{-1}(\nabla f_i-B^\star\bDelta_i)+\left\{(I-K_i)B_i^{-1}-\mathcal R(B^\star)^{-1}\right\}\bbm_i\right\|\1_{\{\tau_{k,r}>i\}}\right] \nonumber\\
&\lesssim \sum_{i=k}^{t-1}\|\Pi_{t,i+1}\|\varphi_i\eta_i=o(\sqrt{\varphi_t}).
\end{align}

Equations~\eqref{eq:effective-ema-negligible}, \eqref{eq:effective-martingale-negligible}, and~\eqref{eq:effective-remainder-negligible} control the three remainder sums in~\eqref{eq:effective-direct-unrolling} on $\{\tau_{k,r}>t\}$. We next remove the stopping time. For every $\zeta>0$,~\eqref{eq:effective-direct-unrolling} gives us the following:
\begin{align*}
&\PP\left(\left\|\bDelta_t-\Pi_{t,k}\bDelta_k+\sum_{i=k}^{t-1}\Pi_{t,i+1}\varphi_i\mathcal R(B^\star)^{-1}\nabla F(\bx^\star;\xi_i)\right\|>\frac{\zeta\sqrt{\varphi_t}}{2}\right) \\
&\leq \PP(\tau_{k,r}\leq t)+\PP\left(\left\|\bDelta_t-\Pi_{t,k}\bDelta_k+\sum_{i=k}^{t-1}\Pi_{t,i+1}\varphi_i\mathcal R(B^\star)^{-1}\nabla F(\bx^\star;\xi_i)\right\|\1_{\{\tau_{k,r}>t\}}>\frac{\zeta\sqrt{\varphi_t}}{2}\right).
\end{align*}
For the fixed $(k,r)$, the vector inside the second probability is the sum of the three remainder terms in~\eqref{eq:effective-direct-unrolling}. By the triangle inequality and the union bound, it suffices to show that the probability associated with each remainder term, with threshold $\zeta\sqrt{\varphi_t}/6$, converges to zero.

For the moving-average difference term, by Markov's inequality and~\eqref{eq:effective-ema-negligible}, we have
\begin{align*}
&\PP\left(\left\|\sum_{i=k}^{t-1}\Pi_{t,i+1}\varphi_i\frac{1-\eta_i}{\eta_i}\mathcal R(B^\star)^{-1}(\bbm_i-\bbm_{i-1})\right\|\1_{\{\tau_{k,r}>t\}}>\frac{\zeta\sqrt{\varphi_t}}{6}\right) \\
&\leq \frac{6}{\zeta\sqrt{\varphi_t}}\mE\left[\left\|\sum_{i=k}^{t-1}\Pi_{t,i+1}\varphi_i\frac{1-\eta_i}{\eta_i}\mathcal R(B^\star)^{-1}(\bbm_i-\bbm_{i-1})\right\|\1_{\{\tau_{k,r}>t\}}\right]=o(1).
\end{align*}

For the centered term, by Chebyshev's inequality and~\eqref{eq:effective-martingale-negligible}, we have
\begin{align*}
&\PP\left(\left\|\sum_{i=k}^{t-1}\Pi_{t,i+1}\varphi_i\left[(\tK_i-K_i)B_i^{-1}\bbm_i-\mathcal R(B^\star)^{-1}\{(\bg_i-\nabla f_i)-\nabla F(\bx^\star;\xi_i)\}\right]\right\|\1_{\{\tau_{k,r}>t\}}>\frac{\zeta\sqrt{\varphi_t}}{6}\right) \\
&\leq \frac{36}{\zeta^2\varphi_t}\mE\left[\left\|\sum_{i=k}^{t-1}\Pi_{t,i+1}\varphi_i\left[(\tK_i-K_i)B_i^{-1}\bbm_i-\mathcal R(B^\star)^{-1}\{(\bg_i-\nabla f_i)-\nabla F(\bx^\star;\xi_i)\}\right]\right\|^2\1_{\{\tau_{k,r}>t\}}\right]=o(1).
\end{align*}

For the noncentered term, by Markov's inequality and~\eqref{eq:effective-remainder-negligible}, we have
\begin{align*}
&\PP\left(\left\|\sum_{i=k}^{t-1}\Pi_{t,i+1}\varphi_i\left[\mathcal R(B^\star)^{-1}(\nabla f_i-B^\star\bDelta_i)+\left\{(I-K_i)B_i^{-1}-\mathcal R(B^\star)^{-1}\right\}\bbm_i\right]\right\|\1_{\{\tau_{k,r}>t\}}>\frac{\zeta\sqrt{\varphi_t}}{6}\right) \\
&\leq \frac{6}{\zeta\sqrt{\varphi_t}}\mE\left[\left\|\sum_{i=k}^{t-1}\Pi_{t,i+1}\varphi_i\left[\mathcal R(B^\star)^{-1}(\nabla f_i-B^\star\bDelta_i)+\left\{(I-K_i)B_i^{-1}-\mathcal R(B^\star)^{-1}\right\}\bbm_i\right]\right\|\1_{\{\tau_{k,r}>t\}}\right]=o(1).
\end{align*}

Since second probability in the stopping-time decomposition converges to zero, we have
\begin{equation*}
\limsup_{t\rightarrow\infty}\PP\left(\left\|\bDelta_t-\Pi_{t,k}\bDelta_k+\sum_{i=k}^{t-1}\Pi_{t,i+1}\varphi_i\mathcal R(B^\star)^{-1}\nabla F(\bx^\star;\xi_i)\right\|>\frac{\zeta\sqrt{\varphi_t}}{2}\right)\leq\PP(\tau_{k,r}<\infty)<\frac{\epsilon}{2}.
\end{equation*}

Lastly, we extend the stochastic sum from $i=k$ to $i=0$ and we have
\begin{multline*}
\bDelta_t+\sum_{i=0}^{t-1}\Pi_{t,i+1}\varphi_i\mathcal R(B^\star)^{-1}\nabla F(\bx^\star;\xi_i) \\
= \bDelta_t-\Pi_{t,k}\bDelta_k+\sum_{i=k}^{t-1}\Pi_{t,i+1}\varphi_i\mathcal R(B^\star)^{-1}\nabla F(\bx^\star;\xi_i) \\
\quad+\Pi_{t,k}\left\{\bDelta_k+\sum_{i=0}^{k-1}\Pi_{k,i+1}\varphi_i\mathcal R(B^\star)^{-1}\nabla F(\bx^\star;\xi_i)\right\}.
\end{multline*}
For fixed $k$, the quantity inside the braces does not depend on $t$ and is finite almost surely. Moreover,~\eqref{eq:effective-kernel-summary} gives $\|\Pi_{t,k}\|=o(\sqrt{\varphi_t})$. Therefore, we obtain
\begin{equation*}
\frac{1}{\sqrt{\varphi_t}}\left\|\Pi_{t,k}\left\{\bDelta_k+\sum_{i=0}^{k-1}\Pi_{k,i+1}\varphi_i\mathcal R(B^\star)^{-1}\nabla F(\bx^\star;\xi_i)\right\}\right\|\stackrel{a.s.}{\rightarrow}0.
\end{equation*}
Hence, the second term on the right-hand side of the preceding decomposition is $o_p(\sqrt{\varphi_t})$.

For every $\zeta>0$, the triangle inequality further yields
\begin{multline*}
\PP\left(\left\|\bDelta_t+\sum_{i=0}^{t-1}\Pi_{t,i+1}\varphi_i\mathcal R(B^\star)^{-1}\nabla F(\bx^\star;\xi_i)\right\|>\zeta\sqrt{\varphi_t}\right) \\
\leq \PP\left(\left\|\bDelta_t-\Pi_{t,k}\bDelta_k+\sum_{i=k}^{t-1}\Pi_{t,i+1}\varphi_i\mathcal R(B^\star)^{-1}\nabla F(\bx^\star;\xi_i)\right\|>\frac{\zeta\sqrt{\varphi_t}}{2}\right) \\
+\PP\left(\left\|\Pi_{t,k}\left\{\bDelta_k+\sum_{i=0}^{k-1}\Pi_{k,i+1}\varphi_i\mathcal R(B^\star)^{-1}\nabla F(\bx^\star;\xi_i)\right\}\right\|>\frac{\zeta\sqrt{\varphi_t}}{2}\right).
\end{multline*}
For the fixed $k=k(\epsilon,r)$, the preceding bound controls the first probability, whereas the second probability converges to zero by the almost-sure convergence established above. It follows that
\begin{equation*}
\limsup_{t\rightarrow\infty}\PP\left(\left\|\bDelta_t+\sum_{i=0}^{t-1}\Pi_{t,i+1}\varphi_i\mathcal R(B^\star)^{-1}\nabla F(\bx^\star;\xi_i)\right\|>\zeta\sqrt{\varphi_t}\right)\leq\PP(\tau_{k,r}<\infty)<\frac{\epsilon}{2}.
\end{equation*}

Since $\epsilon>0$ is arbitrary and the term inside the probability is independent of $k$, we conclude that
\begin{equation*}
\bDelta_t+\sum_{i=0}^{t-1}\Pi_{t,i+1}\varphi_i\mathcal R(B^\star)^{-1}\nabla F(\bx^\star;\xi_i)=o_p(\sqrt{\varphi_t}).
\end{equation*}
Recalling that $\mathcal R=I-K^\star$ and dividing both sides by $\sqrt{\varphi_t}$ yield
\begin{equation*}
\frac{1}{\sqrt{\varphi_t}}\bDelta_t=-\frac{1}{\sqrt{\varphi_t}}\sum_{i=0}^{t-1}\Pi_{t,i+1}\varphi_i(I-K^\star)(B^\star)^{-1}\nabla F(\bx^\star;\xi_i)+o_p(1).
\end{equation*}
Then we complete the proof of~\eqref{eq:last-effective-noise}.

\noindent $\bullet$ \textbf{Proof of~\eqref{eq:averaged-effective-noise}.}
Suppose throughout this part that $\varphi<1$. We return to the fixed localized system $(k,r)$ used above. Before averaging~\eqref{eq:effective-direct-recursion}, we first derive a localized first-moment bound for $\bDelta_t$.

By Assumptions~\ref{ass:1} and~\ref{ass:2}, $(\nabla F(\bx^\star;\xi_i),\mF_i)_{i\geq0}$ forms a martingale difference sequence and $\mE[\|\nabla F(\bx^\star;\xi_i)\|^2]\lesssim1$. Therefore, by~\eqref{eq:effective-kernel-summary}, we have
\begin{multline}\label{eq:effective-leading-tightness}
\mE\left[\left\|\sum_{i=k}^{t-1}\Pi_{t,i+1}\varphi_i\mathcal R(B^\star)^{-1}\nabla F(\bx^\star;\xi_i)\right\|^2\right] \\
= \sum_{i=k}^{t-1}\mE\left[\left\|\Pi_{t,i+1}\varphi_i\mathcal R(B^\star)^{-1}\nabla F(\bx^\star;\xi_i)\right\|^2\right]
\lesssim \sum_{i=k}^{t-1}\|\Pi_{t,i+1}\|^2\varphi_i^2
\lesssim\varphi_t.
\end{multline}

Taking norms in~\eqref{eq:effective-direct-unrolling}, multiplying by $\1_{\{\tau_{k,r}>t\}}$, and using $\1_{\{\tau_{k,r}>t\}}\leq\1_{\{\tau_{k,r}>i\}}$ for every $i\leq t$, the triangle inequality and the Cauchy--Schwarz inequality yield
\begin{align}\label{eq:effective-strong-local-rate}
&\mE[\|\bDelta_t\|\1_{\{\tau_{k,r}>t\}}]
\leq r\|\Pi_{t,k}\|
+\left\{\mE\left[\left\|\sum_{i=k}^{t-1}\Pi_{t,i+1}\varphi_i\mathcal R(B^\star)^{-1}\nabla F(\bx^\star;\xi_i)\right\|^2\right]\right\}^{1/2} \nonumber\\
&\quad+\mE\left[\left\|\sum_{i=k}^{t-1}\Pi_{t,i+1}\varphi_i\frac{1-\eta_i}{\eta_i}\mathcal R(B^\star)^{-1}(\bbm_i-\bbm_{i-1})\right\|\1_{\{\tau_{k,r}>t\}}\right] \nonumber\\
&\quad+\left\{\mE\left[\left\|\sum_{i=k}^{t-1}\Pi_{t,i+1}\varphi_i\left[(\tK_i-K_i)B_i^{-1}\bbm_i-\mathcal R(B^\star)^{-1}\{(\bg_i-\nabla f_i)-\nabla F(\bx^\star;\xi_i)\}\right]\1_{\{\tau_{k,r}>i\}}\right\|^2\right]\right\}^{1/2} \nonumber\\
&\quad+\mE\left[\left\|\sum_{i=k}^{t-1}\Pi_{t,i+1}\varphi_i\left[\mathcal R(B^\star)^{-1}(\nabla f_i-B^\star\bDelta_i)+\left\{(I-K_i)B_i^{-1}-\mathcal R(B^\star)^{-1}\right\}\bbm_i\right]\right\|\1_{\{\tau_{k,r}>t\}}\right] \nonumber\\
&\lesssim r\|\Pi_{t,k}\|+\sqrt{\varphi_t}+o(\sqrt{\varphi_t})+\{o(\varphi_t)\}^{1/2}+o(\sqrt{\varphi_t})
\lesssim\sqrt{\varphi_t}.
\end{align}
where the first inequality uses $\|\bDelta_k\|\leq r$ on $\{\tau_{k,r}>t\}$, the second inequality follows from~\eqref{eq:effective-ema-negligible}, \eqref{eq:effective-martingale-negligible}, \eqref{eq:effective-remainder-negligible} and~\eqref{eq:effective-leading-tightness}, while the last inequality uses $\|\Pi_{t,k}\|=o(\sqrt{\varphi_t})$ from~\eqref{eq:effective-kernel-summary}.

We now rearrange~\eqref{eq:effective-direct-recursion} as follows,
\begin{align}\label{eq:effective-average-decomposition}
\bDelta_i &= -(B^\star)^{-1}\nabla F(\bx^\star;\xi_i)
-\mathcal R^{-1}\frac{\bDelta_{i+1}-\bDelta_i}{\varphi_i}
+\frac{1-\eta_i}{\eta_i}(B^\star)^{-1}(\bbm_i-\bbm_{i-1}) \nonumber\\
&\quad+\mathcal R^{-1}\left[(\tK_i-K_i)B_i^{-1}\bbm_i-\mathcal R(B^\star)^{-1}\{(\bg_i-\nabla f_i)-\nabla F(\bx^\star;\xi_i)\}\right] \nonumber\\
&\quad-\mathcal R^{-1}\left[\mathcal R(B^\star)^{-1}(\nabla f_i-B^\star\bDelta_i)+\left\{(I-K_i)B_i^{-1}-\mathcal R(B^\star)^{-1}\right\}\bbm_i\right].
\end{align}

Since $k$ is some fixed index, the averaged iterate can be decomposed as
\begin{equation*}
\sqrt t(\bar{\bx}_t-\bx^\star) = \frac{1}{\sqrt t}\sum_{i=0}^{t-1}\bDelta_i = \frac{1}{\sqrt t}\sum_{i=0}^{k-1}\bDelta_i + \frac{1}{\sqrt t}\sum_{i=k}^{t-1}\bDelta_i.
\end{equation*}
We first analyze the second term and restore the fixed finite prefix at the end. Summing~\eqref{eq:effective-average-decomposition} from $i=k$ to $t-1$ and dividing both sides by $\sqrt t$, we obtain
\begin{align}\label{eq:effective-averaged-decomposition}
\frac{1}{\sqrt t}\sum_{i=k}^{t-1}\bDelta_i
&= -\frac{1}{\sqrt t}\sum_{i=k}^{t-1}(B^\star)^{-1}\nabla F(\bx^\star;\xi_i)
-\frac{1}{\sqrt t}\mathcal R^{-1}\sum_{i=k}^{t-1}\frac{\bDelta_{i+1}-\bDelta_i}{\varphi_i} \nonumber\\
&\quad+\frac{1}{\sqrt t}(B^\star)^{-1}\sum_{i=k}^{t-1}\frac{1-\eta_i}{\eta_i}(\bbm_i-\bbm_{i-1}) \nonumber\\
&\quad+\frac{1}{\sqrt t}\mathcal R^{-1}\sum_{i=k}^{t-1}\left[(\tK_i-K_i)B_i^{-1}\bbm_i-\mathcal R(B^\star)^{-1}\{(\bg_i-\nabla f_i)-\nabla F(\bx^\star;\xi_i)\}\right] \nonumber\\
&\quad-\frac{1}{\sqrt t}\mathcal R^{-1}\sum_{i=k}^{t-1}\left[\mathcal R(B^\star)^{-1}(\nabla f_i-B^\star\bDelta_i)+\left\{(I-K_i)B_i^{-1}-\mathcal R(B^\star)^{-1}\right\}\bbm_i\right].
\end{align}
The first term on the right-hand side is the leading stochastic term, and it remains to show that each of the other four terms is $o_p(1)$.

For the second term on the right-hand side of~\eqref{eq:effective-averaged-decomposition}, summation by parts gives us
\begin{equation}\label{eq:effective-iterate-abel}
\sum_{i=k}^{t-1}\frac{\bDelta_{i+1}-\bDelta_i}{\varphi_i} = \frac{\bDelta_t}{\varphi_{t-1}}-\frac{\bDelta_k}{\varphi_k}+\sum_{i=k+1}^{t-1}\left(\frac{1}{\varphi_{i-1}}-\frac{1}{\varphi_i}\right)\bDelta_i.
\end{equation}
Since $\varphi_i^{-1}=(i+1)^\varphi/C_\varphi$, by the mean-value theorem, we have
\begin{equation*}
\left|\frac{1}{\varphi_i}-\frac{1}{\varphi_{i-1}}\right| = \frac{1}{C_\varphi}\left|(i+1)^\varphi-i^\varphi\right| \lesssim (i+1)^{\varphi-1}.
\end{equation*}
Therefore, by~\eqref{eq:effective-strong-local-rate} and $\1_{\{\tau_{k,r}>t\}}\leq\1_{\{\tau_{k,r}>i\}}$, we obtain
\begin{align}\label{eq:effective-iterate-abel-negligible}
&\frac{1}{\sqrt t}\mE\left[\left\|\sum_{i=k}^{t-1}\frac{\bDelta_{i+1}-\bDelta_i}{\varphi_i}\right\|\1_{\{\tau_{k,r}>t\}}\right] \nonumber\\
&\leq \frac{1}{\sqrt t}\left\{\frac{\mE[\|\bDelta_t\|\1_{\{\tau_{k,r}>t\}}]}{\varphi_{t-1}}+\frac{r}{\varphi_k}
+\sum_{i=k+1}^{t-1}\left|\frac{1}{\varphi_{i-1}}-\frac{1}{\varphi_i}\right|\mE[\|\bDelta_i\|\1_{\{\tau_{k,r}>i\}}]\right\} \nonumber\\
&\lesssim \frac{1}{\sqrt t}\left\{\frac{1}{\sqrt{\varphi_t}}+1+\sum_{i=k+1}^{t-1}(i+1)^{\varphi-1}\sqrt{\varphi_i}\right\}
\lesssim\frac{1}{t^{(1-\varphi)/2}}=o(1).
\end{align}
Here, $r/\varphi_k$ follows from $\|\bDelta_k\|\leq r$ on $\{\tau_{k,r}>t\}$.

For the third term on the right-hand side of~\eqref{eq:effective-averaged-decomposition}, again we apply summation by parts and obtain
\begin{multline}\label{eq:effective-ema-average-abel}
\sum_{i=k}^{t-1}\frac{1-\eta_i}{\eta_i}(\bbm_i-\bbm_{i-1})=\frac{1-\eta_{t-1}}{\eta_{t-1}}\bbm_{t-1}-\frac{1-\eta_k}{\eta_k}\bbm_{k-1} \\
+\sum_{i=k}^{t-2}\left(\frac{1-\eta_i}{\eta_i}-\frac{1-\eta_{i+1}}{\eta_{i+1}}\right)\bbm_i.
\end{multline}
In addition, the mean-value theorem gives us
\begin{equation*}
\left|\frac{1-\eta_i}{\eta_i}-\frac{1-\eta_{i+1}}{\eta_{i+1}}\right| \lesssim (i+1)^{\eta-1} \lesssim \frac{1}{(i+1)\eta_i}.
\end{equation*}
Using~\eqref{eq:effective-localized-m-moment}, $\1_{\{\tau_{k,r}>t\}}\leq\1_{\{\tau_{k,r}>i\}}$, and Jensen's inequality, we further obtain
\begin{align}\label{eq:effective-ema-average-negligible}
&\frac{1}{\sqrt t}\mE\left[\left\|\sum_{i=k}^{t-1}\frac{1-\eta_i}{\eta_i}(\bbm_i-\bbm_{i-1})\right\|\1_{\{\tau_{k,r}>t\}}\right] \nonumber\\
&\leq \frac{1}{\sqrt t}\Bigg\{\frac{1}{\eta_{t-1}}\mE[\|\bbm_{t-1}\|\1_{\{\tau_{k,r}>t\}}]+\frac{1-\eta_k}{\eta_k}{\red \mE[\|\bbm_{k-1}\|\1_{\{\tau_{k,r}>k\}}]}+\sum_{i=k}^{t-2}\frac{1}{(i+1)\eta_i}\mE[\|\bbm_i\|\1_{\{\tau_{k,r}>t\}}]\Bigg\} \nonumber\\
&\leq \frac{1}{\sqrt t}\Bigg\{\frac{1}{\eta_{t-1}}\left\{\mE[\|\bbm_{t-1}\|^2\1_{\{\tau_{k,r}>t-1\}}]\right\}^{1/2}
+\frac{1-\eta_k}{\eta_k}{\red \mE[\|\bbm_{k-1}\|\1_{\{\tau_{k,r}>k\}}]} \nonumber\\
&\qquad+\sum_{i=k}^{t-2}\frac{1}{(i+1)\eta_i}\left\{\mE[\|\bbm_i\|^2\1_{\{\tau_{k,r}>i\}}]\right\}^{1/2}\Bigg\} \nonumber\\
&\lesssim \frac{1}{\sqrt t}\left\{\frac{\sqrt{\eta_{t-1}}}{\eta_{t-1}}+1+\sum_{i=k}^{t-2}\frac{\sqrt{\eta_i}}{(i+1)\eta_i}\right\}=\frac{1}{\sqrt t}\left\{\frac{1}{\sqrt{\eta_{t-1}}}+1+\sum_{i=k}^{t-2}\frac{1}{(i+1)\sqrt{\eta_i}}\right\} \nonumber\\
&\lesssim \frac{1}{\sqrt t}\left\{t^{\eta/2}+1+\sum_{i=k}^{t-2}(i+1)^{-1+\eta/2}\right\} \lesssim \frac{1}{\sqrt t}\left\{t^{\eta/2}+1+(t+1)^{\eta/2}\right\} \nonumber\\
&\lesssim \frac{1}{t^{(1-\eta)/2}}=o(1).
\end{align}
Here, {\red $\|\bbm_{k-1}\|\leq(1+\Upsilon_H)r$ on $\{\tau_{k,r}>k\}$, so the fixed boundary term is absorbed into the hidden constant}; the last equality directly follows from $\eta<\varphi<1$.

For the fourth term on the right-hand side of~\eqref{eq:effective-averaged-decomposition}, by~\eqref{eq:effective-centered-mds} and~\eqref{eq:effective-centered-moment}, we obtain
\begin{align}\label{eq:effective-centered-average-negligible}
&\mE\left[\left\|\frac{1}{\sqrt t}\sum_{i=k}^{t-1}\mathcal R^{-1}\left[(\tK_i-K_i)B_i^{-1}\bbm_i-\mathcal R(B^\star)^{-1}\{(\bg_i-\nabla f_i)-\nabla F(\bx^\star;\xi_i)\}\right]\right\|^2\1_{\{\tau_{k,r}>t\}}\right] \nonumber\\
&\leq \mE\left[\left\|\frac{1}{\sqrt t}\sum_{i=k}^{t-1}\mathcal R^{-1}\left[(\tK_i-K_i)B_i^{-1}\bbm_i-\mathcal R(B^\star)^{-1}\{(\bg_i-\nabla f_i)-\nabla F(\bx^\star;\xi_i)\}\right]\1_{\{\tau_{k,r}>i\}}\right\|^2\right] \nonumber\\
&= \frac{1}{t}\sum_{i=k}^{t-1}\mE\left[\left\|\mathcal R^{-1}\left[(\tK_i-K_i)B_i^{-1}\bbm_i-\mathcal R(B^\star)^{-1}\{(\bg_i-\nabla f_i)-\nabla F(\bx^\star;\xi_i)\}\right]\right\|^2\1_{\{\tau_{k,r}>i\}}\right] \nonumber\\
&\lesssim \frac{1}{t}\sum_{i=k}^{t-1}\eta_i {\red +\frac{1}{t}\sum_{i=k}^{t-1}\mE[\|(\bg_i-\nabla f_i)-\nabla F(\bx^\star;\xi_i)\|^2\1_{\{\tau_{k,r}>i\}}]} = o(1).
\end{align}
{\red Here, the first sum is $O(t^{-\eta})$ since $\eta<1$; the second is $o(1)$ by~\eqref{eq:effective-sample-gradient-remainder} and Cesàro's lemma.}

For the final term on the right-hand side of~\eqref{eq:effective-averaged-decomposition}, by the triangle inequality and~\eqref{eq:effective-noncentered-moment}, we obtain
\begin{align}\label{eq:effective-noncentered-average-negligible}
&\mE\left[\left\|\frac{1}{\sqrt t}\sum_{i=k}^{t-1}\mathcal R^{-1}\left[\mathcal R(B^\star)^{-1}(\nabla f_i-B^\star\bDelta_i)+\left\{(I-K_i)B_i^{-1}-\mathcal R(B^\star)^{-1}\right\}\bbm_i\right]\right\|\1_{\{\tau_{k,r}>t\}}\right] \nonumber\\
&\leq \frac{\|\mathcal R^{-1}\|}{\sqrt t}\sum_{i=k}^{t-1}\mE\left[\left\|\mathcal R(B^\star)^{-1}(\nabla f_i-B^\star\bDelta_i)+\left\{(I-K_i)B_i^{-1}-\mathcal R(B^\star)^{-1}\right\}\bbm_i\right\|\1_{\{\tau_{k,r}>i\}}\right] \nonumber\\
&\lesssim \frac{1}{\sqrt t}\sum_{i=k}^{t-1}\eta_i
\lesssim \frac{1}{t^{\eta - 1/2}}=o(1),
\end{align}
where the last equality follows from $\eta>1/2$.

Similarly to the proof of~\eqref{eq:last-effective-noise}, it remains to remove the stopping time and restore the finitely many indices preceding $k$. For every $\zeta>0$, the stopping-time decomposition gives
\begin{align*}
&\PP\left(\left\|\frac{1}{\sqrt t}\sum_{i=k}^{t-1}\bDelta_i+\frac{1}{\sqrt t}\sum_{i=k}^{t-1}(B^\star)^{-1}\nabla F(\bx^\star;\xi_i)\right\|>\frac{\zeta}{2}\right) \\
&\leq \PP(\tau_{k,r}\leq t)
+\PP\left(\left\|\frac{1}{\sqrt t}\sum_{i=k}^{t-1}\bDelta_i+\frac{1}{\sqrt t}\sum_{i=k}^{t-1}(B^\star)^{-1}\nabla F(\bx^\star;\xi_i)\right\|\1_{\{\tau_{k,r}>t\}}>\frac{\zeta}{2}\right).
\end{align*}
On $\{\tau_{k,r}>t\}$, by~\eqref{eq:effective-averaged-decomposition} and the triangle inequality, we have
\begin{align*}
&\left\|\frac{1}{\sqrt t}\sum_{i=k}^{t-1}\bDelta_i+\frac{1}{\sqrt t}\sum_{i=k}^{t-1}(B^\star)^{-1}\nabla F(\bx^\star;\xi_i)\right\|\1_{\{\tau_{k,r}>t\}} \\
&\leq \frac{\|\mathcal R^{-1}\|}{\sqrt t}\left\|\sum_{i=k}^{t-1}\frac{\bDelta_{i+1}-\bDelta_i}{\varphi_i}\right\|\1_{\{\tau_{k,r}>t\}}+\frac{\|(B^\star)^{-1}\|}{\sqrt t}\left\|\sum_{i=k}^{t-1}\frac{1-\eta_i}{\eta_i}(\bbm_i-\bbm_{i-1})\right\|\1_{\{\tau_{k,r}>t\}} \\
&\quad+\left\|\frac{1}{\sqrt t}\sum_{i=k}^{t-1}\mathcal R^{-1}\left[(\tK_i-K_i)B_i^{-1}\bbm_i-\mathcal R(B^\star)^{-1}\{(\bg_i-\nabla f_i)-\nabla F(\bx^\star;\xi_i)\}\right]\right\|\1_{\{\tau_{k,r}>t\}} \\
&\quad+\left\|\frac{1}{\sqrt t}\sum_{i=k}^{t-1}\mathcal R^{-1}\left[\mathcal R(B^\star)^{-1}(\nabla f_i-B^\star\bDelta_i)+\left\{(I-K_i)B_i^{-1}-\mathcal R(B^\star)^{-1}\right\}\bbm_i\right]\right\|\1_{\{\tau_{k,r}>t\}}.
\end{align*}
Therefore, if the left-hand side is greater than $\zeta/2$, then at least one term on the right-hand side must be greater than $\zeta/8$. By the union bound, it remains to control the corresponding four probabilities.

For the iterate-difference term, by Markov's inequality and~\eqref{eq:effective-iterate-abel-negligible}, we have
\begin{multline*}
\PP\left(\frac{\|\mathcal R^{-1}\|}{\sqrt t}\left\|\sum_{i=k}^{t-1}\frac{\bDelta_{i+1}-\bDelta_i}{\varphi_i}\right\|\1_{\{\tau_{k,r}>t\}}>\frac{\zeta}{8}\right) \\
\leq \frac{8\|\mathcal R^{-1}\|}{\zeta\sqrt t}\mE\left[\left\|\sum_{i=k}^{t-1}\frac{\bDelta_{i+1}-\bDelta_i}{\varphi_i}\right\|\1_{\{\tau_{k,r}>t\}}\right]=o(1).
\end{multline*}

For the moving-average difference term, by Markov's inequality and~\eqref{eq:effective-ema-average-negligible}, we have
\begin{multline*}
\PP\left(\frac{\|(B^\star)^{-1}\|}{\sqrt t}\left\|\sum_{i=k}^{t-1}\frac{1-\eta_i}{\eta_i}(\bbm_i-\bbm_{i-1})\right\|\1_{\{\tau_{k,r}>t\}}>\frac{\zeta}{8}\right) \\
\leq \frac{8\|(B^\star)^{-1}\|}{\zeta\sqrt t}\mE\left[\left\|\sum_{i=k}^{t-1}\frac{1-\eta_i}{\eta_i}(\bbm_i-\bbm_{i-1})\right\|\1_{\{\tau_{k,r}>t\}}\right]=o(1).
\end{multline*}

For the centered term, by Chebyshev's inequality and~\eqref{eq:effective-centered-average-negligible}, we have
\begin{align*}
&\PP\left(\left\|\frac{1}{\sqrt t}\sum_{i=k}^{t-1}\mathcal R^{-1}\left[(\tK_i-K_i)B_i^{-1}\bbm_i-\mathcal R(B^\star)^{-1}\{(\bg_i-\nabla f_i)-\nabla F(\bx^\star;\xi_i)\}\right]\right\|\1_{\{\tau_{k,r}>t\}}>\frac{\zeta}{8}\right) \\
&\leq \frac{64}{\zeta^2}\mE\left[\left\|\frac{1}{\sqrt t}\sum_{i=k}^{t-1}\mathcal R^{-1}\left[(\tK_i-K_i)B_i^{-1}\bbm_i-\mathcal R(B^\star)^{-1}\{(\bg_i-\nabla f_i)-\nabla F(\bx^\star;\xi_i)\}\right]\right\|^2\1_{\{\tau_{k,r}>t\}}\right]=o(1).
\end{align*}

For the noncentered term, by Markov's inequality and~\eqref{eq:effective-noncentered-average-negligible}, we have
\begin{align*}
&\PP\left(\left\|\frac{1}{\sqrt t}\sum_{i=k}^{t-1}\mathcal R^{-1}\left[\mathcal R(B^\star)^{-1}(\nabla f_i-B^\star\bDelta_i)+\left\{(I-K_i)B_i^{-1}-\mathcal R(B^\star)^{-1}\right\}\bbm_i\right]\right\|\1_{\{\tau_{k,r}>t\}}>\frac{\zeta}{8}\right) \\
&\leq \frac{8}{\zeta}\mE\left[\left\|\frac{1}{\sqrt t}\sum_{i=k}^{t-1}\mathcal R^{-1}\left[\mathcal R(B^\star)^{-1}(\nabla f_i-B^\star\bDelta_i)+\left\{(I-K_i)B_i^{-1}-\mathcal R(B^\star)^{-1}\right\}\bbm_i\right]\right\|\1_{\{\tau_{k,r}>t\}}\right]=o(1).
\end{align*}

Since all four probabilities converge to zero, the second probability in the stopping-time decomposition converges to zero. As a result, we obtain
\begin{equation*}
\limsup_{t\rightarrow\infty} \PP\left(\left\|\frac{1}{\sqrt t}\sum_{i=k}^{t-1}\bDelta_i+\frac{1}{\sqrt t}\sum_{i=k}^{t-1}(B^\star)^{-1}\nabla F(\bx^\star;\xi_i)\right\|>\frac{\zeta}{2}\right) \leq\PP(\tau_{k,r}<\infty)
<\frac{\epsilon}{2}.
\end{equation*}

Lastly, we extend both sums from $i=k$ to $i=0$. For the fixed $k$, we have
\begin{equation*}
\frac{1}{\sqrt t}\sum_{i=0}^{k-1}\left\{\bDelta_i+(B^\star)^{-1}\nabla F(\bx^\star;\xi_i)\right\}\stackrel{a.s.}{\rightarrow}\zero.
\end{equation*}
Therefore, by the triangle inequality and the union bound,
\begin{multline*}
\PP\left(\left\|\frac{1}{\sqrt t}\sum_{i=0}^{t-1}\bDelta_i+\frac{1}{\sqrt t}\sum_{i=0}^{t-1}(B^\star)^{-1}\nabla F(\bx^\star;\xi_i)\right\|>\zeta\right) \\
\leq \PP\left(\left\|\frac{1}{\sqrt t}\sum_{i=k}^{t-1}\bDelta_i+\frac{1}{\sqrt t}\sum_{i=k}^{t-1}(B^\star)^{-1}\nabla F(\bx^\star;\xi_i)\right\|>\frac{\zeta}{2}\right) \\
+\PP\left(\left\|\frac{1}{\sqrt t}\sum_{i=0}^{k-1}\left\{\bDelta_i+(B^\star)^{-1}\nabla F(\bx^\star;\xi_i)\right\}\right\|>\frac{\zeta}{2}\right).
\end{multline*}
Since the second probability on the right-hand side converges to zero by the preceding almost-sure convergence, then it follows that
\begin{equation*}
\limsup_{t\rightarrow\infty}
\PP\left(\left\|\frac{1}{\sqrt t}\sum_{i=0}^{t-1}\bDelta_i+\frac{1}{\sqrt t}\sum_{i=0}^{t-1}(B^\star)^{-1}\nabla F(\bx^\star;\xi_i)\right\|>\zeta\right)\leq\PP(\tau_{k,r}<\infty)
<\frac{\epsilon}{2}.
\end{equation*}

Since $\epsilon>0$ is arbitrary, we conclude that
\begin{equation*}
\sqrt{t} (\bar{\bx}_t - \bx^{\star}) = \frac{1}{\sqrt t}\sum_{i=0}^{t-1}(\bx_i-\bx^\star) = -\frac{1}{\sqrt t}\sum_{i=0}^{t-1}(B^\star)^{-1}\nabla F(\bx^\star;\xi_i) + o_p(1),
\end{equation*}
which proves~\eqref{eq:averaged-effective-noise} and completes the proof.

\subsection{Proof of Theorem~\ref{thm:last-normality}}\label{app:last-normality}

By Lemma~\ref{lem:operator-stabilization}, $\mathcal R$ is symmetric positive definite. By~\eqref{eq:last-effective-noise} and Slutsky's theorem, it suffices to establish the asymptotic normality of its leading stochastic term. For every $0\leq i\leq t-1$, define
\begin{equation}\label{eq:last-leading-array}
W_{t,i}\coloneqq-\frac{\varphi_i}{\sqrt{\varphi_t}}\Pi_{t,i+1}\mathcal R(B^\star)^{-1}\nabla F(\bx^\star;\xi_i).
\end{equation}

\noindent $\bullet$ \textbf{Verification of~\eqref{hh1}.}
By the unbiasedness in Assumption~\ref{ass:2} and $\nabla f(\bx^\star)=\zero$, we have
\begin{equation*}
\mE[\nabla F(\bx^\star;\xi_i)\mid\mF_{i-1}]=\nabla f(\bx^\star)=\zero.
\end{equation*}
Since $\Pi_{t,i+1}$ is deterministic, it follows that
\begin{equation*}
\mE[W_{t,i}\mid\mF_{i-1}]=-\frac{\varphi_i}{\sqrt{\varphi_t}}\Pi_{t,i+1}\mathcal R(B^\star)^{-1}\mE[\nabla F(\bx^\star;\xi_i)\mid\mF_{i-1}]=\zero.
\end{equation*}

\noindent $\bullet$ \textbf{Verification of~\eqref{hh2}.}
By the definitions of $\Omega$ and $\Gamma_{\bbm,\mathrm{as}}$, we have
\begin{equation*}
\mE\left[\mathcal R(B^\star)^{-1}\nabla F(\bx^\star;\xi_i)\nabla F(\bx^\star;\xi_i)^\top(B^\star)^{-1}\mathcal R^\top\mid\mF_{i-1}\right]
=\mathcal R\Omega\mathcal R^\top=\Gamma_{\bbm,\mathrm{as}}.
\end{equation*}
Therefore, the conditional covariance sum of the triangular array is deterministic and satisfies
\begin{equation}\label{eq:last-array-covariance}
\mathcal V_t \coloneqq \sum_{i=0}^{t-1}\mE[W_{t,i}W_{t,i}^\top\mid\mF_{i-1}] = \frac{1}{\varphi_t}\sum_{i=0}^{t-1}\Pi_{t,i+1}\varphi_i^2\Gamma_{\bbm,\mathrm{as}}\Pi_{t,i+1}^\top.
\end{equation}
Since $\Pi_{t+1,i+1}=(I-\varphi_t\mathcal R)\Pi_{t,i+1}$ for every $i\leq t-1$, we obtain the following recursion:
\begin{equation}\label{eq:last-array-covariance-recursion}
\mathcal V_{t+1}=\frac{\varphi_t}{\varphi_{t+1}}(I-\varphi_t\mathcal R)\mathcal V_t(I-\varphi_t\mathcal R)^\top+\frac{\varphi_t^2}{\varphi_{t+1}}\Gamma_{\bbm,\mathrm{as}}.
\end{equation}

If $\varphi<1$, the mean-value theorem gives $\varphi_t/\varphi_{t+1}-1=O((t+1)^{-1})=o(\varphi_t)$. If $\varphi=1$, we have $\varphi_t/\varphi_{t+1}=1+\varphi_t/C_\varphi$. Hence, in both cases, the following relations hold:
\begin{equation}\label{eq:last-stepsize-ratio}
\frac{\varphi_t}{\varphi_{t+1}}=1+2\zeta_\varphi\varphi_t+o(\varphi_t),
\qquad
\frac{\varphi_t^2}{\varphi_{t+1}}=\varphi_t+o(\varphi_t).
\end{equation}
Moreover, the condition on $C_\varphi$ when $\varphi=1$ ensures that $\lambda_m-\zeta_\varphi>0$. For all sufficiently large $t$, we also have $\|I-\varphi_t\mathcal R\|=1-\lambda_m\varphi_t$. Consequently,
\begin{equation*}
\frac{\varphi_t}{\varphi_{t+1}}\|I-\varphi_t\mathcal R\|^2
=1-2(\lambda_m-\zeta_\varphi)\varphi_t+o(\varphi_t).
\end{equation*}
Therefore, there exists a sufficiently large fixed integer $T$ such that, for every $t\geq T$,
\begin{equation*}
\frac{\varphi_t}{\varphi_{t+1}}\|I-\varphi_t\mathcal R\|^2
\leq1-(\lambda_m-\zeta_\varphi)\varphi_t,
\qquad
\frac{\varphi_t^2}{\varphi_{t+1}}\leq2\varphi_t,
\qquad
0<(\lambda_m-\zeta_\varphi)\varphi_t\leq1.
\end{equation*}
Then, we take norms in~\eqref{eq:last-array-covariance-recursion} and obtain
\begin{multline*}
\|\mathcal V_{t+1}\| \leq\frac{\varphi_t}{\varphi_{t+1}}\|I-\varphi_t\mathcal R\|^2\|\mathcal V_t\|+\frac{\varphi_t^2}{\varphi_{t+1}}\|\Gamma_{\bbm,\mathrm{as}}\|\leq\left\{1-(\lambda_m-\zeta_\varphi)\varphi_t\right\}\|\mathcal V_t\|
+2\|\Gamma_{\bbm,\mathrm{as}}\|\varphi_t \\
\leq(1-c\varphi_t)\|\mathcal V_t\|+C\varphi_t,
\end{multline*}
where $c\coloneqq\lambda_m-\zeta_\varphi>0$ and $C\coloneqq2(\|\Gamma_{\bbm,\mathrm{as}}\|\vee1)>0$. Now, we define
\begin{equation*}
M\coloneqq\max\left\{\max_{1\leq s\leq T}\|\mathcal V_s\|,\frac{C}{c}\right\}.
\end{equation*}
Suppose that $\|\mathcal V_t\|\leq M$ for some $t\geq T$. Then, we have
\begin{align*}
\|\mathcal V_{t+1}\|\leq(1-c\varphi_t)M+C\varphi_t=M-c\varphi_t\left(M-\frac{C}{c}\right) \leq M.
\end{align*}
Since $\|\mathcal V_T\|\leq M$, we can conclude that $\|\mathcal V_t\|\leq M$ for every $t\geq T$ by induction. Together with the finitely many indices preceding $T$, this proves that $\{\mathcal V_t\}_{t\geq1}$ is uniformly bounded.

Expanding~\eqref{eq:last-array-covariance-recursion} and using~\eqref{eq:last-stepsize-ratio} together with the uniform boundedness of $\mathcal V_t$, we further obtain
\begin{equation}\label{eq:last-array-covariance-expansion}
\mathcal V_{t+1}=\mathcal V_t-\varphi_t\left[\{\mathcal R-\zeta_\varphi I\}\mathcal V_t+\mathcal V_t\{\mathcal R-\zeta_\varphi I\}-\Gamma_{\bbm,\mathrm{as}}\right]+o(\varphi_t).
\end{equation}
Let $\mathcal R_\varphi\coloneqq\mathcal R-\zeta_\varphi I$ and $\mathcal Q_\varphi\coloneqq I\otimes\mathcal R_\varphi+\mathcal R_\varphi\otimes I$. Since $\mathcal R_\varphi$ is symmetric positive definite, $\mathcal Q_\varphi$ is symmetric positive definite with $\lambda_{\min}(\mathcal Q_\varphi)=2(\lambda_m-\zeta_\varphi)>0$.
Define
\begin{equation*}
s_t\coloneqq\operatorname{vec}(\mathcal V_t),
\qquad
s^\star\coloneqq\operatorname{vec}(\Sigma_{\bbm,\mathrm{as}}^{l})=\mathcal Q_\varphi^{-1}\operatorname{vec}(\Gamma_{\bbm,\mathrm{as}}).
\end{equation*}
By vectorizing the recursion~\eqref{eq:last-array-covariance-expansion}, we obtain
\begin{equation*}
s_{t+1}-s^\star=(I-\varphi_t\mathcal Q_\varphi)(s_t-s^\star)+\varphi_t\varepsilon_t,
\end{equation*}
where $\varepsilon_t\rightarrow\zero$. Therefore, for all sufficiently large $t$,
\begin{equation*}
\|s_{t+1}-s^\star\|
\leq\left\{1-2(\lambda_m-\zeta_\varphi)\varphi_t\right\}\|s_t-s^\star\|+\varphi_t\|\varepsilon_t\|.
\end{equation*}
Applying Lemma~\ref{lem:polynomial_product_rate} to the preceding recursion, for every $\varphi\in(0.5,1]$, we obtain
\begin{equation*}
\limsup_{t\rightarrow\infty}\|s_t-s^\star\| \leq \frac{1}{2(\lambda_m-\zeta_\varphi)} \limsup_{t\rightarrow\infty}\|\varepsilon_t\| = 0.
\end{equation*}
Therefore, $s_t \rightarrow s^\star$, which is equivalent to $\mathcal V_t\rightarrow\Sigma_{\bbm,\mathrm{as}}^{l}$.

Moreover, since $s^\star=\operatorname{vec}(\Sigma_{\bbm,\mathrm{as}}^{l})$, the equation $\mathcal Q_\varphi s^\star=\operatorname{vec}(\Gamma_{\bbm,\mathrm{as}})$ is equivalent to
\begin{equation*}
(\mathcal R-\zeta_\varphi I)\Sigma_{\bbm,\mathrm{as}}^{l} + \Sigma_{\bbm,\mathrm{as}}^{l}(\mathcal R-\zeta_\varphi I) = \Gamma_{\bbm,\mathrm{as}},
\end{equation*}
which is precisely~\eqref{eq:last-normality}. Since $\mathcal Q_\varphi$ is positive definite, the equation $\mathcal Q_\varphi s=\operatorname{vec}(\Gamma_{\bbm,\mathrm{as}})$ admits a unique solution, and hence so does~\eqref{eq:last-normality}. Finally, each $\mathcal V_t$ is positive semidefinite because it is a sum of conditional covariance matrices. Therefore, the convergence of $\mathcal{V}_t$ implies that $\Sigma_{\bbm,\mathrm{as}}^{l}$ is also positive semidefinite, and condition~\eqref{hh2} in Lemma~\ref{lemma:hall} is satisfied.

\noindent $\bullet$ \textbf{Verification of~\eqref{hh3}.}
Let $\delta\coloneqq q_g-2>0$. {\red The argument before~\eqref{eq:effective-gradient-remainders} establishes a $q_g$-moment bound for $\nabla F(\bx^\star;\xi)$. Since $q_g=2+\delta$, we have
\begin{equation}\label{eq:last-gradient-moment}
\mE[\|\nabla F(\bx^\star;\xi)\|^{2+\delta}]\leq C_{g,2}<\infty.
\end{equation}}

Define $A_t\coloneqq\sum_{i=0}^{t-1}\varphi_i^{2+\delta}\|\Pi_{t,i+1}\|^{2+\delta}$. Then, for all sufficiently large $t$, we have
\begin{equation*}
A_{t+1}\leq(1-\lambda_m\varphi_t)^{2+\delta}A_t+\varphi_t^{2+\delta}=(1-\lambda_m\varphi_t)^{2+\delta}A_t+\varphi_t^{2+\delta/2}\varphi_t^{\delta/2}.
\end{equation*}
By Lemma~\ref{lem:polynomial_product_rate}, for $\varphi\in(0.5,1]$, we obtain
\begin{align}\label{eq:last-lindeberg-kernel}
\limsup_{t\rightarrow\infty}\frac{A_t}{\varphi_t^{1+\delta/2}} &\leq
\begin{cases}
\displaystyle
\frac{1}{(2+\delta)\lambda_m}\limsup_{t\rightarrow\infty}\varphi_t^{\delta/2},
& \varphi\in(0.5,1),\\[8pt]
\displaystyle
\frac{C_\varphi}{(2+\delta)C_\varphi\lambda_m-(1+\delta/2)}
\limsup_{t\rightarrow\infty}\varphi_t^{\delta/2},
& \varphi=1,
\end{cases}
=0.
\end{align}
Here, the denominator in the second case is positive because
\begin{equation*}
1+\frac{\delta}{2}=\frac{2+\delta}{2}<(2+\delta)C_\varphi\lambda_m,
\end{equation*}
where the inequality follows from $C_\varphi\lambda_m>1/2$. Consequently, we obtain $A_t=o(\varphi_t^{1+\delta/2})$.

For every $\epsilon>0$, by the inequality $a^2\1_{\{a>\epsilon\}}\leq a^{2+\delta}/\epsilon^\delta$,~\eqref{eq:last-gradient-moment} and~\eqref{eq:last-lindeberg-kernel}, we obtain
\begin{multline}\label{eq:last-conditional-lindeberg}
\sum_{i=0}^{t-1}\mE[\|W_{t,i}\|^2\1_{\{\|W_{t,i}\|>\epsilon\}}\mid\mF_{i-1}]\leq\frac{1}{\epsilon^\delta}\sum_{i=0}^{t-1}\mE[\|W_{t,i}\|^{2+\delta}\mid\mF_{i-1}] \\
\lesssim\frac{1}{\epsilon^\delta\varphi_t^{1+\delta/2}}\sum_{i=0}^{t-1}\varphi_i^{2+\delta}\|\Pi_{t,i+1}\|^{2+\delta}=\frac{A_t}{\epsilon^\delta\varphi_t^{1+\delta/2}}\rightarrow0.
\end{multline}
Therefore, condition~\eqref{hh3} is satisfied.

Combining~\eqref{hh1}--\eqref{hh3}, Lemma~\ref{lemma:hall} yields
\begin{equation}\label{eq:last-leading-normality}
\sum_{i=0}^{t-1}W_{t,i} = -\frac{1}{\sqrt{\varphi_t}}\sum_{i=0}^{t-1}\Pi_{t,i+1}\varphi_i(I-K^\star)(B^\star)^{-1}\nabla F(\bx^\star;\xi_i) \xrightarrow{d} \mN(\zero,\Sigma_{\bbm,\mathrm{as}}^{l}).
\end{equation}
Finally, combining~\eqref{eq:last-effective-noise} and~\eqref{eq:last-leading-normality} and applying Slutsky's theorem, we obtain
\begin{equation*}
\frac{1}{\sqrt{\varphi_t}}(\bx_t-\bx^\star)\xrightarrow{d}\mN(\zero,\Sigma_{\bbm,\mathrm{as}}^{l}),
\end{equation*}
And we complete the proof.

\subsection{Proof of Proposition~\ref{prop:covariance-comparison}}\label{app:covariance-comparison}

Recall that $A_\varphi^\star=(I-K^\star)-\zeta_\varphi I$. By Lemma~\ref{lem:operator-stabilization}, $K^\star$ is symmetric and $I-K^\star\succ\zero$. If $\varphi<1$, then $\zeta_\varphi=0$. If $\varphi=1$, the condition on $C_\varphi$ in Theorem~\ref{thm:sketch-noise-reduction} gives $\zeta_\varphi=1/(2C_\varphi)<\lambda_{\min}(I-K^\star)$. Therefore, $A_\varphi^\star\succ\zero$ under both stepsize regimes.

\noindent $\bullet$ \textbf{Proof of part~\textbf{(a)}.}
Since $K^\star=\mE[\tK^\star]$, we have $\mE[\tK^\star-K^\star]=\zero$. Also, we have
\begin{align*}
\Gamma_{\bg,\mathrm{as}} &=\Gamma_{\bbm,\mathrm{as}}-(I-K^\star)\Omega\mE[(\tK^\star-K^\star)^\top] \\
&\quad -\mE[\tK^\star-K^\star]\Omega(I-K^\star)^\top+\mE\left[(\tK^\star-K^\star)\Omega(\tK^\star-K^\star)^\top\right] \\
&=\Gamma_{\bbm,\mathrm{as}}+\mE\left[(\tK^\star-K^\star)\Omega(\tK^\star-K^\star)^\top\right].
\end{align*}
Since $\Omega\succeq\zero$, this identity implies
$\Gamma_{\bbm,\mathrm{as}}\preceq\Gamma_{\bg,\mathrm{as}}$.

Subtracting~\eqref{eq:last-normality} from the defining Lyapunov equation for $\Sigma_{\bg,\mathrm{as}}^{l}$, we obtain
\begin{multline*}
\left(\mathcal{R}-\zeta_\varphi I\right)(\Sigma_{\bg,\mathrm{as}}^{l}-\Sigma_{\bbm,\mathrm{as}}^{l})+(\Sigma_{\bg,\mathrm{as}}^{l}-\Sigma_{\bbm,\mathrm{as}}^{l})\left(\mathcal{R}-\zeta_\varphi I\right) \\
=\mE\left[(\tK^\star-K^\star)\Omega(\tK^\star-K^\star)^\top\right]\succeq\zero.
\end{multline*}
Since $\mathcal{R}-\zeta_\varphi I\succ\zero$, the unique solution of this Lyapunov equation admits the representation
\begin{multline*}
\Sigma_{\bg,\mathrm{as}}^{l}-\Sigma_{\bbm,\mathrm{as}}^{l} \\
=\int_0^\infty\exp\left\{-\left(\mathcal{R}-\zeta_\varphi I\right)s\right\}\mE\left[(\tK^\star-K^\star)\Omega(\tK^\star-K^\star)^\top\right]\exp\left\{-\left(\mathcal{R}-\zeta_\varphi I\right)s\right\}\,ds\succeq\zero,
\end{multline*}
which completes the proof of part~\textbf{(a)}.

\noindent $\bullet$ \textbf{Proof of part~\textbf{(b)}.}
Suppose that the Newton system is solved exactly. Then, we have $K^\star=\tK^\star=\zero$ and $\Gamma_{\bbm,\mathrm{as}}=\Gamma_{\bg,\mathrm{as}}=\Omega$. In this regime, the Lyapunov equations in~\eqref{eq:last-normality} and~\eqref{eq:single-last-normality} reduce to
\begin{equation*}
2(1-\zeta_\varphi)\Sigma_{\bbm,\mathrm{e}}^{l}
=2(1-\zeta_\varphi)\Sigma_{\bg,\mathrm{e}}^{l}
=\Omega.
\end{equation*}
If $\varphi<1$, then $\zeta_\varphi=0$, which gives $\Sigma_{\bbm,\mathrm{e}}^{l}=\Sigma_{\bg,\mathrm{e}}^{l}=0.5\Omega$. If $\varphi=1$, then $\zeta_\varphi=1/(2C_\varphi)$ and hence
\begin{equation*}
\Sigma_{\bbm,\mathrm{e}}^{l}
=\Sigma_{\bg,\mathrm{e}}^{l}
=\frac{1}{2\{1-1/(2C_\varphi)\}}\Omega
=\frac{C_\varphi}{2C_\varphi-1}\Omega.
\end{equation*}
This completes the proof of part~\textbf{(b)}.

\noindent $\bullet$ \textbf{Proof of part~\textbf{(c)}.}
When $\varphi=C_\varphi=1$, part~\textbf{(b)} gives $\Sigma_{\bbm,\mathrm{e}}^{l}=\Sigma_{\bg,\mathrm{e}}^{l}=\Omega$, and $t\varphi_t\rightarrow1$. Therefore, by Theorems~\ref{thm:last-normality} and~\ref{thm:single-normality} and Slutsky's theorem, exact online Newton, with or without gradient averaging, satisfies
\begin{equation*}
\sqrt{t}\,(\bx_t-\bx^\star)\xrightarrow{d}\mN(\zero,\Omega).
\end{equation*}

We next establish the covariance ordering for sketched Newton with the same stepsizes. In this case, $\zeta_\varphi=1/2$, and Lemma~\ref{lem:operator-stabilization} gives
\begin{equation*}
\lambda_m=\lambda_{\min}(I-K^\star)\geq\frac{3}{4}>\frac{1}{2},
\qquad
\frac{1}{2}I-K^\star\succ\zero.
\end{equation*}
Moreover, by the definition of $\Gamma_{\bbm,\mathrm{as}}$, we have
\begin{multline*}
\Gamma_{\bbm,\mathrm{as}}
-\left\{\left(\frac{1}{2}I-K^\star\right)\Omega
+\Omega\left(\frac{1}{2}I-(K^\star)^\top\right)\right\}\\
=(I-K^\star)\Omega(I-K^\star)^\top
-\left\{\left(\frac{1}{2}I-K^\star\right)\Omega
+\Omega\left(\frac{1}{2}I-(K^\star)^\top\right)\right\}
=K^\star\Omega(K^\star)^\top.
\end{multline*}
Subtracting the expression inside the braces from~\eqref{eq:last-normality} gives
\begin{equation*}
\left(\frac{1}{2}I-K^\star\right)
(\Sigma_{\bbm,\mathrm{as}}^{l}-\Omega)
+(\Sigma_{\bbm,\mathrm{as}}^{l}-\Omega)
\left(\frac{1}{2}I-(K^\star)^\top\right)
=K^\star\Omega(K^\star)^\top\succeq\zero.
\end{equation*}
Since $\frac{1}{2}I-K^\star\succ\zero$, it follows that
\begin{multline*}
\Sigma_{\bbm,\mathrm{as}}^{l}-\Omega\\
=\int_0^\infty
\exp\left\{-\left(\frac{1}{2}I-K^\star\right)s\right\}
K^\star\Omega(K^\star)^\top
\exp\left\{-\left(\frac{1}{2}I-(K^\star)^\top\right)s\right\}
\,ds\succeq\zero.
\end{multline*}
Therefore, $\Omega\preceq\Sigma_{\bbm,\mathrm{as}}^{l}$. Combining this inequality with parts~\textbf{(a)} and~\textbf{(b)} gives
\begin{equation*}
\Omega
=\Sigma_{\bg,\mathrm{e}}^{l}
=\Sigma_{\bbm,\mathrm{e}}^{l}
\preceq\Sigma_{\bbm,\mathrm{as}}^{l}
\preceq\Sigma_{\bg,\mathrm{as}}^{l},
\end{equation*}
which completes the proof of part~\textbf{(c)}.

\subsection{Proof of Theorem~\ref{thm:averaged-normality}}\label{app:averaged-normality}

Since $\nabla f(\bx^\star)=\zero$, we have
\begin{align*}
&\mE[(B^\star)^{-1}\nabla F(\bx^\star;\xi_i)]=(B^\star)^{-1}\nabla f(\bx^\star)=\zero, \\
&\mE\left[(B^\star)^{-1}\nabla F(\bx^\star;\xi_i)\nabla F(\bx^\star;\xi_i)^\top(B^\star)^{-1}\right]=\Omega.
\end{align*}

Moreover, by~\eqref{eq:last-gradient-moment} and Jensen's inequality, we further obtain
\begin{multline*}
\mE[\|(B^\star)^{-1}\nabla F(\bx^\star;\xi_i)\|^2] \leq \frac{1}{\gamma_H^2}\mE[\|\nabla F(\bx^\star;\xi_i)\|^2] \\
\leq\frac{1}{\gamma_H^2}\left\{\mE[\|\nabla F(\bx^\star;\xi_i)\|^{2+\delta}]\right\}^{2/(2+\delta)}<\infty.
\end{multline*}
Therefore, the central limit theorem for i.i.d. random variables gives us
\begin{equation*}
-\frac{1}{\sqrt t}\sum_{i=0}^{t-1}(B^\star)^{-1}\nabla F(\bx^\star;\xi_i)\xrightarrow{d}\mN(\zero,\Omega).
\end{equation*}

Finally, by~\eqref{eq:averaged-effective-noise}, we have
\begin{equation*}
\sqrt{t}\,(\bar{\bx}_t-\bx^\star)=-\frac{1}{\sqrt t}\sum_{i=0}^{t-1}(B^\star)^{-1}\nabla F(\bx^\star;\xi_i)+o_p(1).
\end{equation*}
Combining the preceding two equations and applying Slutsky's theorem, we conclude that
\begin{equation*}
\sqrt{t}\,(\bar{\bx}_t-\bx^\star)\xrightarrow{d}\mN(\zero,\Sigma_{\bbm,\mathrm{as}}^{a}),
\end{equation*}
where $\Sigma_{\bbm,\mathrm{as}}^{a}=\Omega$. This proves~\eqref{eq:averaged-normality} and completes the proof.

\section{Experimental Details and Additional Results}\label{app:experiments}

We provide the experimental details and additional results for Section~\ref{sec:experiments}. The comparisons below complement the main paper with higher-dimensional problems, full-matrix covariance errors, method-specific normal QQ plots, and estimation-error trajectories.

\subsection{Detailed experimental setup}\label{app:experiment-design}

\noindent$\bullet$ \textbf{Regression models.}
For both models, we set $\bx^\star=(0,1/(d-1),\ldots,1)^\top$ and generate $\ba\sim\mN(\zero,\Sigma_a)$. We consider $d\in\{20,40\}$ and three covariate covariances: $\Sigma_a=I$, the equicorrelation matrix $(1-\rho)I+\rho\mathbf1\mathbf1^\top$, and the Toeplitz matrix with entries $(\Sigma_a)_{jk}=\rho^{|j-k|}$, where $\rho=0.4$.
For linear regression, $y=\ba^\top\bx^\star+\epsilon$ with independent $\epsilon\sim\mN(0,1)$, and
\begin{equation*}
F(\bx;(\ba,y))=\tfrac12(\ba^\top\bx-y)^2.
\end{equation*}
For logistic regression, $y\mid\ba\sim\operatorname{Bernoulli}\{\sigma(\ba^\top\bx^\star)\}$, where $\sigma(z)=(1+e^{-z})^{-1}$, and
\begin{equation*}
F(\bx;(\ba,y))=\log\{1+\exp(\ba^\top\bx)\}-y\ba^\top\bx
+\tfrac{c}{2}\|\bx-\bx^\star\|^2,\qquad c=2.
\end{equation*}
The penalty is centered at $\bx^\star$ so that the population minimizer remains unchanged. We use this penalty only in simulations, since $\bx^\star$ is unknown in practice. In both models, observations are drawn from a fixed distribution, and the gradient and Hessian are evaluated at the current iterate using the same observation. The Gaussian covariates are unbounded, and the linear sample Hessian is rank one. Thus, the uniform sampled Hessian bounds imposed in the existing sketched Newton literature do not hold in these experiments, while the population-level conditions in Assumption~\ref{ass:1} do.

\noindent$\bullet$ \textbf{Methods and sketching parameters.}
We compare online Newton with and without gradient averaging, using either accelerated sketching or exact solves, and include SGD. All methods start from $\bx_0=\mathbf1$. Newton methods use the Hessian average $B_t=\{I+\sum_{i=0}^{t-1}H_i\}/(t+1)$, which is fixed before $\xi_t$ is drawn. The term $I/(t+1)$ keeps $B_t$ nonsingular and vanishes as $t$ increases. We use Kaczmarz sketches $S=\be_j$ with $j$ uniform on $\{1,\ldots,d\}$, or Gaussian sketches $S\sim\mN(\zero,I)$, with $\tau\in\{5,10\}$. Each inner solve starts from $\bz_{t,0}=\bv_{t,0}=\zero$. We compute the acceleration parameters from the current $B_t$ at $t=0,500,1000,\ldots$ and keep them fixed between updates. The expectations in~\eqref{eq:mu-nu} are computed exactly by averaging over all $d$ coordinates for Kaczmarz sketches. For Gaussian sketches, we use the same $M_{\mathrm{acc}}=16{,}384$ Gaussian vectors at each update, generated independently of the training data and inner sketches.

\noindent$\bullet$ \textbf{Stepsizes.}
We set $\varphi=0.8$ for all methods and use $\eta_t=(t+1)^{-0.6}$ with $\bbm_0=\bg_0$ for gradient averaging. For logistic regression, the Newton stepsize is $\varphi_t=(t+1)^{-0.8}$. For linear regression, all four Newton methods use the same initial stepsizes, obtained by interpolating $\log\varphi_t$ linearly in $\log(t+1)$ through
\begin{equation*}
(t,\varphi_t)\in\big\{(0,0.02),(39,0.0005),(99,0.01),(199,0.014427),(499,0.020794)\big\},
\end{equation*}
where the final two stepsizes are rounded. For $t\geq500$, we use $\varphi_t=3/(t+1)^{0.8}$. Linear SGD uses $\min\{(t+1)^{-0.8},0.02\}$ for the first 500 updates and $(t+1)^{-0.8}$ thereafter, while logistic SGD uses $0.4/(t+1)^{0.8}$. These choices are fixed across runs. Within each problem setting, the Newton methods use the same stepsizes with and without gradient averaging, for both sketching distributions and both values of $\tau$.

\noindent$\bullet$ \textbf{Runs and paired comparisons.}
The two models, two dimensions, and three covariate covariance matrices give 12 problem settings. For each setting, we compare eight accelerated-sketching configurations, two exact-Newton configurations, and SGD, giving 132 configurations in total. For each configuration, we conduct $R=200$ independent runs of $T=10^5$ iterations. Within each setting, all methods use the same observations in each run. Sketched Newton with and without gradient averaging also uses the same sketches for the same sketching distribution and $\tau$. The observations and sketches are generated independently.

\subsection{Covariance estimation and statistical comparisons}\label{app:experiment-estimands}

We use the averaged iterate $\bar{\bx}_t=t^{-1}\sum_{i=0}^{t-1}\bx_i$. At iteration $t$, we compute centered sample covariances across the $R$ runs of
\begin{equation*}
Z_{r,t}^{l}=\frac{\bx_t^{(r)}-\bx^\star}{\sqrt{\varphi_t}},
\qquad
Z_{r,t}^{a}=\sqrt{t}\,(\bar{\bx}_t^{(r)}-\bx^\star),
\end{equation*}
with denominator $R-1$. For the last iterate, normalization uses $C/(t+1)^{0.8}$ with the corresponding constant $C$ for each method, rather than the modified initial stepsizes. We measure directional variance along $\mathbf1$. To compare estimation accuracy, we also report the empirical bias norm, MSE, and mean population objective gap using unnormalized errors. The bias norm is the norm of the mean error across runs.

For the last iterate, we compare each empirical covariance with the corresponding theoretical covariance, which generally differs from $\Omega$. For the averaged iterate, online sketched Newton with gradient averaging and exact online Newton have limiting covariance $\Omega$, whereas vanilla sketched Newton has an additional covariance term. We therefore compare averaged-iterate covariances both with their respective theoretical values and with $\Omega$. We use covariance traces to compare total variance and Frobenius errors to measure differences between covariance matrices.

We construct pointwise 95\% confidence intervals using 1,000 bootstrap resamples of the independent runs. Within each problem setting, we use the same resampled run indices for all methods and iterations. These intervals measure uncertainty across runs, with the theoretical covariances held fixed. For the ratio tables, we compute each ratio within a matched setting and then summarize these point estimates across settings. For the QQ plots, we use theoretical rather than empirical standard deviations and do not subtract the empirical means.

We compute population moments analytically for linear regression and by numerical Gaussian integration for logistic regression. The acceleration parameters used in the theoretical covariances are computed at $B^\star$ using the same procedure as in the experiments. We estimate the sketch-dependent covariance terms from $M_K=131{,}072$ independent realizations of the inner solver at $B^\star$, using sketches generated independently of the training and parameter-estimation samples. For these Monte Carlo estimates, we require both the relative Frobenius standard error and the relative standard error of the variance along $\mathbf1$ to be below 2\%. The theoretical covariances are computed using the known data distributions and the true parameter $\bx^\star$.

\subsection{Additional covariance comparisons}\label{app:experiment-grid}

\begin{figure}[t]
\centering
\captionsetup{font=footnotesize,skip=5pt}
\captionsetup[subfigure]{font=footnotesize,skip=1pt}
\begin{subfigure}[t]{0.247\linewidth}
\centering
\includegraphics[width=\linewidth]{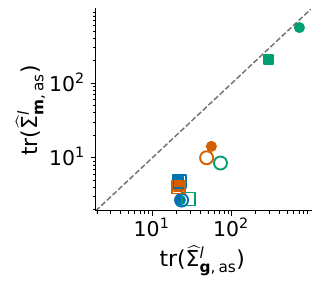}
\caption{Linear: last}
\end{subfigure}\hfill
\begin{subfigure}[t]{0.247\linewidth}
\centering
\includegraphics[width=\linewidth]{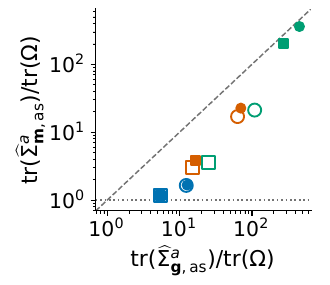}
\caption{Linear: averaged}
\end{subfigure}\hfill
\begin{subfigure}[t]{0.247\linewidth}
\centering
\includegraphics[width=\linewidth]{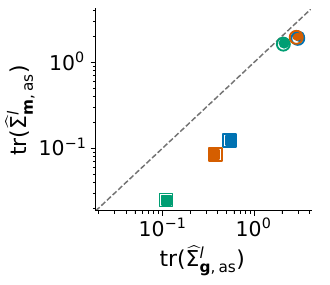}
\caption{Logistic: last}
\end{subfigure}\hfill
\begin{subfigure}[t]{0.247\linewidth}
\centering
\includegraphics[width=\linewidth]{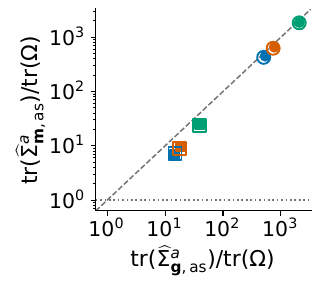}
\caption{Logistic: averaged}
\end{subfigure}
\par\vspace{1pt}
\includegraphics[width=0.94\linewidth]{figures/experiments/covariance_legend.pdf}
\caption{\textit{Covariance comparisons for $d=40$, $T=10^5$, and $R=200$. Horizontal and vertical axes show traces without and with gradient averaging, respectively; averaged-iterate traces are divided by $\operatorname{tr}(\Omega)$. Each panel contains 12 settings. Colors identify covariate covariances, circles/squares distinguish $\tau=5/10$, and filled/open markers denote Kaczmarz/Gaussian sketches. Diagonals indicate equality; dotted horizontal lines indicate one. Open markers are enlarged to distinguish overlaps without changing their positions.}}
\label{fig:covariance-d40}
\end{figure}

From Figure~\ref{fig:covariance-d40}, we observe smaller covariance traces with gradient averaging for both the last and averaged iterates in all 24 matched settings with $d=40$. As in the $d=20$ results in Figure~\ref{fig:covariance-comparison}(a--d), this reduction holds for both models, both sketching distributions, and all three covariate covariance matrices.

Table~\ref{tab:experiment-ratios} reports additional comparisons of directional variance, bias, and objective gap, together with the ranges of the ratios across settings. We observe smaller values with gradient averaging for both the last and averaged iterates in all 48 sketching settings. Therefore, gradient averaging improves estimation accuracy as well as reducing covariance. In contrast, the covariance-trace ratios for exact online Newton remain close to one, consistent with Proposition~\ref{prop:covariance-comparison}.

\begin{table}[t]
\centering
\captionsetup{font=footnotesize,skip=5pt}
\caption{\textit{Moving-average/vanilla ratios at $T=10^5$ over $R=200$ runs per configuration. Entries give median [minimum, maximum] over 48 sketching settings or 12 exact-solve settings. All sketching ratios are below one. Directional variance uses $\mathbf1$; bias, MSE, and objective gap use unnormalized errors.}}
\label{tab:experiment-ratios}
\small
\begin{tabular}{llcc}
\toprule
Solver & Quantity & Last iterate & Averaged iterate\\
\midrule
Accelerated sketching & Covariance trace & 0.232 [0.098, 0.787] & 0.409 [0.130, 0.896]\\
& $\mathbf1$-direction variance & 0.322 [0.110, 0.814] & 0.427 [0.148, 0.906]\\
& Empirical bias norm & 0.525 [0.290, 0.985] & 0.818 [0.270, 0.986]\\
& MSE & 0.234 [0.099, 0.935] & 0.428 [0.192, 0.939]\\
& Population objective gap & 0.241 [0.116, 0.935] & 0.428 [0.220, 0.939]\\
\midrule
Exact solve & Covariance trace & 1.003 [0.981, 1.041] & 1.003 [1.001, 1.009]\\
\bottomrule
\end{tabular}
\end{table}

We also compare the averaged-iterate covariance matrices with $\Omega$ using the relative Frobenius error $E_u=\|\widehat\Sigma_{u,\mathrm{as}}^{a}-\Omega\|_F/\|\Omega\|_F$ for $u\in\{\bg,\bbm\}$, where $\|\cdot\|_F$ denotes the Frobenius norm. From Table~\ref{tab:experiment-groups}, we observe smaller errors with gradient averaging in all 48 matched settings. Thus, gradient averaging brings the covariance matrices closer to $\Omega$, not only their traces. Increasing $\tau$ further reduces these errors, although some settings still have large errors at $T=10^5$.

\begin{table}[t]
\centering
\captionsetup{font=footnotesize,skip=5pt}
\caption{\textit{Relative Frobenius errors of averaged-iterate covariances against $\Omega$ at $T=10^5$. Entries give median [minimum, maximum] over 12 dimension--covariate-covariance--sketch combinations. The final column aggregates settingwise error ratios, not ratios of the two median errors.}}
\label{tab:experiment-groups}
\footnotesize
\setlength{\tabcolsep}{4pt}
\begin{tabular}{lcccc}
\toprule
Model & $\tau$ & $E_{\bg}$ & $E_{\bbm}$ & $E_{\bbm}/E_{\bg}$\\
\midrule
Linear & 5 & 14.979 [4.125, 604.798] & 2.240 [0.343, 514.822] & 0.148 [0.069, 0.851]\\
Linear & 10 & 5.034 [1.824, 344.949] & 0.556 [0.312, 275.337] & 0.140 [0.090, 0.798]\\
Logistic & 5 & 289.971 [11.986, 2867.951] & 241.862 [5.051, 2590.490] & 0.717 [0.421, 0.903]\\
Logistic & 10 & 8.937 [2.150, 52.009] & 3.813 [0.464, 32.736] & 0.344 [0.203, 0.629]\\
\bottomrule
\end{tabular}
\end{table}

Table~\ref{tab:experiment-uncertainty} reports the final covariance-trace ratios and 95\% bootstrap intervals for Figure~\ref{fig:covariance-comparison}(e,f). We observe that gradient averaging brings the covariance traces closer to those of exact online Newton. The bootstrap intervals for sketched Newton with and without gradient averaging do not overlap.

\begin{table}[t]
\centering
\captionsetup{font=footnotesize,skip=5pt}
\caption{\textit{Averaged-iterate trace divided by $\operatorname{tr}(\Omega)$, with pointwise 95\% bootstrap intervals, for $d=20$, identity covariate covariance, Gaussian sketches, $\tau=10$, $T=10^5$, and $R=200$. Exact online Newton uses $\bg_t$.}}
\label{tab:experiment-uncertainty}
\small
\begin{tabular}{lcc}
\toprule
Method & Linear regression & Logistic regression\\
\midrule
Vanilla sketched Newton & 2.725 [2.604, 2.829] & 3.078 [2.925, 3.201]\\
Sketched Newton with gradient averaging & 1.047 [1.002, 1.085] & 1.248 [1.188, 1.292]\\
Exact online Newton & 1.012 [0.968, 1.046] & 1.071 [1.019, 1.112]\\
\bottomrule
\end{tabular}
\end{table}

\subsection{Normal QQ plots with method-specific covariances}\label{app:experiment-normality}

\begin{figure}[t]
\centering
\captionsetup{font=footnotesize,skip=5pt}
\captionsetup[subfigure]{font=footnotesize,skip=1pt}
\begin{subfigure}[t]{0.247\linewidth}
\centering
\includegraphics[width=\linewidth]{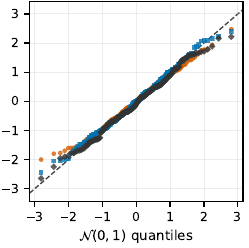}
\caption{Linear: last}
\end{subfigure}\hfill
\begin{subfigure}[t]{0.247\linewidth}
\centering
\includegraphics[width=\linewidth]{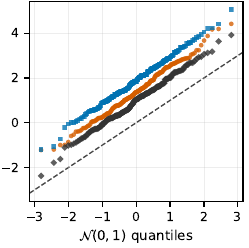}
\caption{Linear: averaged}
\end{subfigure}\hfill
\begin{subfigure}[t]{0.247\linewidth}
\centering
\includegraphics[width=\linewidth]{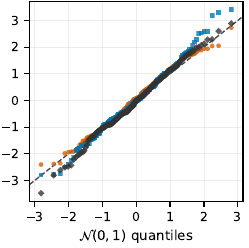}
\caption{Logistic: last}
\end{subfigure}\hfill
\begin{subfigure}[t]{0.247\linewidth}
\centering
\includegraphics[width=\linewidth]{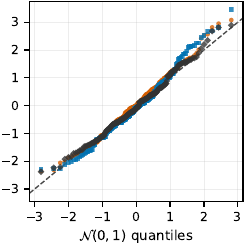}
\caption{Logistic: averaged}
\end{subfigure}
\par\vspace{1pt}
\includegraphics[width=0.88\linewidth]{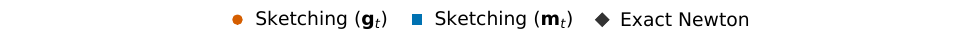}
\caption{\textit{Method-specific normal QQ plots for $d=20$, identity covariate covariance, Gaussian sketches, $\tau=10$, $T=10^5$, and $R=200$. Horizontal axes show standard normal quantiles; vertical axes show quantiles of $\mathbf1^\top Z_{r,T}^{q}/\sqrt{\mathbf1^\top\Sigma_{u,s}^{q}\mathbf1}$, using each method's own theoretical covariance. Sketching uses $\bg_t/\bbm_t$ without/with gradient averaging; exact Newton uses $\bg_t$. Dashed lines indicate equality. Empirical means are retained.}}
\label{fig:qq-method-specific}
\end{figure}

Figure~\ref{fig:qq-method-specific} shows normal QQ plots using each method's theoretical covariance. We consider $d=20$, identity covariate covariance, Gaussian sketches, and $\tau=10$, as in Figure~\ref{fig:covariance-comparison}(e--h). For each method and iterate, we project $Z_{r,T}^{q}$ onto $\mathbf1$ and divide by the theoretical standard deviation $\sqrt{\mathbf1^\top\Sigma_{u,s}^{q}\mathbf1}$. Unlike the averaged-iterate QQ plots in the main paper, vanilla sketched Newton is normalized by its own larger covariance rather than by $\Omega$.

From panels (a,c,d), we observe that the empirical quantiles closely follow the standard normal reference over most quantiles. Panel (b) shows a mean shift for the linear averaged iterate, which remains visible because we do not subtract the empirical means. These plots examine the normal approximation for each method using its corresponding limiting covariance.

\subsection{Estimation error and computational cost}\label{app:experiment-optimization}

\begin{figure}[t]
\centering
\captionsetup{font=footnotesize,skip=5pt}
\includegraphics[width=\linewidth]{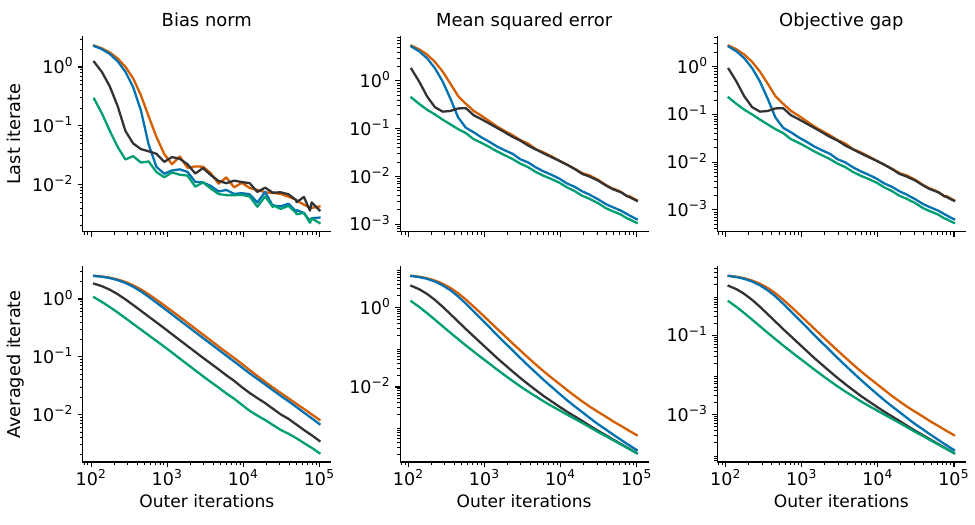}
\par\vspace{1pt}
\includegraphics[width=0.94\linewidth]{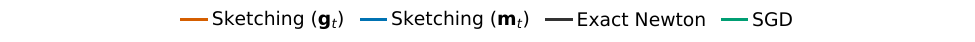}
\caption{\textit{Estimation errors for linear regression with $d=20$, identity covariate covariance, Gaussian sketches, $\tau=10$, and $R=200$. Columns show empirical bias norm, MSE, and population objective gap; rows show last and averaged iterates. Sketching uses $\bg_t/\bbm_t$ without/with gradient averaging; exact Newton uses $\bg_t$, and SGD is the first-order baseline. All quantities use unnormalized errors.}}
\label{fig:optimization-linear}
\end{figure}

\begin{figure}[t]
\centering
\captionsetup{font=footnotesize,skip=5pt}
\includegraphics[width=\linewidth]{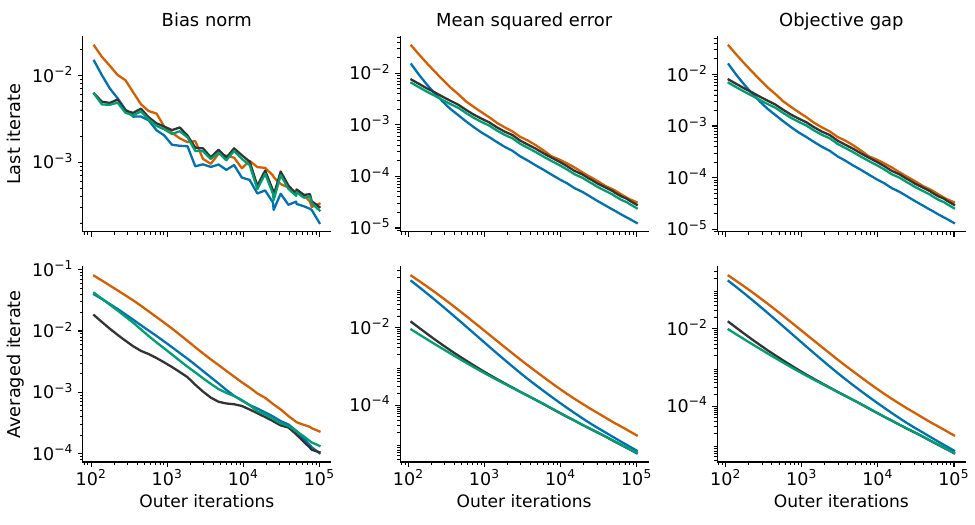}
\par\vspace{1pt}
\includegraphics[width=0.94\linewidth]{figures/experiments/optimization_legend.pdf}
\caption{\textit{Estimation errors for logistic regression with $d=20$, identity covariate covariance, Gaussian sketches, $\tau=10$, and $R=200$. Methods, quantities, and layout follow Figure~\ref{fig:optimization-linear}.}}
\label{fig:optimization-logistic}
\end{figure}

Figures~\ref{fig:optimization-linear} and~\ref{fig:optimization-logistic} show bias, MSE, and population objective gap over iterations for the same representative problems. We include SGD and exact online Newton as first- and second-order baselines, respectively. Sketched Newton with and without gradient averaging uses the same observations, sketches, and outer stepsizes. {\red Across both figures, the methods exhibit comparable error decay rates, consistent with our asymptotic results. Among the baselines, SGD performs better in the simpler, well-conditioned linear regression setting of Figure~\ref{fig:optimization-linear}, where second-order information provides limited additional benefits, whereas exact Newton performs comparably to SGD in the more complex logistic regression setting of Figure~\ref{fig:optimization-logistic} and even achieves lower averaged-iterate bias, benefiting from its ability to adapt to varying curvature.}
We also observe smaller MSE and objective gaps with gradient averaging toward the end of the runs. Table~\ref{tab:experiment-ratios} also shows smaller empirical bias in every matched sketching setting. Therefore, although the moving-average gradient is generally conditionally biased, the resulting parameter estimates have smaller bias in these experiments.

Gradient averaging requires no additional derivative evaluations and only $O(d)$ extra operations per iteration, preserving the $O(d^2)$ cost for a fixed sketching budget. Across the 48 matched sketching settings, the per-iteration time ratio with versus without gradient averaging has median $1.038$ and range $[0.877,1.564]$. Thus, the improvements in covariance and estimation accuracy require neither additional observations nor more sketching steps, with a median running-time increase of only $3.8\%$.

{\red \subsection{Statistical inference and computational cost}\label{app:experiment-inference}

We study statistical inference for the coordinate average $d^{-1}\sum_{j=1}^{d}x_j^\star$ to complement the covariance and estimation error comparisons in Section~\ref{sec:experiments}. The experiments use the same linear and logistic regression settings at $d=40$, with all three design covariance structures. For the sketched Newton methods, we use Kaczmarz sketching with $\tau=10$ steps at each iteration. After 100,000 updates, we construct nominal 95\% confidence intervals with a burn-in of 5,000 updates. The comparison includes both the last and averaged iterates of the exact and sketched Newton methods, as well as averaged SGD (ASGD). Table~\ref{tab:IC-coverage} reports empirical coverage, mean interval length, and mean absolute error (MAE) over 200 runs per setting. It also reports mean runtime and GFLOPs per run, including both iterate updates and confidence interval construction.

For exact and sketched Newton, Table~\ref{tab:IC-coverage} shows that the methods achieve comparably high estimation accuracy. Most MAEs are on the order of $10^{-4}$ or smaller, and all are below $2\times10^{-3}$. Empirical coverage for the Newton methods ranges from 89.5\% to 98.5\%, with most values close to the nominal level of 95\% and only the lowest one slightly below 90\%. Interval lengths reflect the uncertainty in the estimated coordinate average. The averaged iterates of exact Newton and Sketch ($\bbm_t$) generally yield shorter intervals than the other Newton estimators. For example, in linear regression with identity covariance, the averaged iterates of exact Newton and Sketch ($\bbm_t$) yield mean interval lengths of $2.567\times10^{-3}$ and $2.357\times10^{-3}$, respectively. The mean interval lengths of the other Newton estimators in this setting range from $3.455\times10^{-3}$ to $7.626\times10^{-3}$. This pattern is consistent with Theorem~\ref{thm:averaged-normality}, which shows that the averaged iterate of Sketch ($\bbm_t$) attains the same optimal limiting covariance $\Omega$ as averaged exact Newton.

The results also show that sketching reduces the computational cost of Newton inference, as measured by both FLOPs and runtime. Across all six settings, exact Newton uses 2.733--2.750 GFLOPs per run. Compared with exact Newton in the corresponding settings, Sketch ($\bg_t$) uses about 74\% fewer FLOPs and reduces runtime by 27--53\%, while Sketch ($\bbm_t$) uses about 73\% fewer FLOPs and reduces runtime by 24--39\%. These savings support our cost analysis, which gives an average cost per outer iteration of $O(d^2)$ for coordinate sketching with fixed $\tau$ and $O(d^3)$ for exact Newton, and illustrate the computational advantage of sketching solvers.

Table~\ref{tab:IC-coverage} also demonstrates the ``free lunch'' from gradient averaging: it improves both estimation accuracy and statistical efficiency with negligible additional computational cost. With the solver and sketching budget held fixed, Sketch ($\bbm_t$) yields lower MAEs and shorter confidence intervals than Sketch ($\bg_t$) for both the last and averaged iterates across all six settings. For example, in linear regression with Toeplitz covariance, gradient averaging reduces the MAE of the averaged iterate from $0.464\times10^{-3}$ to $0.279\times10^{-3}$ and its mean interval length from $2.704\times10^{-3}$ to $1.654\times10^{-3}$, with the same 93\% empirical coverage for both methods. These substantial improvements in estimation accuracy and statistical efficiency come with little extra computational cost, with the reported FLOP count increasing by only about 1.7\%.

As to ASGD, the results show that its estimation accuracy is comparable to that of the Newton methods. However, for inference, its nominal 95\% batch-means confidence intervals exhibit substantial undercoverage in all six settings, with empirical coverage of only 57--65\%. For example, in linear regression with identity covariance, despite the similar MAEs, ASGD achieves only 58.5\% empirical coverage, compared with 93.5\% for the averaged iterate of exact Newton. At 100,000 updates, ASGD indeed uses fewer FLOPs and runs faster than the Newton methods, as shown in Table~\ref{tab:IC-coverage}. However, its savings over sketched Newton are modest when measured in runtime. For example, in linear regression with equicorrelated covariates, ASGD takes 1.173 seconds per run on average, compared with 1.285 seconds for Sketch ($\bg_t$). These time savings offer limited benefit given the observed substantial undercoverage.

More importantly, for inference the relevant comparison is the time and computation required to reach a common coverage target. Under this requirement, ASGD shows no computational advantage over sketched Newton. We run an experiment in which we evaluate coverage every 1,000 updates and identify the first checkpoint with at least 90\% empirical coverage. Table~\ref{tab:coverage-endpoint-cost} reports the stopping step, mean runtime, and mean GFLOPs per run. All exact and sketched Newton methods reach the target within 100,000 updates. Many of these methods attain the required coverage at an early stage, with some reaching the target even within the first 10,000 updates. Their mean runtimes to reach the target range from 0.111 to 2.151 seconds per run. In contrast, ASGD does not reach 90\% coverage in any of the six settings within the 30-second limit per run. By that limit, ASGD has spent substantially more time than any Newton method needs to reach the target and has already used more FLOPs than the corresponding Newton method in almost all comparisons. In the few comparisons with lower FLOP counts for ASGD, those counts reflect only the computation performed before ASGD was stopped at the time limit without reaching the target, so they do not establish a computational advantage at the target coverage. The runtime and FLOP advantages of ASGD at a fixed number of updates therefore do not carry over to inference under a common coverage requirement.

Notably, Table~\ref{tab:coverage-endpoint-cost} also shows the benefits of gradient averaging at a common empirical coverage target of 90\%. In all six settings, the averaged iterate of Sketch ($\bbm_t$) reaches the required empirical coverage rate in fewer updates, with less time and fewer FLOPs than the averaged iterate of Sketch ($\bg_t$). These results complement the comparison at 100,000 updates: compared with sketched Newton without gradient averaging, the method with gradient averaging achieves higher statistical efficiency for a similar computational cost after a fixed number of updates and reaches the common coverage target sooner while using less computation.}

\begin{table}[t]
\centering
\captionsetup{font=footnotesize,skip=5pt}
\caption{{\red \textit{Inference for the coordinate average $d^{-1}\sum_{j=1}^{d}x_j^\star$ after 100,000 updates, with a 5,000-update burn-in. Empirical coverage is reported for nominal 95\% confidence intervals. Each entry summarizes 200 runs. CI length is the mean full interval length. For each setting, we highlight a method's coverage rate and mean CI length if no other method achieves comparably good coverage with a shorter mean interval. When both measures are comparable, we favor the method with fewer FLOPs per iteration.}}}
\label{tab:IC-coverage}
\footnotesize
\setlength{\tabcolsep}{4pt}
\begin{tabular}{llcrrrrr}
\toprule
\multicolumn{8}{l}{Linear regression} \\
Covariance & Method & Iterate
& \multicolumn{1}{c}{MAE} & \multicolumn{1}{c}{Coverage}
& \multicolumn{1}{c}{CI length} & \multicolumn{1}{c}{GFLOPs} & \multicolumn{1}{c}{Runtime} \\
& & & \multicolumn{1}{c}{$(10^{-3})$} & \multicolumn{1}{c}{(\%)}
& \multicolumn{1}{c}{$(10^{-3})$} & & \multicolumn{1}{c}{(s)} \\
\midrule
\multirow{7}{*}{Identity} & SGD & Average & 0.417 & 58.5 & 1.005 & 0.024 & 0.578 \\
\cmidrule(lr){2-8}
 & \multirow{2}{*}{Newton} & Last & 1.717 & 91.0 & 7.626 & \multirow{2}{*}{2.733} & \multirow{2}{*}{1.491} \\
 &  & Average & 0.410 & 93.5 & 2.567 & & \\
\cmidrule(lr){2-8}
 & \multirow{2}{*}{Sketch ($\bg_t$)} & Last & 1.530 & 93.0 & 7.122 & \multirow{2}{*}{0.711} & \multirow{2}{*}{0.697} \\
 &  & Average & 0.855 & 90.0 & 4.527 & & \\
\cmidrule(lr){2-8}
 & \multirow{2}{*}{Sketch ($\bbm_t$)} & Last & 0.781 & 90.5 & 3.455 & \multirow{2}{*}{0.723} & \multirow{2}{*}{0.916} \\
 &  & Average & 0.422 & \textcolor{red}{92.5} & \textcolor{red}{2.357} & & \\
\midrule
\multirow{7}{*}{\shortstack[l]{Toeplitz\\$r=0.4$}} & SGD & Average & 0.275 & 65.0 & 0.737 & 0.024 & 0.878 \\
\cmidrule(lr){2-8}
 & \multirow{2}{*}{Newton} & Last & 1.115 & 94.5 & 5.083 & \multirow{2}{*}{2.733} & \multirow{2}{*}{1.760} \\
 &  & Average & 0.273 & 94.0 & 1.672 & & \\
\cmidrule(lr){2-8}
 & \multirow{2}{*}{Sketch ($\bg_t$)} & Last & 1.415 & 93.5 & 6.486 & \multirow{2}{*}{0.711} & \multirow{2}{*}{1.038} \\
 &  & Average & 0.464 & 93.0 & 2.704 & & \\
\cmidrule(lr){2-8}
 & \multirow{2}{*}{Sketch ($\bbm_t$)} & Last & 0.949 & 93.0 & 4.092 & \multirow{2}{*}{0.723} & \multirow{2}{*}{1.192} \\
 &  & Average & 0.279 & \textcolor{red}{93.0} & \textcolor{red}{1.654} & & \\
\midrule
\multirow{7}{*}{\shortstack[l]{Equicorrelation\\$r=0.4$}} & SGD & Average & 0.103 & 57.0 & 0.269 & 0.024 & 1.173 \\
\cmidrule(lr){2-8}
 & \multirow{2}{*}{Newton} & Last & 0.409 & 96.0 & 1.850 & \multirow{2}{*}{2.733} & \multirow{2}{*}{1.990} \\
 &  & Average & 0.104 & 91.5 & 0.605 & & \\
\cmidrule(lr){2-8}
 & \multirow{2}{*}{Sketch ($\bg_t$)} & Last & 0.608 & 98.0 & 3.184 & \multirow{2}{*}{0.711} & \multirow{2}{*}{1.285} \\
 &  & Average & 0.169 & \textcolor{red}{93.0} & \textcolor{red}{1.052} & & \\
\cmidrule(lr){2-8}
 & \multirow{2}{*}{Sketch ($\bbm_t$)} & Last & 0.464 & 93.5 & 2.137 & \multirow{2}{*}{0.723} & \multirow{2}{*}{1.476} \\
 &  & Average & 0.118 & 90.5 & 0.745 & & \\
\midrule
\multicolumn{8}{l}{Logistic regression} \\
Covariance & Method & Iterate
& \multicolumn{1}{c}{MAE} & \multicolumn{1}{c}{Coverage}
& \multicolumn{1}{c}{CI length} & \multicolumn{1}{c}{GFLOPs} & \multicolumn{1}{c}{Runtime} \\
& & & \multicolumn{1}{c}{$(10^{-3})$} & \multicolumn{1}{c}{(\%)}
& \multicolumn{1}{c}{$(10^{-3})$} & & \multicolumn{1}{c}{(s)} \\
\midrule
\multirow{7}{*}{Identity} & SGD & Average & 0.038 & 59.0 & 0.097 & 0.036 & 0.854 \\
\cmidrule(lr){2-8}
 & \multirow{2}{*}{Newton} & Last & 0.086 & 92.0 & 0.394 & \multirow{2}{*}{2.750} & \multirow{2}{*}{2.140} \\
 &  & Average & 0.038 & 89.5 & 0.225 & & \\
\cmidrule(lr){2-8}
 & \multirow{2}{*}{Sketch ($\bg_t$)} & Last & 0.138 & 95.5 & 0.818 & \multirow{2}{*}{0.727} & \multirow{2}{*}{1.553} \\
 &  & Average & 0.235 & 93.0 & 1.105 & & \\
\cmidrule(lr){2-8}
 & \multirow{2}{*}{Sketch ($\bbm_t$)} & Last & 0.045 & 98.5 & 0.399 & \multirow{2}{*}{0.739} & \multirow{2}{*}{1.587} \\
 &  & Average & 0.116 & \textcolor{red}{95.5} & \textcolor{red}{0.592} & & \\
\midrule
\multirow{7}{*}{\shortstack[l]{Toeplitz\\$r=0.4$}} & SGD & Average & 0.042 & 57.0 & 0.103 & 0.036 & 1.051 \\
\cmidrule(lr){2-8}
 & \multirow{2}{*}{Newton} & Last & 0.097 & 94.0 & 0.435 & \multirow{2}{*}{2.750} & \multirow{2}{*}{2.414} \\
 &  & Average & 0.042 & 91.5 & 0.244 & & \\
\cmidrule(lr){2-8}
 & \multirow{2}{*}{Sketch ($\bg_t$)} & Last & 0.117 & 95.0 & 0.689 & \multirow{2}{*}{0.727} & \multirow{2}{*}{1.581} \\
 &  & Average & 0.182 & 93.5 & 0.927 & & \\
\cmidrule(lr){2-8}
 & \multirow{2}{*}{Sketch ($\bbm_t$)} & Last & 0.050 & 98.0 & 0.344 & \multirow{2}{*}{0.739} & \multirow{2}{*}{1.824} \\
 &  & Average & 0.098 & \textcolor{red}{96.5} & \textcolor{red}{0.479} & & \\
\midrule
\multirow{7}{*}{\shortstack[l]{Equicorrelation\\$r=0.4$}} & SGD & Average & 0.025 & 62.5 & 0.069 & 0.036 & 1.049 \\
\cmidrule(lr){2-8}
 & \multirow{2}{*}{Newton} & Last & 0.060 & 95.0 & 0.260 & \multirow{2}{*}{2.750} & \multirow{2}{*}{2.703} \\
 &  & Average & 0.025 & 92.5 & 0.155 & & \\
\cmidrule(lr){2-8}
 & \multirow{2}{*}{Sketch ($\bg_t$)} & Last & 0.063 & 98.5 & 0.525 & \multirow{2}{*}{0.727} & \multirow{2}{*}{1.603} \\
 &  & Average & 0.152 & 95.0 & 0.762 & & \\
\cmidrule(lr){2-8}
 & \multirow{2}{*}{Sketch ($\bbm_t$)} & Last & 0.030 & \textcolor{red}{96.5} & \textcolor{red}{0.275} & \multirow{2}{*}{0.739} & \multirow{2}{*}{1.898} \\
 &  & Average & 0.076 & 97.5 & 0.410 & & \\
\bottomrule
\end{tabular}
\end{table}

\begin{table}[t]
\centering
\captionsetup{font=footnotesize,skip=5pt,justification=justified}
\caption{\red {\textit{Time and computational cost to reach 90\% empirical coverage at $d=40$, with 200 runs per setting. Coverage is evaluated every 1,000 updates. The table reports the first checkpoint with at least 90\% coverage, along with the corresponding runtime and GFLOPs. If the threshold is not reached, the reported step is the mean number of updates completed by the 30-second cap, rounded to the nearest integer. Runtime and GFLOPs are per-run means that include iteration updates and confidence interval construction.}}}
\label{tab:coverage-endpoint-cost}
\footnotesize
\setlength{\tabcolsep}{3.9pt}
\begin{tabular}{@{}llcrrrrrr@{}}
\toprule
& & & \multicolumn{3}{c}{Linear regression} & \multicolumn{3}{c}{Logistic regression} \\
\cmidrule(lr){4-6}\cmidrule(lr){7-9}
Covariance & Method & Iterate
& \multicolumn{1}{c}{Stop step} & \multicolumn{1}{c}{Runtime} & \multicolumn{1}{c}{GFLOPs}
& \multicolumn{1}{c}{Stop step} & \multicolumn{1}{c}{Runtime} & \multicolumn{1}{c}{GFLOPs} \\
& & & & \multicolumn{1}{c}{(s)} & & & \multicolumn{1}{c}{(s)} & \\
\midrule
\multirow{7}{*}{Identity} & SGD & Average & 2,281,833 & 30.000 & 0.554 & 1,834,092 & 30.000 & 0.671 \\
\cmidrule(lr){2-9}
 & \multirow{2}{*}{Newton} & Last & 7,000 & 0.137 & 0.192 & 21,000 & 0.472 & 0.578 \\
 &  & Average & 10,000 & 0.196 & 0.274 & 53,000 & 1.197 & 1.457 \\
\cmidrule(lr){2-9}
 & \multirow{2}{*}{Sketch ($\bg_t$)} & Last & 22,000 & 0.264 & 0.156 & 32,000 & 0.454 & 0.232 \\
 &  & Average & 99,000 & 1.250 & 0.703 & 52,000 & 0.753 & 0.378 \\
\cmidrule(lr){2-9}
 & \multirow{2}{*}{Sketch ($\bbm_t$)} & Last & 22,000 & 0.304 & 0.159 & 20,000 & 0.312 & 0.148 \\
 &  & Average & 59,000 & 0.848 & 0.426 & 31,000 & 0.498 & 0.229 \\
\midrule
\multirow{7}{*}{\shortstack[l]{Toeplitz\\$r=0.4$}} & SGD & Average & 1,921,905 & 30.000 & 0.467 & 1,664,748 & 30.000 & 0.609 \\
\cmidrule(lr){2-9}
 & \multirow{2}{*}{Newton} & Last & 7,000 & 0.149 & 0.192 & 14,000 & 0.333 & 0.385 \\
 &  & Average & 11,000 & 0.235 & 0.301 & 67,000 & 1.598 & 1.842 \\
\cmidrule(lr){2-9}
 & \multirow{2}{*}{Sketch ($\bg_t$)} & Last & 9,000 & 0.111 & 0.064 & 29,000 & 0.453 & 0.211 \\
 &  & Average & 26,000 & 0.357 & 0.184 & 63,000 & 0.995 & 0.458 \\
\cmidrule(lr){2-9}
 & \multirow{2}{*}{Sketch ($\bbm_t$)} & Last & 10,000 & 0.146 & 0.072 & 26,000 & 0.451 & 0.192 \\
 &  & Average & 11,000 & 0.165 & 0.079 & 38,000 & 0.664 & 0.281 \\
\midrule
\multirow{7}{*}{\shortstack[l]{Equicorrelation\\$r=0.4$}} & SGD & Average & 1,921,647 & 30.000 & 0.467 & 1,674,177 & 30.000 & 0.613 \\
\cmidrule(lr){2-9}
 & \multirow{2}{*}{Newton} & Last & 8,000 & 0.172 & 0.219 & 12,000 & 0.282 & 0.330 \\
 &  & Average & 10,000 & 0.215 & 0.274 & 91,000 & 2.151 & 2.502 \\
\cmidrule(lr){2-9}
 & \multirow{2}{*}{Sketch ($\bg_t$)} & Last & 9,000 & 0.112 & 0.064 & 20,000 & 0.305 & 0.145 \\
 &  & Average & 13,000 & 0.170 & 0.092 & 38,000 & 0.603 & 0.276 \\
\cmidrule(lr){2-9}
 & \multirow{2}{*}{Sketch ($\bbm_t$)} & Last & 10,000 & 0.146 & 0.072 & 18,000 & 0.308 & 0.133 \\
 &  & Average & 10,000 & 0.145 & 0.072 & 29,000 & 0.510 & 0.214 \\
\bottomrule
\end{tabular}
\end{table}

\end{document}